\documentclass[review]{elsarticle}

\usepackage{hyperref}

\usepackage{multirow}
\usepackage{amsfonts}
\usepackage{algorithm}
\usepackage{stackengine}
\usepackage{algorithmicx}
\usepackage[noend]{algpseudocode}
\usepackage{balance}
\usepackage{dsfont}
\usepackage{amsmath}
\usepackage{xcolor}
\usepackage{enumerate}
\usepackage[caption=false,font=footnotesize,labelfont=sf,textfont=sf]{subfig}
\usepackage[absolute]{textpos}
\DeclareMathOperator*{\argmax}{argmax}
\DeclareMathOperator*{\argmin}{argmin}

\journal{Neurocomputing}

\begin{document}

\begin{textblock}{12}(2,0.3)
	\noindent Please cite as follows: K. Malialis, C. G. Panayiotou, M. M. Polycarpou, Nonstationary data stream classification with online active learning and siamese neural networks, Neurocomputing, Volume 512, Pages 235-252, 2022, doi: 10.1016/j.neucom.2022.09.065.  
\end{textblock}

\begin{frontmatter}

\title{
	Nonstationary Data Stream Classification with Online Active Learning and Siamese Neural Networks
\tnoteref{mytitlenote}}
\tnotetext[mytitlenote]{This work has been supported by the European Research Council (ERC) under grant agreement No 951424 (Water-Futures), by the European Union’s Horizon 2020 research and innovation programme under grant agreements No 883484 (PathoCERT) and No 739551 (TEAMING KIOS CoE), and from the Republic of Cyprus through the Deputy Ministry of Research, Innovation and Digital Policy.}


\author[a,b]{Kleanthis Malialis\corref{mycorrespondingauthor}}
\cortext[mycorrespondingauthor]{Corresponding author}
\ead{malialis.kleanthis@ucy.ac.cy}

\author[a,b]{Christos G. Panayiotou}
\ead{christosp@ucy.ac.cy}

\author[a,b]{Marios M. Polycarpou}
\ead{mpolycar@ucy.ac.cy}

\address[a]{KIOS Research and Innovation Center of Excellence, University of Cyprus, Cyprus}
\address[b]{Department of Electrical and Computer Engineering, University of Cyprus, Cyprus}

\begin{abstract}
We have witnessed in recent years an ever-growing volume of information becoming available in a streaming manner in various application areas. As a result, there is an emerging need for online learning methods that train predictive models on-the-fly. A series of open challenges, however, hinder their deployment in practice. These are, learning as data arrive in real-time one-by-one, learning from data with limited ground truth information, learning from nonstationary data, and learning from severely imbalanced data, while occupying a limited amount of memory for data storage. We propose the ActiSiamese algorithm, which addresses these challenges by combining online active learning, siamese networks, and a multi-queue memory. It develops a new density-based active learning strategy which considers similarity in the latent (rather than the input) space. We conduct an extensive study that compares the role of different active learning budgets and strategies, the performance with/without memory, the performance with/without ensembling, in both synthetic and real-world datasets, under different data nonstationarity characteristics and class imbalance levels. ActiSiamese outperforms baseline and state-of-the-art algorithms, and is effective under severe imbalance, even only when a fraction of the arriving instances' labels is available. We publicly release our code to the community.
\end{abstract}


\end{frontmatter}


\section{Introduction}
There is at present an emerging need for online learning algorithms that train predictive models on-the-fly as new information is continually becoming available in a diverse set of applications, such as, monitoring systems (e.g. fault detection in critical infrastructure systems \cite{kyriakides2014intelligent}, environmental monitoring \cite{ditzler2015learning}), security (e.g., spam filtering \cite{wang2018systematic}), finance (e.g., credit card fraud detection \cite{dal2015credit}), and recommender systems \cite{ditzler2015learning}. Despite the great potential impact of online predictive models, there are some key open challenges and problems that affect their deployability in practical applications:

\textbf{Limited labelled data}. Acquiring ground truth information (e.g., labels in classification tasks) as instances arrive one-by-one can be costly or impossible in some real-time applications, such as, fault detection in critical infrastructures.

\textbf{Nonstationary data}. It refers to the problem of having a data distribution that is unknown, and it evolves over time \cite{ditzler2015learning}. For example, nonstationarity can be caused by hardware faults (e.g., sensors in monitoring systems), changes in users’ behaviour (e.g., to evade detection in security systems), seasonality or periodicity effects (e.g., consumption in a water distribution network), or changes in users’ interests (e.g., in recommender systems).

\textbf{Imbalanced data}. Class imbalance refers to the problem of having a skewed distribution of data \cite{he2008learning}.
It renders a learning algorithm ineffective in identifying minority class examples as its performance declines significantly. The problem becomes considerably harder for unlabelled and nonstationary data \cite{wang2018systematic}.

Therefore, for certain applications it is desirable to design an online learning algorithm with these properties: (i) be able to respond in real-time as instances arrive one-by-one; (ii) be able to learn from limited labelled data, since in some applications it is not possible to rely solely on supervision; (iii) be able to adapt, as an algorithm without these capabilities would be ineffective under nonstationary conditions; and (iv) perform well under severe imbalance. Importantly, finding a good trade-off between these requirements is a challenging task.

An effective way to deal with limited labelled data is the active learning paradigm, in which the classifier queries a human expert for the labels of selected instances \cite{settles2009active}. A number of industry-scale applications have been realised using active learning, e.g., Google’s method for labelling malicious advertisements \cite{sculley2011detecting}, and NVIDIA's \cite{nvidia} and Tesla's \cite{tesla} methods for their autonomous vehicles with self-driving capability. This paper focuses on \textit{online active} learning. The key contributions of this work are the following:
\begin{enumerate}
	\item  We propose the ActiSiamese algorithm, which synergistically combines online active learning, siamese networks, and a multi-queue memory. While the majority of existing work focuses on uncertainty-based strategies, ActiSiamese proposes a new density-based strategy which considers similarity in the encoding (rather than the input) space. ActiSiamese aims to address the key challenges of online learning that were previously discussed.
	
	\item We conduct an extensive study that compares and examines the role of different active learning budgets and strategies (uncertainty-based, density-based), different neural network types (standard fully-connected, siamese), the performance with and without memory, the performance with and without ensembling, in both synthetic and real-world datasets, under different nonstationarity characteristics and imbalance severity levels. ActiSiamese outperforms strong baselines and state-of-the-art algorithms, and is effective under conditions of extreme imbalance, even when only a fraction of the arriving instances' labels is available.
	
	\item  We provide new insights into learning from nonstationary and imbalanced data streams, which constitutes a challenging and largely unexplored area even in the presence of supervision \cite{ditzler2015learning, wang2018systematic, malialis2020online}.
\end{enumerate}

ActiSiamese was first introduced in our brief conference paper \cite{malialis2020data}, and this work constitutes a significant extension of that. Unlike our preliminary paper: (i) We extend ActiSiamese to allow ensembling to better address nonstationarity; (ii) We examine the performance of ActiSiamese in real-world datasets, as well as in more synthetic datasets; (iii) We perform an extensive empirical analysis of ActiSiamese that examines the role of its various parameters; (iv) We perform an extensive comparative study as described above; and (iv) related work has been considerably enriched. We publicly release our code\footnote{https://github.com/kmalialis/actisiamese}.

The paper is organised as follows. Preliminary material necessary to understand the contributions made is provided in Section~\ref{sec:background}. Related work is presented in Section~\ref{sec:related}. ActiSiamese is described in Section~\ref{sec:proposed}. Our experimental setup is described in Section~\ref{sec:setup}. An empirical analysis of ActiSiamese is provided in Section~\ref{sec:analysis}. A comparative study is provided in Section~\ref{sec:comparative_study}.
A discussion on important remarks about ActiSiamese, its computational aspects, advantages and limitations is found in Section~\ref{sec:discussion}. We conclude in Section~\ref{sec:conclusion}.

\section{Preliminaries}\label{sec:background}
\textbf{Online} learning considers a data generating process that provides at each time $t$ a sequence of examples $S = \{S^t\}_{t=1}^T$, where $S^t = \{(x^t_i,y^t_i)\}^M_{i=1}$. The number of steps is denoted by $T \in [1, \infty)$ where the data are typically sampled from a long, potentially infinite, sequence. The number of examples at each step is denoted by $M$. If $M=1$, it is termed \textbf{one-by-one online} learning, otherwise it is termed \textbf{batch-by-batch online} learning \cite{ditzler2015learning}. The examples are drawn from an unknown probability distribution $p^{t}(x,y)$, where $x^t \in \mathbb{R}^d$ is a $d$-dimensional vector in the input space $X \subset \mathbb{R}^d$, $y^t \in \{1, ..., K\}$ is the class label in the target space $Y \subset \mathbb{Z}^+$, and $K \geq 2$ is the number of classes.

We focus on one-by-one learning, i.e., $S^t = (x^t, y^t)$, which is important for real-time monitoring. One-by-one learning requires the model to adapt immediately upon seeing a new example, and algorithms intended for batch-by-batch learning are, typically, not applicable for one-by-one learning tasks \cite{wang2018systematic}. A one-by-one online classifier receives a new instance $x^t$ at time $t$ and makes a prediction $\hat{y}^t$ based on a concept $h: X \to Y$. In \textbf{online supervised} learning, the classifier receives the true label $y^t$, its performance is evaluated using a loss function and is then trained based on the loss incurred. This process is repeated at each step. The gradual adaptation of the classifier without complete re-training $h^t = h^{t-1}.train(\cdot)$ is termed \textbf{incremental} learning \cite{losing2018incremental}. The rate or frequency of adaptation, and its associated costs and benefits, have been examined in \cite{zliobaite2015towards}.

In streaming tasks, however, the label cannot be typically provided, as it's either impossible (e.g. in real-time tasks) or costly / impractical. To address this issue, an alternative paradigm is \textbf{active} learning \cite{settles2009active}, which deals with strategies to selectively query for labels from a human expert according to a pre-defined ``budget'' $B \in [0,1]$, e.g., $B=0.2$ means that $20\%$ of the arriving instances can be labelled. A budget spending mechanism must ensure that the labelling spending $b \in [0,1]$ does not exceed the allocated budget.

In \textbf{online active} learning \cite{zliobaite2013active}, a classifier is built that receives a new instance $x^t$ at time $t$. At each time step the classifier calculates the prediction probability $\hat{p}(y | x^t)$. The classifier outputs the best prediction probability $h(x^t) = \max_y \hat{p}(y|x^t)$ and the predicted class $\hat{y}^t = \argmax_y \hat{p}(y|x^t)$. A given active learning strategy $\alpha : X \to \{False,True\}$ decides if the true label $y^t$ is required, which is assumed that an expert will provide. The classifier is evaluated using a loss function and is then trained based on the loss incurred. Note that training occurs only if the budget allows and when $\alpha(x^t) = True$.

Since data are sampled from a long, potentially infinite, sequence which is typically the case in data streams, it is unrealistic to expect that all acquired labels will be available at all times. The classifier should use a fixed amount of memory for data storage. If learning occurs on the most recent example, without using a memory, it is termed \textbf{one-pass} learning \cite{ditzler2015learning}. In such case, the cost $J$ at time $t$ is calculated using the loss function $l$ as follows $J=l(y^t,h(x^t))$.

A significant challenge encountered in some streaming applications is that of data \textbf{nonstationarity} \cite{ditzler2015learning, gama2014survey, lu2018learning}, typically caused by \textbf{concept drift}, which represents a change in the joint probability. The drift between steps $t_i$ and $t_j$, where $i \ne j$, is defined as:
\begin{equation}
\quad p^{t_i}(x,y) \neq p^{t_j}(x,y)
\end{equation}

Another major challenge encountered in some streaming applications is the presence of infrequent events, also known as \textbf{class imbalance} \cite{he2008learning}. It occurs when at least one class is under-represented, thus constituting a minority class. In binary classification, imbalance is defined as follows:
\begin{equation}
\exists y_0, y_1 \in Y \quad p^t(y=y_0) >> p^t(y=y_1),
\end{equation}
\noindent where $y_1$ represents the minority class.

\section{Related Work}\label{sec:related}

\subsection{Online supervised learning}
Methods that address drift fall into three groups. Memory-based methods use data storage (e.g., a sliding window) to maintain a set of recent examples that a classifier is trained on. Change detection-based methods use statistical tests to detect drift. Ensembling refers to a set of classifiers which can incorporate new concepts by adding new classifiers, and ``forget'' old concepts by discarding or updating existing ones \cite{brzezinski2018ensemble}. We direct the interested reader towards these excellent surveys \cite{ditzler2015learning, gama2014survey, krawczyk2017ensemble, lu2018learning, ramirezgallego2017survey, losing2018incremental, gomes2017survey}.

\subsubsection{Class imbalance}
The \textit{combined} problem of class imbalance and concept drift remains an open challenge \cite{aguiar2022survey, wang2018systematic}. There are two types of approaches, algorithm-level and data-level. Examples of algorithm-level methods include modified splitting criteria and distance metrics to make skew-insensitive the Decision Tree (e.g., \cite{ksieniewicz2021prior}) and Nearest Neighbour (e.g., \cite{vaquet2020balanced}) classifiers respectively. Genetic Programming has also been used, and this method \cite{cano2019evolving} increases skew-insensitive rule interpretability and recovery speed from drift. Cost-sensitive learning, one-class classification, and anomaly detection are other types of methods that can handle imbalance. We direct the interested reader towards these excellent surveys  \cite{aguiar2022survey, wang2018systematic}.

Examples of data-level approaches refer to resampling methods, which alter the training set to deal with the skewed distribution. In contrast to stationary environments, resampling methods require dedicated strategies or mechanisms to handle imbalance in data streams. The idea of having separate memories and, specifically, queues for each class to address imbalance is introduced by Queue-Based Resampling \cite{malialis2018queue}. Adaptive REBAlancing (AREBA) \cite{malialis2020online} extends this idea and proposes a dynamic mechanism to adaptively modify the size of the queues to constantly maintain separate and balanced queues per class. To address the problem of imbalance, resampling has been combined with ensembling, specifically, online bagging. Such examples, include, Oversampling-based Online Bagging (OOB) \cite{wang2015resampling}, Kappa Updated Ensemble (KUE) \cite{cano2020kappa}, and Robust Online Self-adjusting Ensemble (ROSE) \cite{cano2022rose}. Other ensembling methods, include, EONN \cite{ghazikhani2013ensemble}, ESOS-ELM \cite{mirza2015ensemble}, GRE \cite{ren2018gradual} and HEEM \cite{siahroudi2021online}.

While these algorithms have been shown to be effective, they typically rely on supervision  \cite{aguiar2022survey, wang2018systematic}.

\subsection{Online active learning}

\subsubsection{Querying strategies}\label{sec:related_strategies}
The most common strategy is \textbf{uncertainty sampling}, where the learner queries the most uncertain instances, which are typically found near the decision boundary \cite{lewis1994sequential}. The majority of existing active learning strategies assume the availability of all training examples $U \subset X$ (offline active learning) \cite{cohn1994improving}. One way to measure uncertainty \cite{settles2009active} is to first find the instance $x_q$ with the least confident best prediction:
\begin{equation}
x_q = \argmin_{x \in U} h(x)
\end{equation}
\noindent where $h(x) = \max_y \hat{p}(y|x)$ and request its label if it satisfies:
\begin{equation}
h(x_q) < \theta,
\end{equation}
where $\theta$ is a threshold which is typically fixed.

Some works consider batch-by-batch online active learning \cite{zhu2007active, lindstrom2010handling, lindstrom2013drift}. To address drift, \cite{zhu2007active} uses ensembling, \cite{lindstrom2010handling} uses a sliding window approach, while \cite{lindstrom2013drift} uses a change detection-based method. Work on one-by-one online active learning is limited. The arriving $x^t$ is queried if:
\begin{equation}
h(x^t) < \theta,
\end{equation}
\noindent where $h(x^t) = \max_y \hat{p}(y|x^t)$ and $\theta$ is a fixed threshold. This is called a \textit{fixed uncertainty sampling} strategy \cite{zliobaite2013active}.

This strategy may not perform well if the threshold is set incorrectly, or if the classifier learns enough so that the uncertainty remains above the fixed threshold most of the time. In \cite{zliobaite2013active} a variable uncertainty sampling strategy is proposed, which incorporates randomisation to ensure that the probability of labelling any instance is non-zero. This is called a \textit{randomised variable uncertainty sampling (RVUS)} strategy and the threshold is modified as follows:
\begin{equation}\label{eq:strategy}
\theta =
\begin{cases}
\theta (1 - s) & \text{if } h(x^t) < \theta_{rdm} \text{ \# request label}\\
\theta (1 + s) & \text{if } h(x^t) \geq \theta_{rdm} \text{ \# don't request}\\
\end{cases}
\end{equation}
\noindent where $s$ is a step size parameter, $\theta_{rdm} = \theta * \eta$ where $\eta$ follows a Normal distribution $\eta \sim N(1,\delta)$ with a standard deviation of $\delta$. Another work that proposes a variable uncertainty sampling strategy is \cite{cesa2006worst}.

\textbf{Query-by-committee} \cite{freund1997selective} is a popular approach where a set of classifiers, referred to as the committee, is in place. Each committee member ``votes'' or predicts the label and the most informative query is considered to be the instance they disagree the most. The majority are offline active learning methods. Early examples are \textit{query by bagging} and \textit{query by boosting} \cite{mamitsuka1998query} while another one is \textit{Active-DECORATE} \cite{melville2004diverse}, which focuses on building a diverse committee. An abstaining mechanism is proposed in \cite{korycki2019active} to temporarily remove uncertain classifiers, with dynamically adjusting the abstaining criterion in favour of minority classes. In \cite{krawczyk2019adaptive}, the problem has been approached as a multi-armed bandit problem, which obtains an efficient and adaptive ensemble active learning procedure by selecting the most competent classifier from the pool for each query.

An alternative strategy is \textbf{density sampling}, whose idea is that informative instances are not only those which lie near the decision boundary, but also those which lie in high density regions i.e. those which inhabit dense regions of the input space according to a distance or similarity metric. As with uncertainty sampling, the majority of work is on offline active learning where the set of all unlabelled instances $U \subset X$ is already available. In \cite{settles2008analysis} the instance queried for labelling is selected by its average similarity to other instances in $U$ as follows:
\begin{equation}
\argmax_{x \in U} \frac{1}{|U|} \sum_{x_u \in U} sim(x, x_u),
\end{equation}
where $x \ne x_u$ and $sim$ is a similarity function (e.g., cosine). This has also been applied in batch-by-batch online learning where an informative instance is the one which is similar to other unlabelled instances in the most recent batch \cite{capo2013active}.

\textbf{Hybrid} strategies have been proposed where various methods are combined to query even more informative instances. In \cite{settles2008analysis}, the authors combine offline density sampling with another strategy, such as, uncertainty sampling or query-by-committee. In \cite{capo2013active}, the authors use a batch-by-batch online active learning strategy where informative instances are selected based on uncertainty, density and overlap. A one-by-one online active learning strategy is proposed in \cite{liu2021online} that uses uncertainty and density sampling.

An important and desirable property of online active learning strategies is their data processing functionality \cite{lughofer2017line}.
Ideally, one-pass (or single-pass) ensures that data is loaded sample-wise, the sample is processed through the active learning strategy,  the classifier is updated if the strategy returned true, and then the sample is immediately discarded.
Examples include the aforementioned \cite{zliobaite2013active, cesa2006worst}, as well as \cite{chu2011unbiased, dasgupta2009analysis}. When storage is used (e.g., window), it is essential that a fixed amount of memory is allowed for any storage \cite{gama2014survey}.

Thus far, we have reviewed ``classical'' incremental approaches inspired by the machine learning and data mining community. An alternative line of research has focussed on enhanced evolving approaches (fuzzy models) inspired by the soft computing community, which can deal with significant novelty content, and allows knowledge expansion of the models on-the-fly based on selected samples \cite{lughofer2017line}. Representative works, include, \cite{lughofer2012single}, \cite{weigl2016improving} and \cite{pratama2016scaffolding}. We direct the interested reader towards \cite{lughofer2017line} for a comprehensive review of online active learning strategies, including, evolving (fuzzy) approaches.

For completeness, active learning has been combined with other learning paradigms, such as, semi-supervised and unsupervised learning \cite{dyer2013compose, abdallah2015adaptive, mohamad2020online}.


One contribution of this work is that we propose a new density-based online active learning strategy, whose novel characteristic is that the similarity is considered in the latent / encoding space rather than in the input / feature space. Also, no online active learning strategy addresses effectively the combination of drift and imbalance, while our work focuses exactly on this joint problem.

\subsubsection{Budget spending mechanisms}\label{sec:budget_spending}
One approach is to count the \textit{exact} labelling spending \cite{zliobaite2013active}. Its drawback is that the contribution of every next label will diminish over infinite time. One way to solve the aforementioned problem is to count the \textit{exact} labelling spending over a sliding window $b^t= \frac{u^t_w}{w}$, where $u^t_w$ is the number of instances queried within the sliding window $w$. This, however, contradicts the incremental learning concept as it needs to record previously labelling decisions. The authors in \cite{zliobaite2013active} approximate the labelling spending $\hat{b}^t= \frac{\hat{u}^t_w}{w}$ by approximating the number of instances queried within the sliding window as follows: $\hat{u}^t_w = \lambda \hat{u}^{t-1}_w + a(x^t)$, where $\lambda = \frac{w-1}{w}$ and $a(x^t)$ is a Boolean value that indicates if the true label for $x^t$ is queried or not. The authors prove that $\hat{b}$ is an unbiased estimate of $b$.

\section{Proposed Method}\label{sec:proposed}
\begin{figure}[t!]
	\centering
	\includegraphics[scale=0.35]{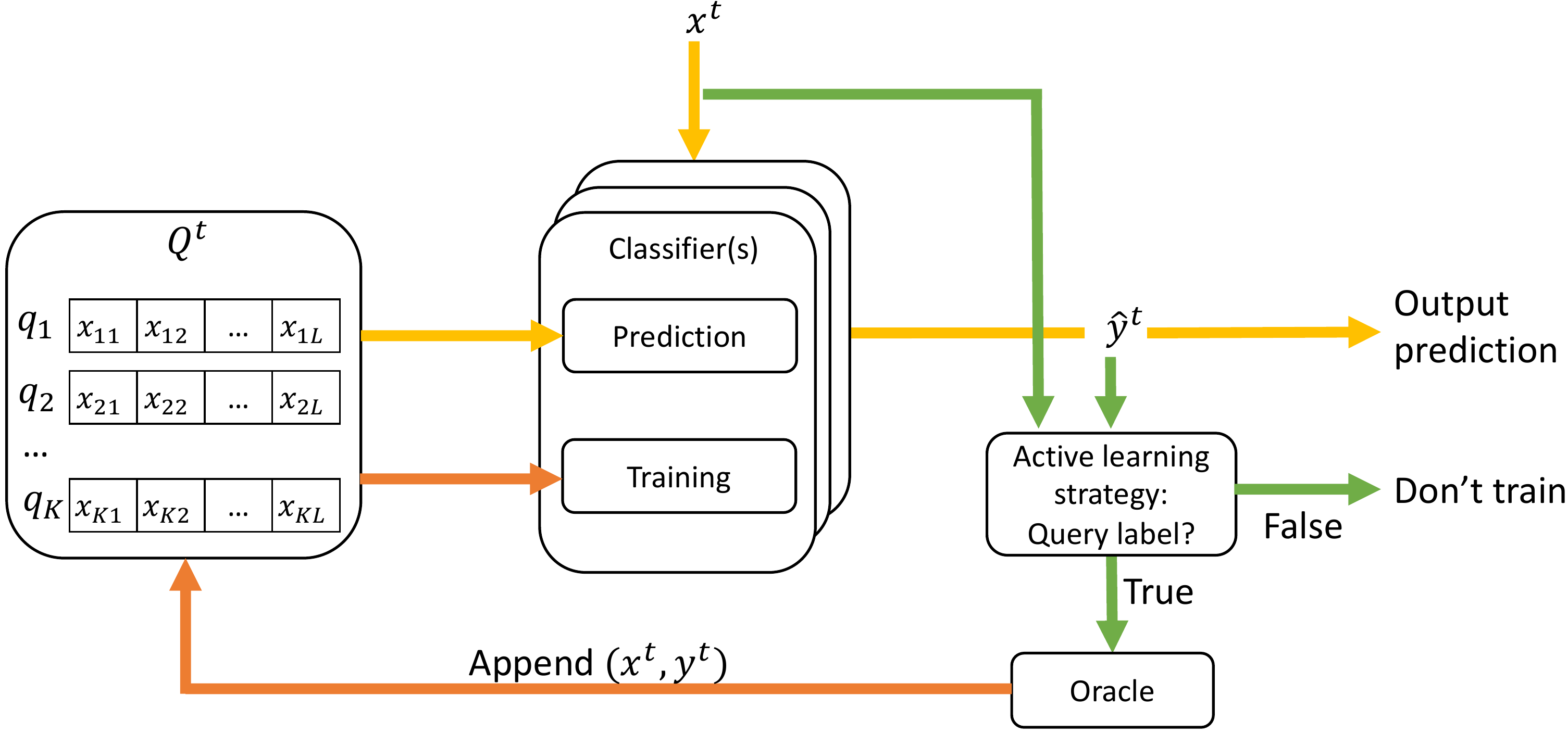}
	
	\caption{ActiSiamese's architecture which synergistically combines a multi-queue memory (left), a classifier or an ensemble of classifiers (middle), and online active learning (right).}
	
	\label{fig:overview}
\end{figure}

ActiSiamese's overview is shown in Fig~\ref{fig:overview}. The component $Q^t$ is a multi-queue memory where each class has its own sliding window which is implemented as a queue. In essence, this is ActiSiamese's way to deal with concept drift. The middle component denotes the classifier; notice that ensembling is supported. The component on the right depicts the active learning strategy. Specifically, at any time $t$, we observe an arriving instance $x^t$. The classifier considers the examples in $Q^t$, and provides the predicted class $\hat{y}^t$. The flow of information for the prediction part is shown in yellow. If the active learning strategy requests the label, this is provided by the oracle. Alternatively, nothing is done and the algorithm waits for the next arriving instance. This is shown in green. The flow of information for incrementally training the classifier is shown in orange. Each component is described below in detail. Towards the end of this section, we discuss our design choices, computational aspects, and other important remarks.

\subsection{Multi-queue memory}\label{sec:storage}
The framework uses multiple first-in-first-out (FIFO) queues that will be populated by examples queried by the active learning strategy. At any time $t$ we maintain a set of $K$ queues, one for each class as follows:
\begin{equation}
Q^t = \{q^t_c\}^K_{c=1},
\end{equation}
\noindent where $c$ is a class and $K \geq 2$ is the number of classes. All queues are of the same capacity $L$ and a queue corresponding to class $c$ is defined as follows:
\begin{equation}
q^t_c = \{x_{c,i}\}^L_{i=1},
\end{equation}
\noindent where for any two $x_{c,i}, x_{c,j} \in q^t_c$ such that $j > i$, $x_{c,j}$ has been observed more recently in time.

ActiSiamese assumes that the queues are initially full, i.e., it assumes the initial availability of $L$ labelled examples per class. While this may be difficult to have in practise, we will show that ActiSiamese is effective for very small values of $L$, e.g., up to ten. We argue that for the vast majority of applications this assumption is realistic. The assumption is not needed for problems in which imbalance does not exist, as the queues will be populated at a similar rate.

\subsection{ActiSiamese}\label{sec:actisiamese}
While the proposed architecture doesn't impose any restrictions on the selection of the classifier, we propose the use of Siamese neural networks which have been demonstrated (in the offline supervised learning framework) to be capable of few-shot learning, that is, learning from a few examples per class \cite{koch2015siamese}.

\begin{figure}[t!]
	\centering
	\includegraphics[scale=0.35]{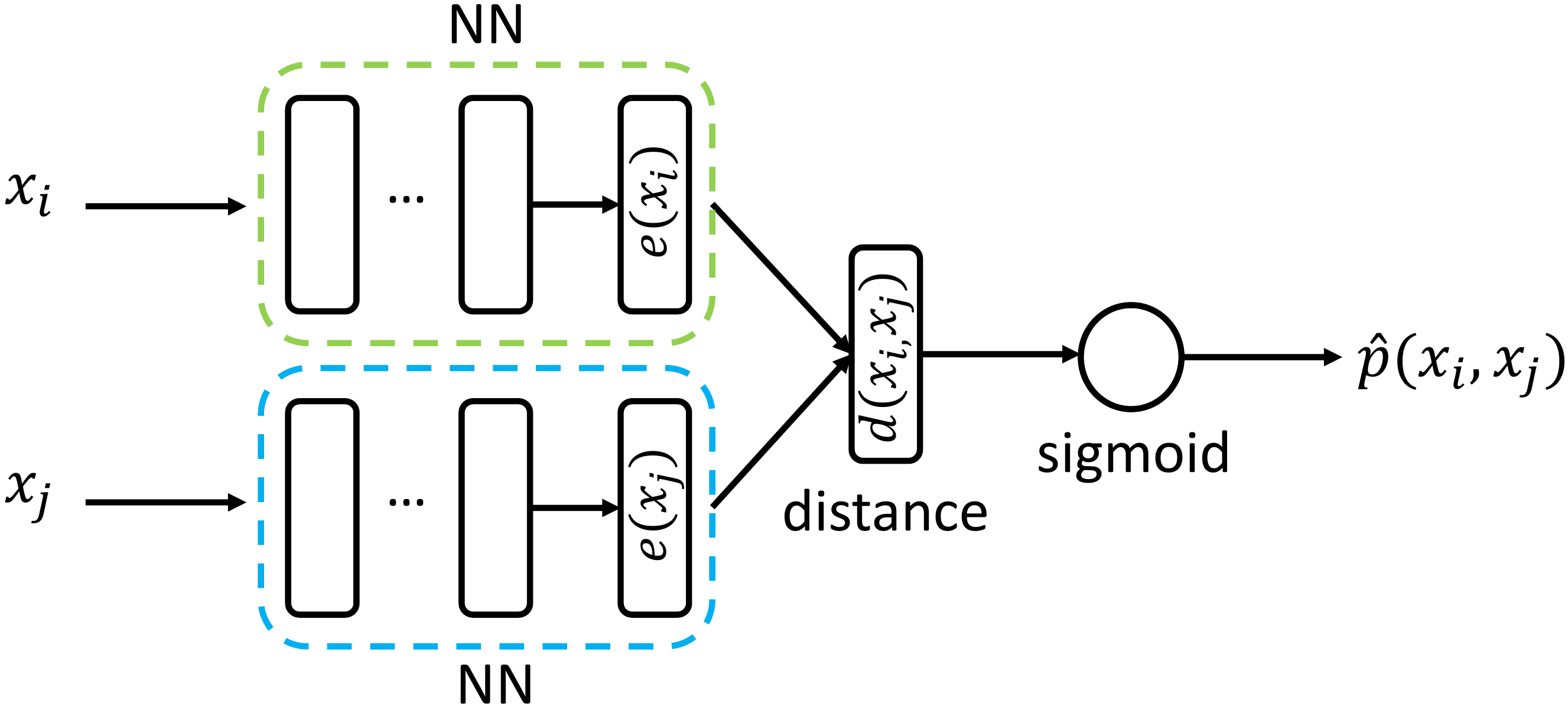}
	
	\caption{A siamese neural network.}
	
	\label{fig:nn_siamese}
\end{figure}

At the core of ActiSiamese lies a siamese neural network \cite{bromley1994signature} that consists of two identical neural networks (the ``twins'') as depicted in Fig~\ref{fig:nn_siamese}. The basic idea is to learn a function $e$ that maps an input pattern $x$ into a \textit{latent} space, thus forming its ``encoding'' $e(x)$, in such a way that a simple distance in the latent space approximates the neighbourhood relationships in the input space.

Traditional models, e.g. the \textit{k-nearest neighbour (k-NN)} algorithm, consider neighbourhood distances in the \textit{input} space and, it is for this reason that siamese networks have significantly outperformed these models in high-dimensional spaces, e.g., for image recognition \cite{koch2015siamese}.

Given a pair of examples $(x_i,x_j)$, the distance metric used is the element-wise absolute difference as in \cite{taigman2014deepface}:
\begin{equation}
d(x_i, x_j) = | e(x_i) - e(x_j) |
\end{equation}

\noindent The calculated distance is then provided to a sigmoid output unit. The siamese network $\hat{p}: X \times X \rightarrow [0,1]$ will learn to output a probability $\hat{p}(x_i, x_j)$ that indicates if the elements of the pair $(x_i,x_j)$ belongs to the same class.

\textbf{Class prediction}: To predict the class of the arriving $x^t$ (yellow part in Fig.~\ref{fig:overview}), ActiSiamese considers all examples in $Q^t$. For each queue, we find the average similarity of $x^t$ to its elements. We choose the queue with the highest average similarity:
\begin{equation}\label{eq:siamese_predict}
\hat{y}^t = \argmax_{c \in \{1, ..., K\}}  \frac{1}{L} \sum_{i=1}^L \hat{p}(x^t, x_{c,i}),
\end{equation}
\noindent where $K$ is the number of classes, and $x_{c,i} \in q^t_c$.

\textbf{Active learning strategy:} This work focuses on one-by-one learning strategies (green part in Fig.~\ref{fig:overview}). ActiSiamese proposes a randomised variable \textit{similarity} sampling (RVSS) strategy, where the selection criterion value is the maximum similarity in the predicted class:
\begin{equation}\label{eq:siamese_criterion}
v = \max_{i \in \{1, ..., L\}} \hat{p}(x^t, x_{c,i}),
\end{equation}
\noindent where $c$ is the class selected using Eq.~(\ref{eq:siamese_predict}). This criterion is compared to a variable threshold to determine whether or not to trigger selection, as follows:
\begin{equation}\label{eq:strategy_proposed}
\theta =
\begin{cases}
\theta (1 - s) & \text{if } v < \theta_{rdm} \text{ \# request label}\\
\theta (1 + s) & \text{if } v \geq \theta_{rdm} \text{ \# don't request}\\
\end{cases}
\end{equation}
\noindent where $s$ is a step size parameter, $\theta_{rdm} = \theta * \eta$ where $\eta$ follows a Normal distribution $\eta \sim N(1,\delta)$ with a standard deviation of $\delta$. This was inspired by RVUS \cite{zliobaite2013active} as shown in Eq.~(\ref{eq:strategy}), although, the selection criterion value is no longer $h(x^t) = \max_y \hat{p}(y|x^t)$, but instead it is the proposed $v$.

Notice that ActiSiamese uses a \textit{one-by-one} online active learning strategy. In other words, when the active learning strategy returns true, it requests the label of only the most recently observed instance $x^t$. Therefore, the proposed method does not need or store past unlabelled data. Recall that the memory (multi-queue) holds only some of the previously labelled data.

Lastly, we use the budget spending mechanism from \cite{zliobaite2013active} described in Section~\ref{sec:budget_spending}, however, any mechanism can be used which ensures that the labelling spending $b$ does not exceed the allocated budget $B$.

\textbf{Incremental training}: Recall that training (orange part in Fig.~\ref{fig:overview}) is initiated only when the active learning strategy requests and receives a class label. At time $t$, we generate from $Q^t$ all possible combinations of size two $C^t_2$. We then generate two subsets of $C^t_2$ as described below. The first subset contains all pairs in which the two examples belong to the same class, while the second subset contains all pairs in which the two examples belong to different queues:
\begin{align}
Q_{pos}^t = \{ \forall (x_{c_1, i_1}, x_{c_2, i_2}) \in C^t_2 | c_1 = c_2 \} \\
Q_{neg}^t = \{ \forall  (x_{c_1, i_1}, x_{c_2, i_2}) \in C^t_2 | c_1 \neq c_2 \}
\end{align}

Importantly, we ensure that the two sets are always balanced and, if necessary, we perform random downsampling to $Q_{neg}^t $. The training set is thus formed as follows:
\begin{equation}\label{eq:training_set}
Q_{train}^t = Q_{pos}^t \cup Q_{neg}^t
\end{equation}

Let $t-\Delta$ be the time of the last training, i.e., the last time a class label was requested and provided by an expert. Let also $t$ be the current time, and let's assume that the active learning strategy requested and received the class label. The cost function $J^t$ at time $t$ is defined as follows:
\begin{equation}\label{eq:siamese_cost}
J^t= \frac{1}{|Q^t_{train}|} \sum_{(x_i, x_j) \in Q^t_{train}} l(y_{i,j}, \hat{p}(x_i, x_j))
\end{equation}
\noindent where $y_{i,j} \in \{0,1\}$ is the ground truth and the loss function $l$ used is the binary cross-entropy. Learning is performed using incremental stochastic gradient descent where each neural network weight $w$ is updated according to the formula $w^t \leftarrow w^{t-\Delta} - \alpha \frac{\partial J^t}{w}$, where $\frac{\partial J^t}{w}$ is the partial derivative with respect to $w$, and $\alpha$ is the learning rate. ActiSiamese is trained using the backpropagation algorithm \cite{rumelhart1986learning}. We clarify that ActiSiamese's architecture remains fixed, and no layers are incrementally added; there is no evolution of any structural components.

ActiSiamese's pseudocode is shown in Algorithm~\ref{alg:actisiamese}.

\begin{algorithm}[h!]
	\caption{ActiSiamese}
	\label{alg:actisiamese}
	\begin{algorithmic}[1]
		
		\Statex \textbf{Input:}
		\State $a$: active learning strategy
		\State $B$: labelling budget
		\State $K$: number of classes
		\State $L$: queue length
		\State $D$: labelled data \Comment Optional, $|D| = K \times L$
		
		\Statex \textbf{Initialisation:}
		\State init queues $Q^0 = FIFO(num=K, capacity=L, init=D)$
		\State init budget expenses $b^0 = 0$
		\State create model $h^0$
		
		\Statex \textbf{Main:}
		\For{each time step $t \in [1, \infty)$}
		\State receive instance $x^t \in \mathbb{R}^d$
		\State predict class $\hat{y}^t$ using Eq.~(\ref{eq:siamese_predict})
		
		\State $Q^t = Q^{t-1}$
		\State $h^t = h^{t-1}$
		\If{$b^{t-1} < B$}\Comment expenses within budget
		\State calculate query criterion value $v$ using Eq.~(\ref{eq:siamese_criterion})
		\If{$a(x^t, v) == True$}\Comment label request using Eq.~\ref{eq:strategy_proposed}
		\State receive true label $y^t$
		\State append example $Q^t = Q^{t-1}.append(x^t,y^t)$
		\State prepare training pairs $Q^t_{train}$ using Eq.~(\ref{eq:training_set})
		\State calculate cost $J^t$ using Eq.~(\ref{eq:siamese_cost})
		\State update classifier $h^t = h^{t-1}.train(J^t)$
		\EndIf
		\EndIf
		
		\State update budget expenses $b^t$ \Comment Section~\ref{sec:budget_spending}
		\EndFor
		
	\end{algorithmic}
\end{algorithm}

\subsection{ActiSiamese-WM}\label{sec:actiwm}
This algorithm refers to a collection of $N$ classifiers $\{S_i\}_{i=1}^N$, where $S_i$ is an ActiSiamese classifier.

\textbf{Class prediction}:
In this work, we consider the popular \textit{Weighted Majority (WM)} algorithm \cite{littlestone1989weighted}. We maintain a pool of weights for each classifier: $\{w_i\}_{i=1}^N$. For each arriving instance $x^t$, each classifier provides a prediction probability $\hat{p}(c|x^t)$ for each class $c \in \{1, ..., K\}$. We then calculate  the weighted average probability for each class:
\begin{equation}
\{ \hat{p}^t_{avg}(c|x^t) \}_{c=1}^K
\end{equation}

The algorithm predicts the class of each $x^t$ as follows:
\begin{equation}
\hat{y}^t = \argmax_c \hat{p}^t_{avg}(c|x^t)
\end{equation}

The weight of each classifier $i$ is updated using the formula below:
\begin{equation}
w^t_i \leftarrow w^{t-1}_i e^{- \beta z^t_i},
\end{equation}
where $z^t_i$ is the zero-one (0/1) loss of classifier $i$ at time $t$ and $\beta$ is a pre-specified multiplicative factor \cite{littlestone1989weighted}.

\textbf{Active learning strategy}: We use the proposed randomised variable \textit{similarity} sampling strategy described in the previous section, however, the selection criterion value is now given as follows:
\begin{equation}
v = \max_{c \in \{1, ..., K\}} \hat{p}^t_{avg}(c|x^t)
\end{equation}

\textbf{Incremental training}: Each Siamese network $S_i$ is trained exactly as described earlier in Eq. (\ref{eq:siamese_cost}).

\section{Experimental Setup}\label{sec:setup}


\subsection{Datasets}

\subsubsection{Synthetic datasets}
These are necessary as they provide us with the flexibility to control the simulation conditions i.e. the imbalance rate, when to introduce drift, and the nature of the drift.

\begin{figure}[h]
	\centering
	
	\subfloat[\textit{sea}]{\includegraphics[scale=0.12]{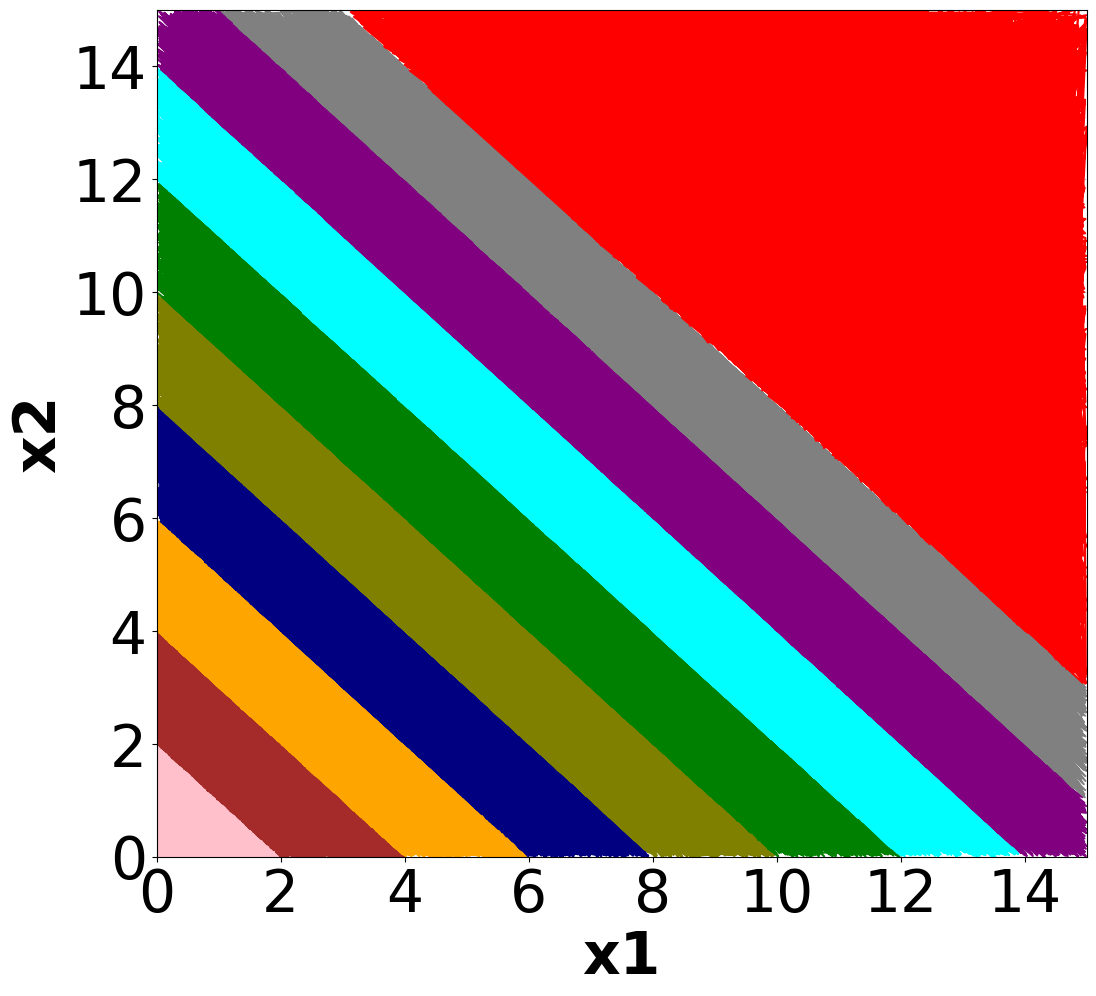}%
		\label{fig:sea}}
	\subfloat[\textit{sea} drifted]{\includegraphics[scale=0.12]{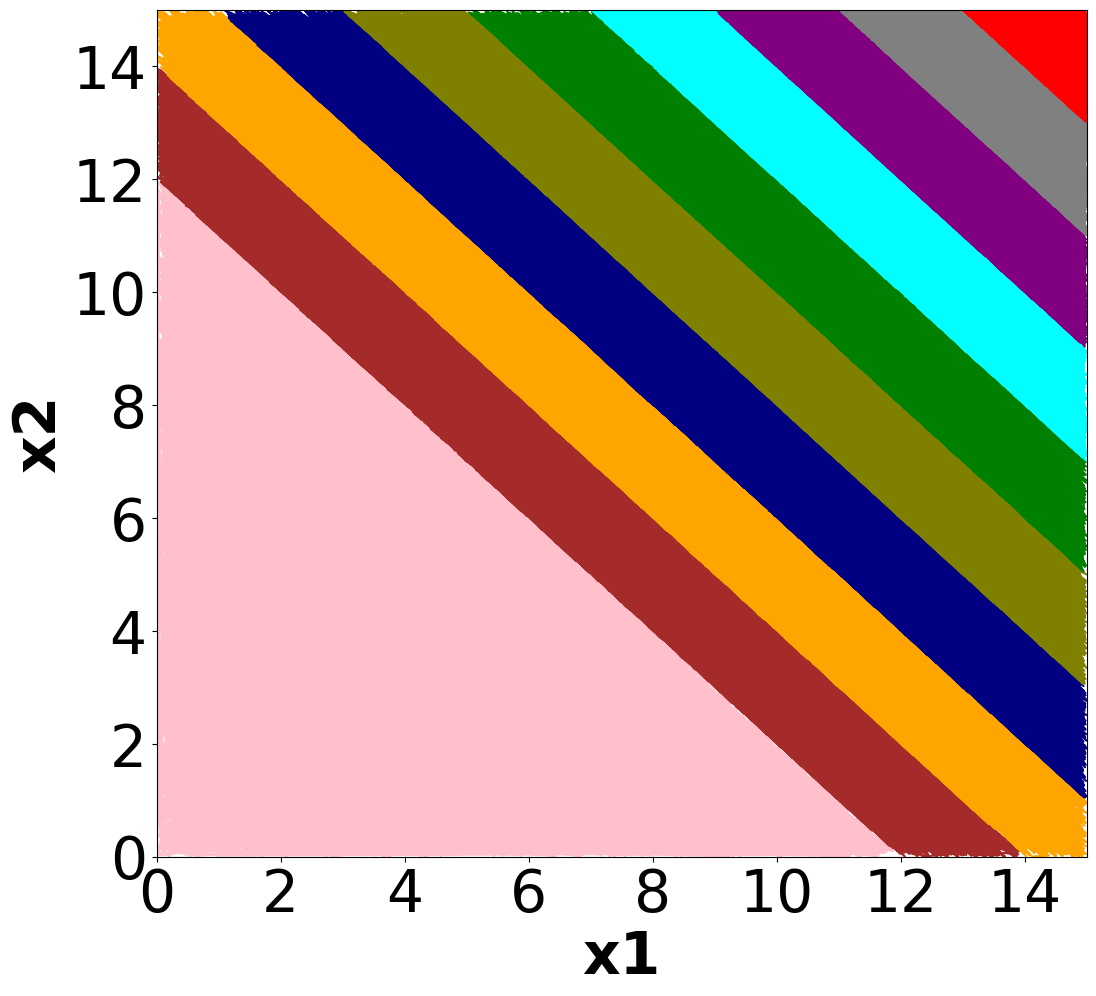}%
		\label{fig:sea_drifted}}
	
	\subfloat[\textit{circles}]{\includegraphics[scale=0.12]{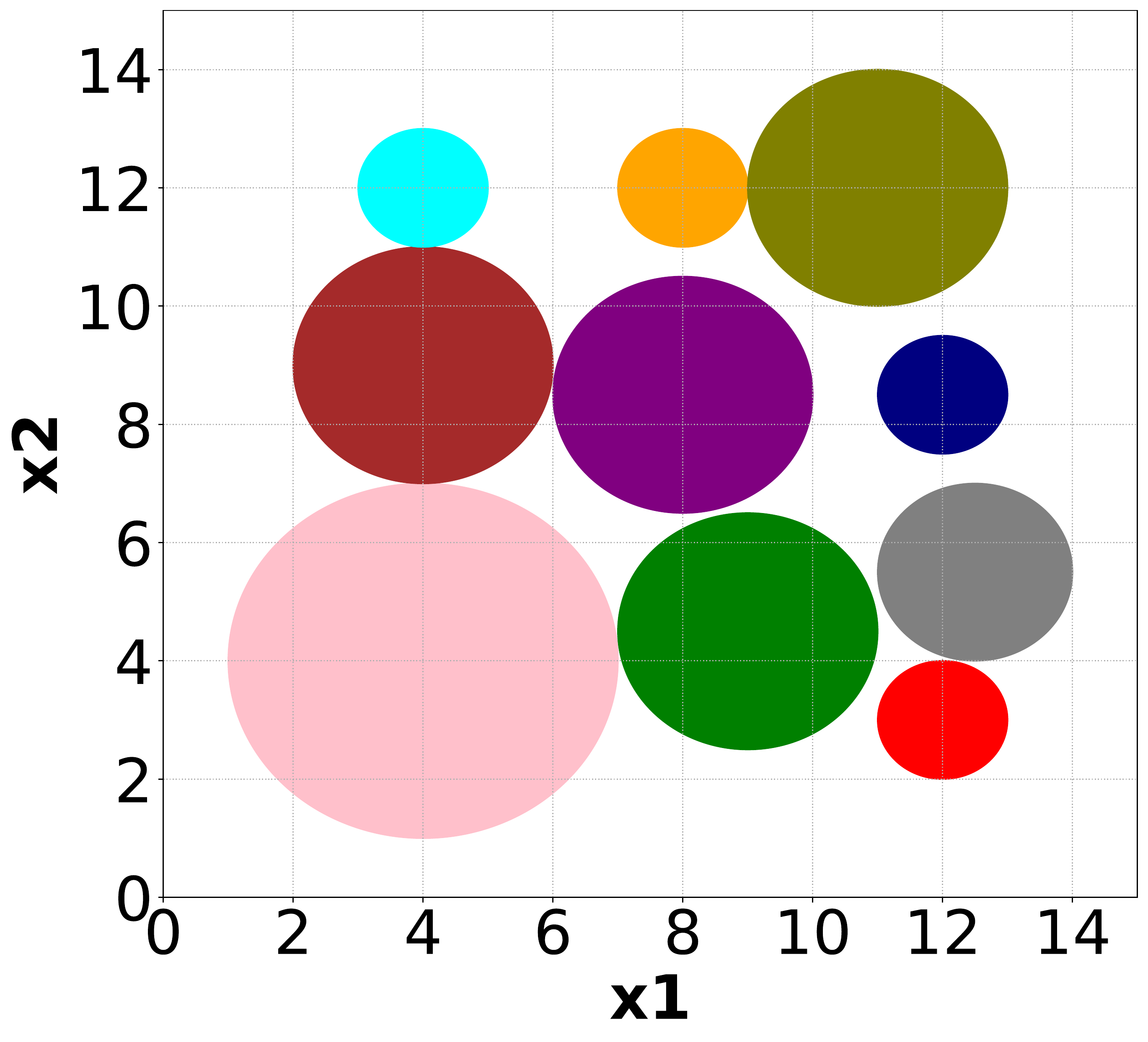}%
		\label{fig:circles}}
	\subfloat[\textit{circles} drifted]{\includegraphics[scale=0.12]{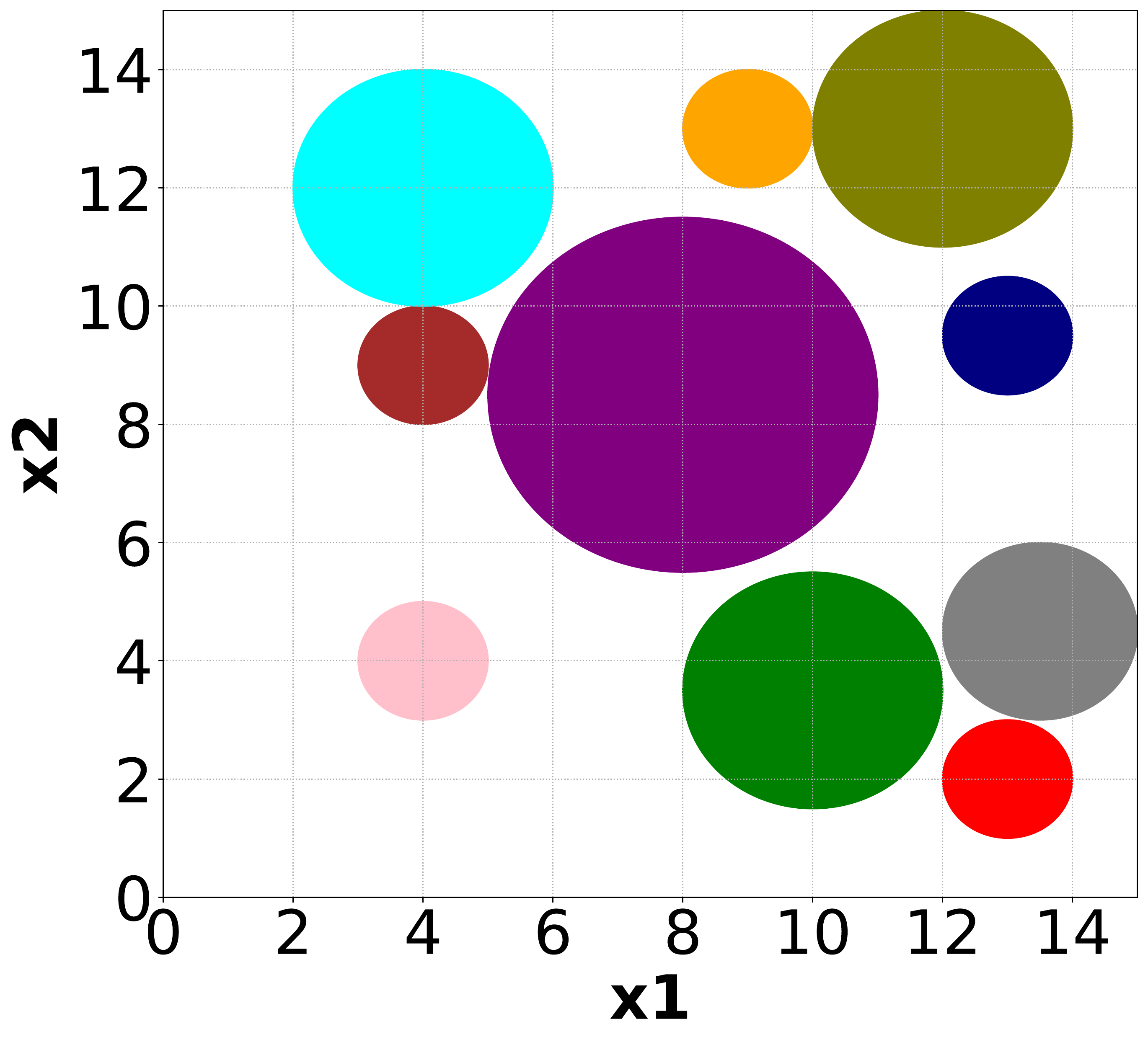}%
		\label{fig:circles_drifted}}
	
	\subfloat[\textit{blobs}]{\includegraphics[scale=0.27]{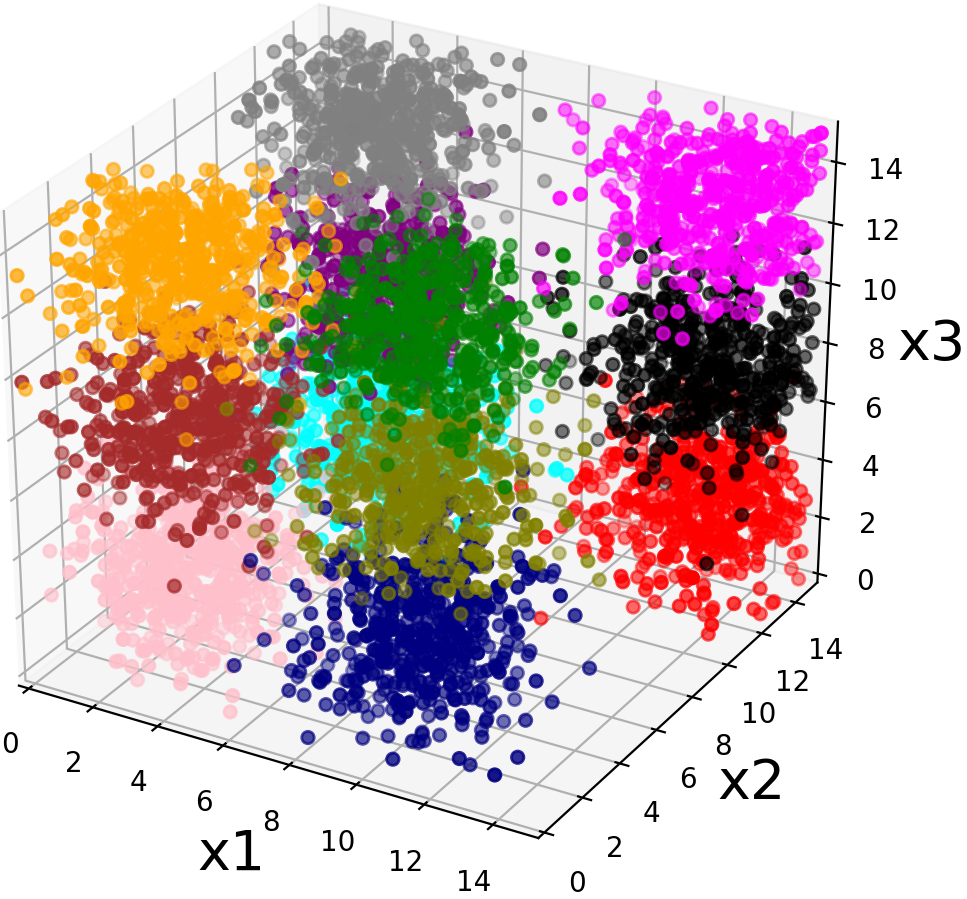}%
		\label{fig:blobs}}
	\subfloat[\textit{blobs} drifted]{\includegraphics[scale=0.27]{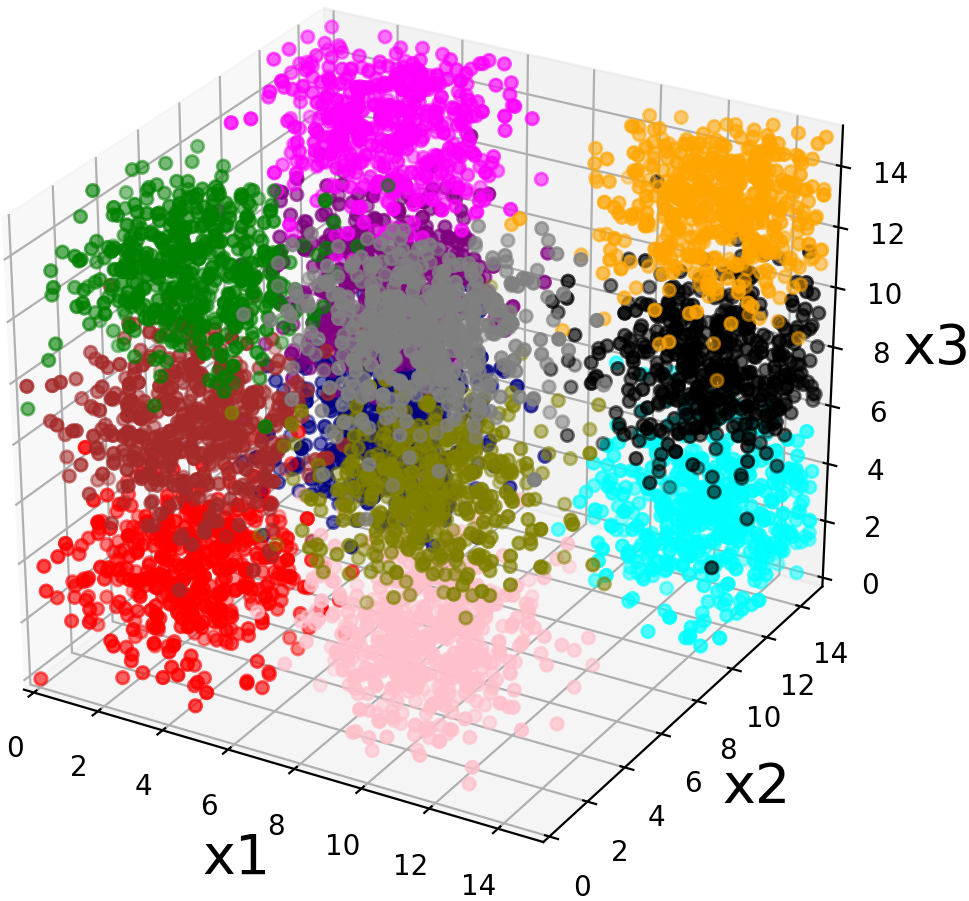}%
		\label{fig:blobs_drifted}}
	
	\caption{The synthetic data used in our experimental study.}
\end{figure}

\textbf{sea} \cite{street2001streaming}: It consists of two features $x_1, x_2 \in [0, 15]$. The original dataset was a binary classification problem; we have extended it here to ten classes. The decision boundaries are shown in Fig.~\ref{fig:sea} and defined as follows:
\begin{equation}
\begin{split}
\rho_0 & \leq x_1 + x_2 < \rho_1 \rightarrow \text{class 1 (pink)}\\
\rho_1 & \leq x_1 + x_2 < \rho_2 \rightarrow \text{class 2 (brown)}\\
...\\
\rho_9 & \leq x_1 + x_2 < \rho_{10} \rightarrow \text{class 10 (red)}\\
\end{split}
\end{equation}
The parameter values are $\rho = <0, 2, 4, ..., 18, 30>$. We will consider a version of this dataset with posterior concept drift as depicted in Fig.~\ref{fig:sea_drifted}, where the parameters become $\rho = <0, 12, 14, ..., 30>$. We will consider cases where the drift occurs abruptly and recurrently. Furthermore, we will consider a version of \textit{sea} with prior drift, where initially the probability of all classes is $p(y) = 0.1$. After a specified time, this will change to a multi-minority imbalance scenario.

\textbf{circles} \cite{gama2004learning}: It consists of two features $x_1, x_2 \in [0,15]$ and ten classes as shown in Fig.~\ref{fig:circles}. Each class is represented by a circle of the form $(x_1 - x_{1c})^2 + (x_2 - x_{2c})^2 = r^2_c$ where $(x_{1c}, x_{2c})$ and $r_c$ are its centre and radius respectively. Like with the \textit{sea} dataset, we will consider posterior (depicted in Fig.~\ref{fig:circles_drifted}) and prior types of drift, as well as abrupt and recurrent drift.

\textbf{blobs}: It consists of three features $x_1, x_2, x_3 \in [0,15]$ and 12 classes as shown in Fig.~\ref{fig:blobs}. Each class is an isotropic Gaussian blob and noise exists due to the standard deviation of the blobs. As before, we will consider abrupt and recurrent posterior concept drift as depicted in Fig.~\ref{fig:blobs_drifted}.

Notice that these datasets exhibit a different level of posterior drift severity, specifically, drift is mild, severe and extreme in the \textit{circles}, \textit{blobs} and \textit{sea} data respectively. Moreover, we will consider abrupt and recurrent types of drift.

Furthermore, we produce imbalanced versions of the above datasets. We consider the more challenging multi-minority class imbalance scenarios \cite{wang2012multiclass}. We have chosen the pink class to be the majority class, and the rest constitute minority classes with the same imbalance rate. We consider cases of severe ($1\%$) and extreme ($0.1\%$) imbalance. For example, for \textit{sea} where the number of classes is $K=10$, the extreme multi-minority scenario would correspond to: $p(y=pink) = 1.0 - (K-1) \times p(y \in minority) = 1.0 - 9 \times 0.001 = 0.991$.

\textbf{Two Patterns} \cite{geurts2002contributions}: A simulated time series dataset in which each class represents the presence of two step patterns (up, down). There are 128 features, 5000 examples, 4 classes (down-down, up-down, down-up, up-up), and it is a balanced dataset.

\textbf{Moving Squares} \cite{losing2016knn}: Four equidistantly separated, squared uniform distributions are moving in horizontal direction with constant speed. The direction is inverted whenever the leading square reaches a predefined boundary. Each square represents a different class. There is a predefined time horizon of 120 examples before old instances may start to overlap current ones.

\textbf{Interchanging RBF} \cite{losing2016knn}: Fifteen Gaussians with random covariance matrices are replacing each other every 3000 samples. Thereby, the number of Gaussians switching their position increases each time by one until all are simultaneously changing their location. Altogether 67 abrupt drifts are occurring within this dataset.

\subsubsection{Real-world datasets}
These are high-dimensional, noisy, and more challenging than synthetic datasets, however, the true nature of concept drift may be unknown.

\textbf{gestures} \cite{gestures}: The task is to classify four human gestures from electrical activity of muscles (electromyography). Eight sensors are placed on skin surface, and each arriving instance includes eight consecutive readings i.e. 64 features. This dataset is balanced as the number of arriving instances per class is 2500.

\textbf{Forest} \cite{blackard1999comparative}: It contains cartographic information from the U.S. Forest Service. The task is to predict the forest cover type for given 30x30m cells from the Roosevelt National Forest in Colorado. There are seven forest cover types each with 20000, 30000, 3500, 275, 1000, 1500, 2000 instances respectively.

\textbf{MNIST} \cite{lecun1998gradient}: The dataset consists of handwritten digits (``0''-``9'') where each image is of size 28x28. While it is, typically used as a benchmark for training image classification systems, we use it in our study as it can stress test streaming algorithms with its high dimensionality of 784 features. We consider two multi-minority scenarios where the majority class is the digit ``0'' and the rest of the digits are the minority classes. In the first scenario the imbalance level is 10\%, that is, 5000 arriving images of digit ``0'', and 500 from each of the remaining digits. In the second scenario the imbalance level is 1\%, that is, 5000 arriving images of digit ``0'', and 50 from each of the remaining digits.

\textbf{Keystroke} \cite{souza2015data}: The dataset was constructed after 51 users were requested to type a specific password which was captured in eight sessions over different days. The task is to identify four different users based on their typing rhythm. It has 1600 examples and ten features which are extracted from the ``flight time'' for each pressed key; defined as the difference between the times when a key is released and when the next one is pressed.

\textbf{UWave Gesture Library Z} \cite{liu2009uwave}: A time series dataset for a set of eight simple gestures generated from accelerometers using the Wii remote. The data consists of the Z coordinates of each motion. There are 8 classes with total 3582 samples with 315 features.

\textbf{Insects} \cite{souza2013classification}: These are data generated by a low-cost laser sensor, where the task is to classify species of insects. This task is important as it constitutes a step towards the development of intelligent traps, which will be able to capture species of interest (e.g., for pest control, during diseases). The data stream is nonstationary as the insects’ metabolisms are influenced by environmental conditions (e.g. temperature), circadian rhythm, and age. The number of features are 33 and the imbalance ratio is ~20\%.

\subsection{Compared methods}\label{sec:setup_methods}

\textbf{RVUS}: The seminal work \cite{zliobaite2013active} described in Sec.~\ref{sec:related_strategies} and shown in Eq.~(\ref{eq:strategy}). It uses an uncertainty-based active learning strategy, a standard neural network, and it is a one-pass learner (no memory).

\textbf{RVUS-WM}: The ensemble version of RVUS with size is $N=10$ and the multiplicative factor of the Weighted Majority algorithm is set to $\beta = 0.5$.

\textbf{ActiQ}: A state-of-the-art uncertainty-based algorithm \cite{malialis2020data} which uses the RVUS, but within the proposed architecture shown in Fig.~\ref{fig:overview}. In other words, similarly to RVUS it uses a standard neural network, and similarly to ActiSiamese it uses the multi-queue memory.

\textbf{ActiQ-WM}: The ensemble version of ActiQ.

\textbf{RVSS}: A randomised variable similarity sampling active learning strategy using cosine similarity as in \cite{settles2009active}. It uses the memory (multi-queue) from the proposed architecture (Fig.~\ref{fig:overview}). Similarity is considered in the input space (rather than the latent space).

\textbf{RVSS-WM}: The ensemble version of RVSS.

\textbf{ActiSiamese}: The proposed method described in Section~\ref{sec:actisiamese} and Algorithm~\ref{alg:actisiamese}. It uses a similarity-based active learning strategy, a Siamese network, and it is a memory-based method. Similarity is considered in the latent space.

\textbf{ActiSiamese-WM}: The ensemble version of ActiSiamese (Section~\ref{sec:actiwm}).

Our comparative study examines many crucial aspects of learning in data streams, which are: i) one-pass learning (RVUS) vs memory-based (RVSS, ActiQ, ActiSiamese); ii) uncertainty-based (RVUS, ActiQ) vs similarity-based (RVSS, ActiSiamese) sampling strategies; iii) similarity-based in the input space (RVSS) vs the latent space (ActiSiamese); and iv) single classifier (RVUS, RVSS, ActiQ, ActiSiamese) vs ensembling (RVUS-WM, RVSS-WM, ActiQ-WM, ActiSiamese-WM).

We have tried to make the comparison as fair as possible. First, we ensure that no \textit{offline} learning, i.e., no pre-training takes place. This means that \textit{all} methods start learning online at time $t=1$ even if ActiQ and ActiSiamese have access to $E$ initial examples per class. Second, all active learning strategies use the same parameter values, as suggested by \cite{zliobaite2013active}: step size $s=0.01$, randomisation threshold $\delta = 1.0$ and sliding window size $w=300$. Third, all methods share the same classifier per dataset, which is standard neural network. For ActiSiamese, it is replicated to form the ``twin network". 

Classifiers are implemented in Keras \cite{chollet2015keras}. For reproducibility, the hyper-parameter values for synthetic datasets are shown in Table~\ref{tab:hyperparameters_synthetic}. The hyper-parameter values for real-world datasets are shown in Table~\ref{tab:hyperparameters_real}; any values not shown are the same as in the synthetic datasets. Notice that the output activation and loss function depend on the type of the neural network used. The standard fully-connected network learns the probability of an arriving example $x^t$ belonging to each class, i.e., $\hat{p}(y | x^t)$. As a result, the categorical cross-entropy loss function and the softmax output activation are used. In contrast, the Siamese network learns to output a probability $\hat{p}(x_i,x_j)$ that indicates if the elements of the pair of examples $(x_i, x_j)$ belongs to the same class (Fig.~\ref{fig:nn_siamese}). As a result, the binary cross-entropy and the sigmoid output activation are used.

%
%
%

\begin{table}
	\centering
	\caption{Hyper-parameter values for synthetic data}
	\label{tab:hyperparameters_synthetic}
	\resizebox{\columnwidth}{!}{%
		\begin{tabular}{l|l|l|l|l|c|c}
			& \textbf{Sea} & \textbf{Circles} & \textbf{Blobs} & \textbf{Two Patterns}             & \textbf{\textbf{Interchanging RBF}} & \textbf{Moving~\textbf{Squares}}  \\ 
			\hline
			\textbf{Learning rate}      & \multicolumn{3}{c|}{0.01}                        & \multicolumn{3}{c}{0.001}                                                                                  \\ 
			\hline
			\textbf{Hidden layers}      & \multicolumn{3}{c|}{{[}32, 32]}                  & \multicolumn{1}{c|}{{[}256, 256]} & {[}64, 64]                         & {[}8, 32]                         \\ 
			\hline
			\textbf{Mini-batch size}    & \multicolumn{3}{c|}{64}                          & \multicolumn{3}{c}{128}                                                                                    \\ 
			\hline
			\textbf{Weight initialiser} & \multicolumn{6}{c}{He Normal \cite{he2015delving}}                                                                                                                                 \\ 
			\hline
			\textbf{Optimiser}          & \multicolumn{6}{c}{Adam \cite{kingma2014adam}}                                                                                                                                      \\ 
			\hline
			\textbf{Hidden activation}  & \multicolumn{6}{c}{Leaky ReLU($\lambda = 0.01$) \cite{maas2013rectifier}}                                                                                                              \\ 
			\hline
			\textbf{Num. of epochs}     & \multicolumn{6}{c}{1}                                                                                                                                         \\ 
			\hline
			\textbf{Output activation}  & \multicolumn{6}{c}{Binary (ActiSiamese) / Categorical cross-entropy }                                                                                                                      \\ 
			\hline
			\textbf{Loss function}      & \multicolumn{6}{c}{Sigmoid (ActiSiamese) / Softmax}                                                                                                                                  
		\end{tabular}
	}
\end{table}

\begin{table}
	\centering
	\caption{Hyper-parameter values for real-world data}
	\label{tab:hyperparameters_real}
		\resizebox{\columnwidth}{!}{%
	\begin{tabular}{l|c|c|c|c|c|c}
		\multicolumn{1}{c|}{}    & \textbf{Gestures} & \textbf{Insects} & \textbf{\textbf{Forest}}        & \textbf{Keystroke} & \textbf{\textbf{UWave Gestures}} & \textbf{MNIST}  \\ 
		\hline
		\textbf{Learning rate}   & \multicolumn{5}{c|}{0.001}                                                                                                     & 0.0001          \\ 
		\hline
		\textbf{Hidden layers}   & \multicolumn{2}{c|}{{[}128, 128]}    & \multicolumn{1}{l|}{{[}32, 32]} & {[}64, 128]        & {[}512, 1024]                    & {[}1024, 1024]  \\ 
		\hline
		\textbf{Mini-batch size} & \multicolumn{6}{c}{128}                                                                                                                         
	\end{tabular}
}
\end{table}

\subsection{Performance metrics and Evaluation method}\label{sec:evaluation}
Classifiers are typically evaluated using the overall accuracy metric. In the presence of imbalance, however, this metric becomes unsuitable as it is biased towards the majority class(es) \cite{he2008learning}. A widely accepted performance metric which is not sensitive to imbalance is the geometric mean: \cite{sun2006boosting}: 
\begin{equation}\label{eq:gmean}
G\text{-}mean = \displaystyle\sqrt[K]{\prod_{c=1}^K R_c},
\end{equation}
\noindent where $R_c$ is the recall of class $c$ and $K$ is the number of classes. Not only $G\text{-}mean$ is not sensitive to imbalance, it has some desirable properties as it is high when all recalls are high and when their difference is small \cite{he2008learning}.

To evaluate and compare sequential learning algorithms, we use the popular \textit{prequential evaluation with fading factors} method. It has been proven to converge to the Bayes error when learning in stationary data \cite{gama2013evaluating}. Moreover, a major advantage is that it does not require a holdout set and the predictive algorithm is always tested on unseen data. The fading factor is set to $\xi = 0.99$. 
In all simulations we plot the prequential $G\text{-}mean$ in every step averaged over 20 repetitions, including the error bars displaying the standard error around the mean. In some plots the error bars are very small but they are always included.

Moreover, we test for statistical significance using a one-way repeated measures ANOVA and then using posthoc multiple comparisons tests with Fisher’s least significant difference correction procedure to show which of the compared method is significantly different from the others.

\section{Empirical Analysis of ActiSiamese}\label{sec:analysis}

\subsection{Effect of the budget \textit{B}}
\begin{figure*}[t!]
	\centering
	
	\subfloat[sea ($t=20000$)]{\includegraphics[scale=0.13]{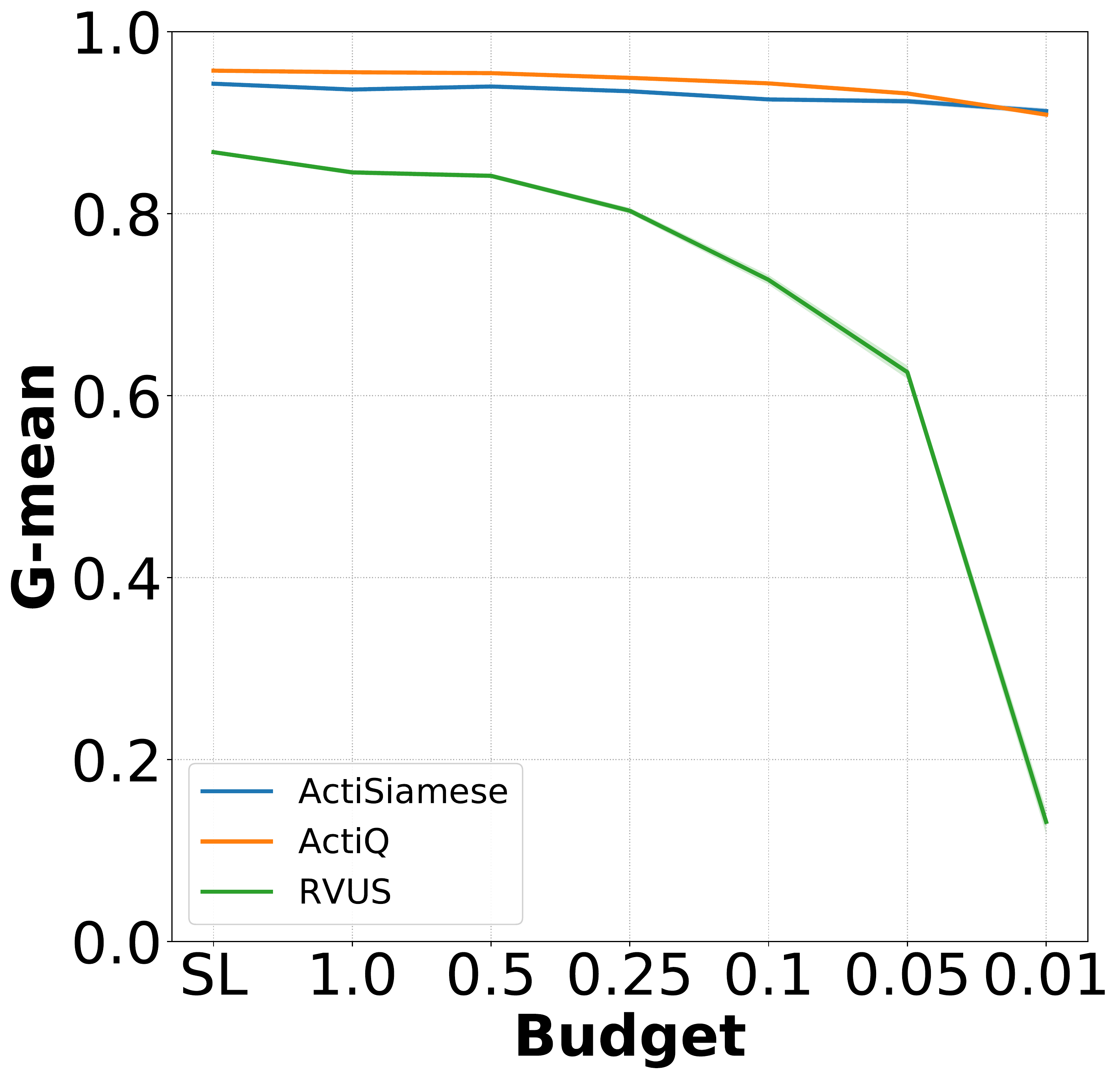}%
		\label{fig:stationary_sea_budget_overall}}
	\subfloat[circles ($t=20000$)]{\includegraphics[scale=0.13]{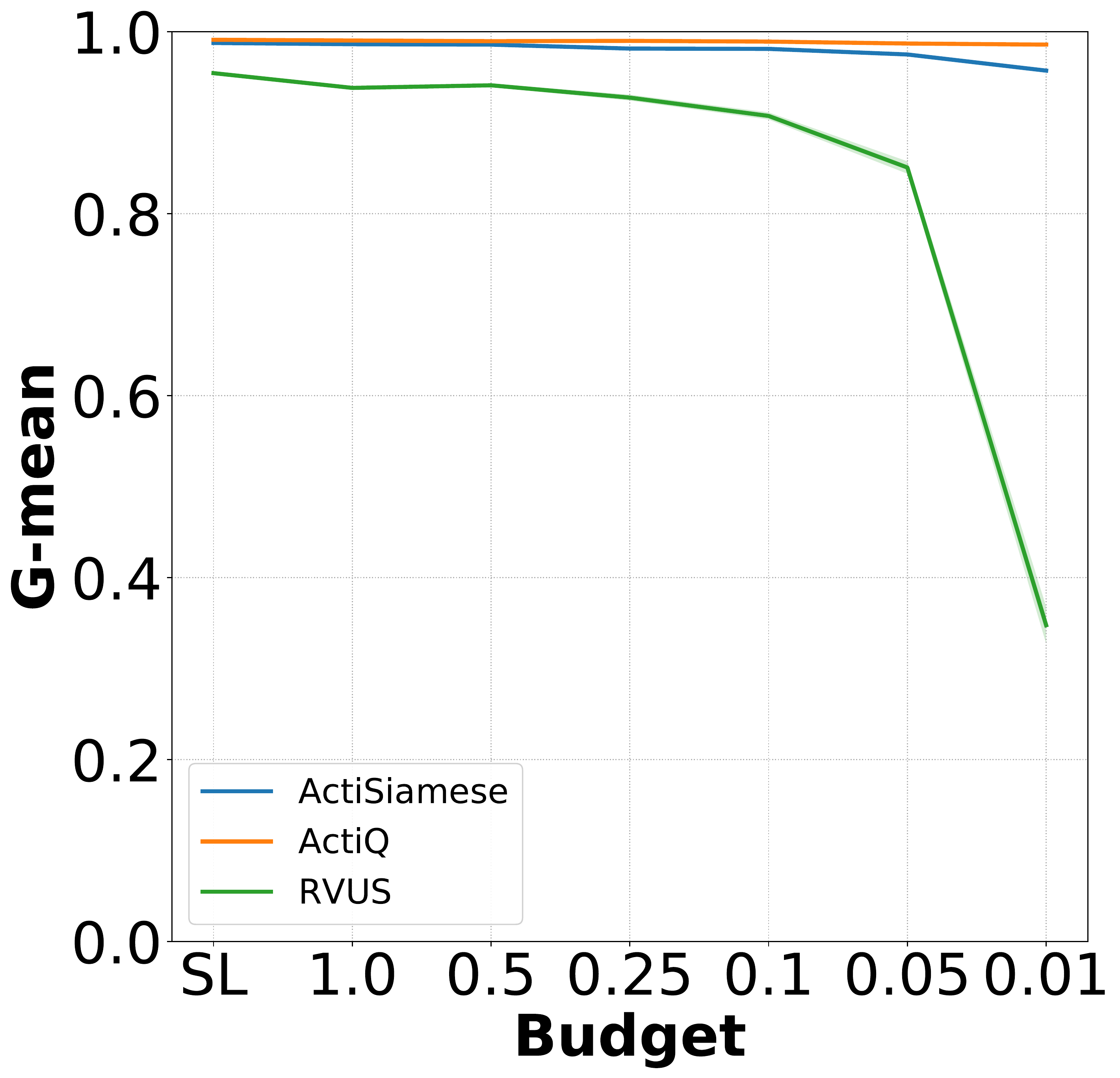}%
		\label{fig:stationary_circles_budget_overall}}
	\subfloat[blobs ($t=20000$)]{\includegraphics[scale=0.13]{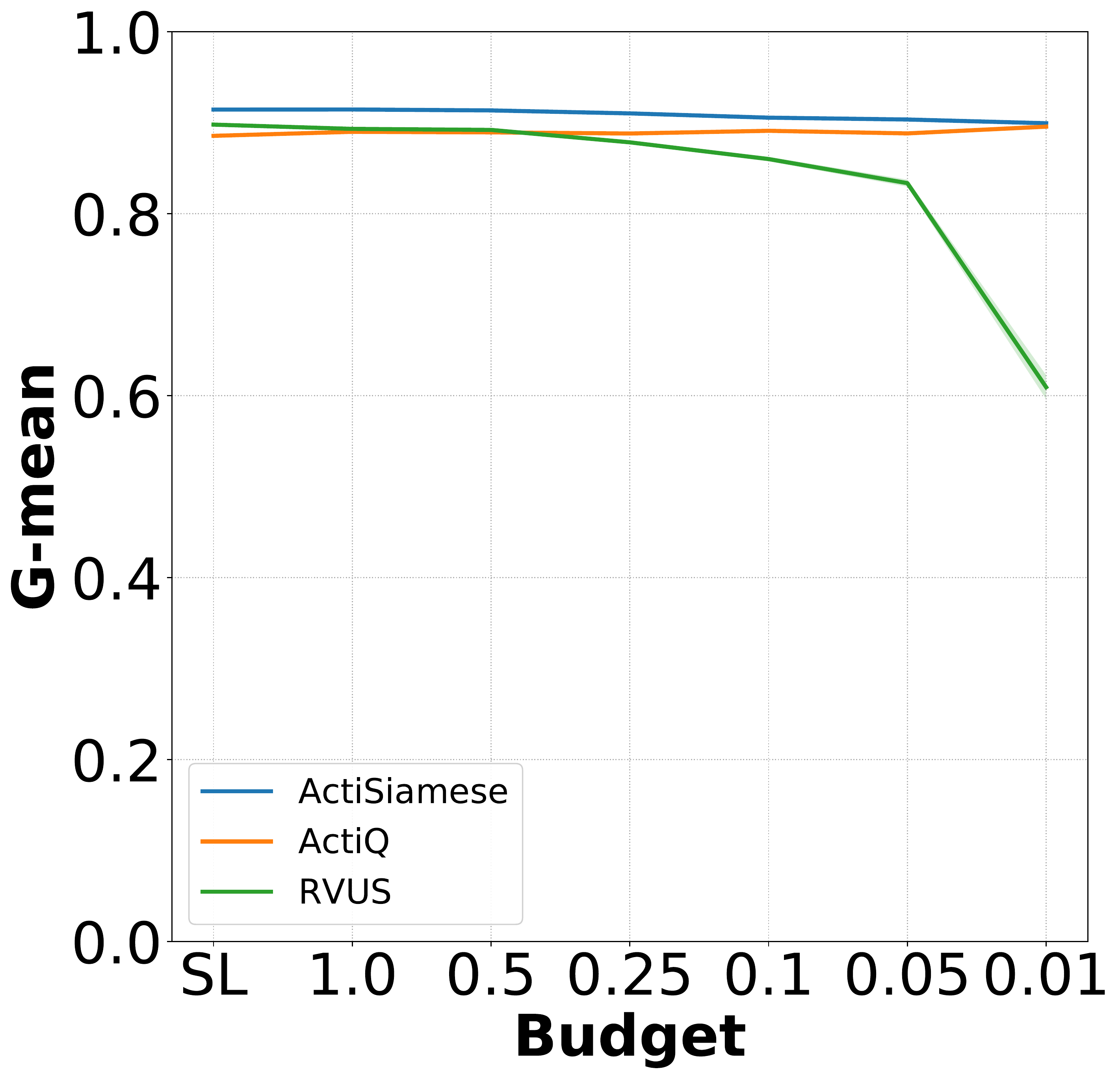}%
		\label{fig:stationary_blobs_budget_overall}}
	
	\subfloat[sea drifted ($t=5500$)]{\includegraphics[scale=0.13]{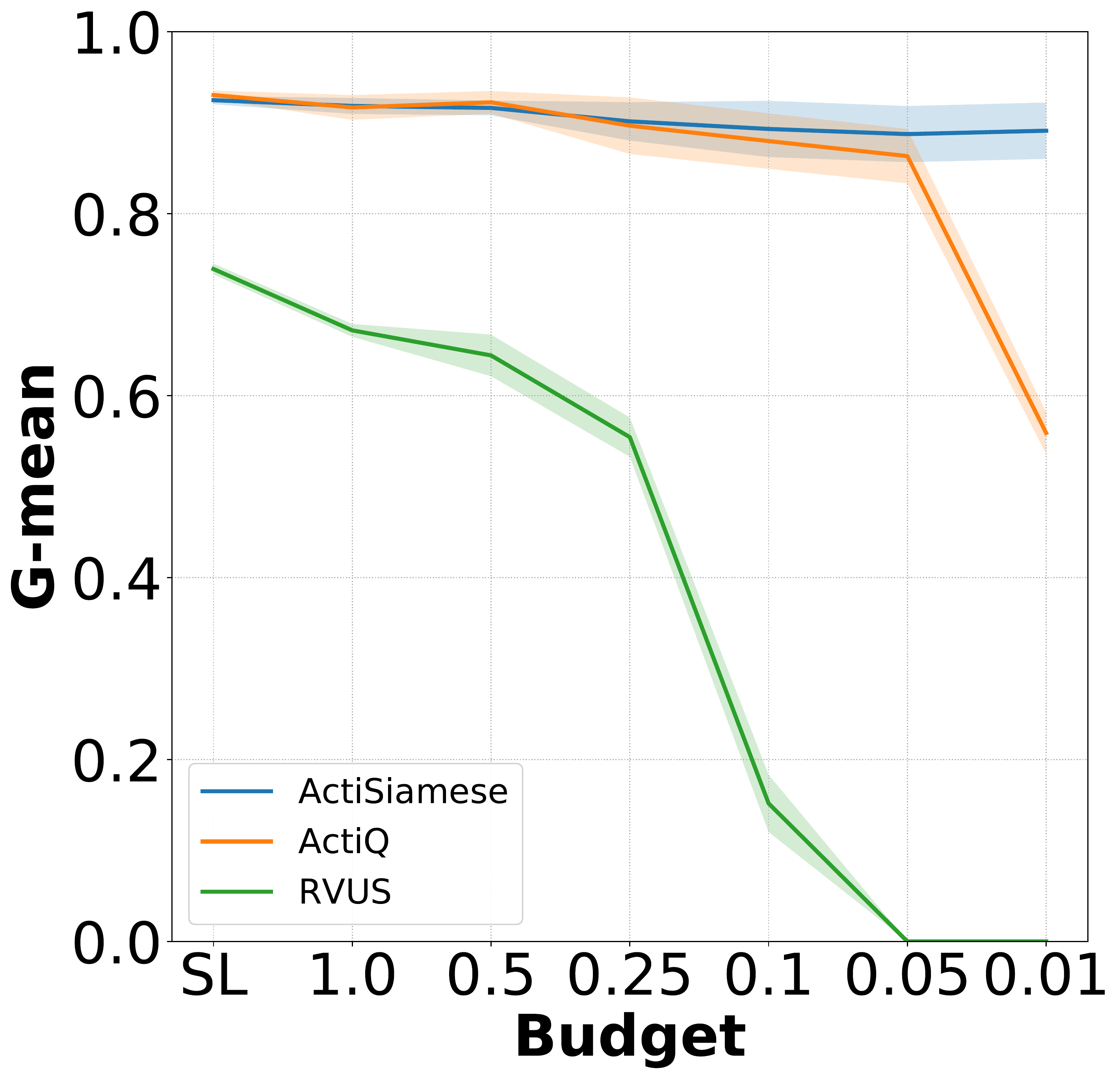}%
		\label{fig:stationary_sea_budget_overall_drift_early}}
	\subfloat[sea drifted ($t=20000$)]{\includegraphics[scale=0.13]{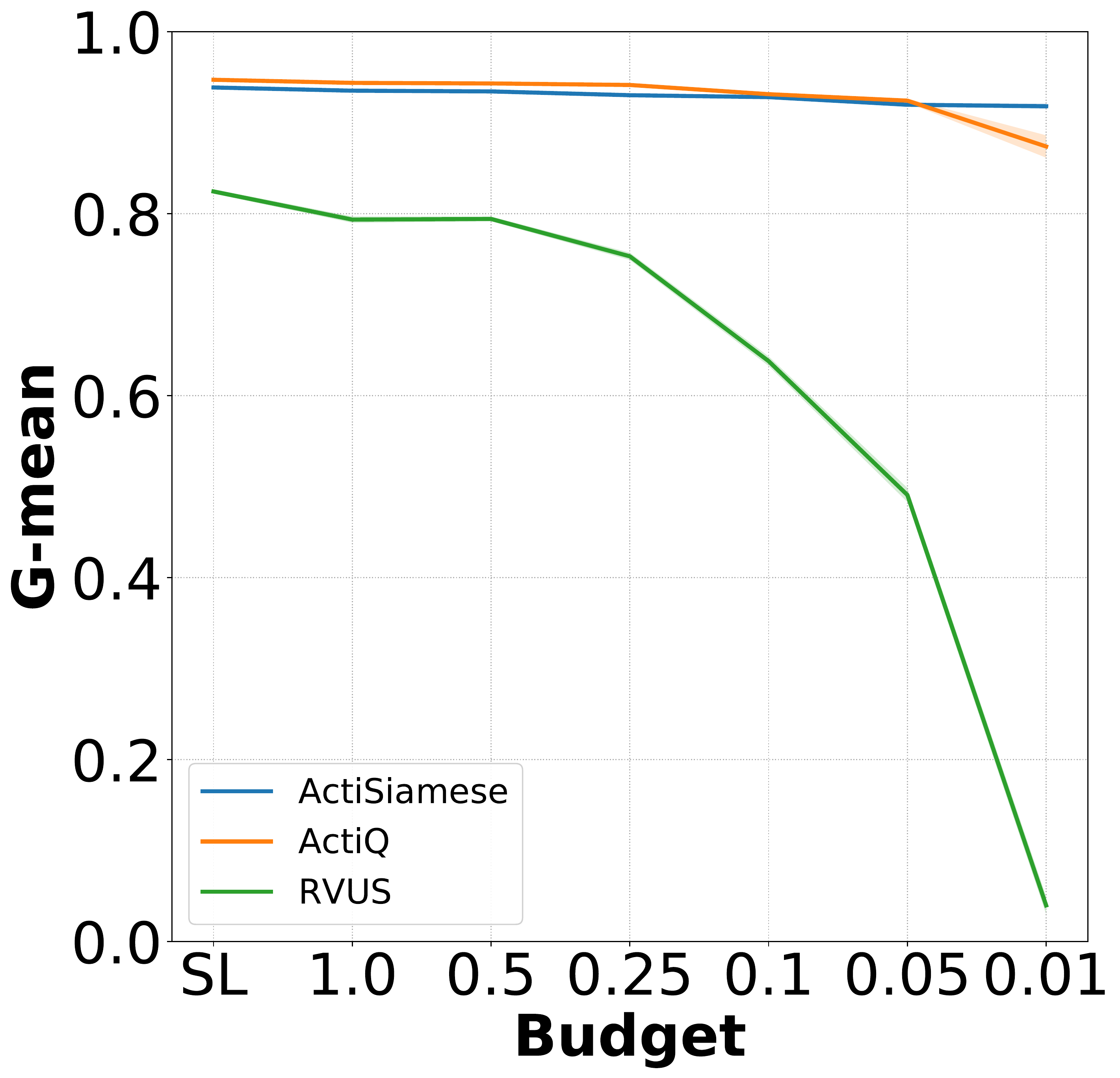}%
		\label{fig:stationary_sea_budget_overall_drift}}
	
	\caption{The effect of the budget $B$ on the performance ($t=20000$) under balanced data.}
\end{figure*}

\textbf{Stationary data}: Figs~\ref{fig:stationary_sea_budget_overall} - \ref{fig:stationary_blobs_budget_overall} depict the results for the three synthetic datasets under stationary conditions. The ``SL'' corresponds to online supervised learning i.e. the label is presented at each step. Simulations were performed for 20000 steps, therefore, the final performance is measured at $t=20000$. The queue capacity is set to $L=10$. RVUS performs significantly worse than the proposed ActiQ and ActiSiamese. The ActiQ and ActiSiamese obtain overall a similar final performance. Specifically, in Fig~\ref{fig:stationary_sea_budget_overall} and \ref{fig:stationary_circles_budget_overall} ActiQ performs slightly better, while in Fig~\ref{fig:stationary_blobs_budget_overall} ActiSiamese performs slightly better. In a later section, however, we will show that their learning speed differs substantially.

\textbf{Nonstationary data}: For \textit{sea}, the same experiment was repeated when drift occurs abruptly at $t=5000$. We measure the performance at $t=5500$, i.e. only 500 steps after the drift, to examine the short-term effect of the drift on performance. This is depicted in Fig~\ref{fig:stationary_sea_budget_overall_drift_early}. The performance of RVUS declines as the budget gets smaller, however, the decline rate is worse compared to Fig~\ref{fig:stationary_sea_budget_overall}. Furthermore, we observe a significant drop in performance for ActiQ when the budget is $B=1\%$. The performance at $t=20000$, that is, 15000 steps after the drift, is shown in Fig~\ref{fig:stationary_sea_budget_overall_drift}. It is almost identical to Fig~\ref{fig:stationary_sea_budget_overall}, i.e., all algorithms have recovered from the drift. In summary, important remarks are as follows:
\begin{itemize}
	\item RVUS' final performance severely declines as the budget gets smaller.
	
	\item For ActiQ and ActiSiamese, the decline rate is very small, even negligible in some cases, with the exception of ActiQ when $B=1\%$ under drift. These approaches are robust to the choice of the budget.
	
	\item We have the first evidence that the multi-queue memory is beneficial compared to the one-pass learner RVUS.
\end{itemize}

\begin{figure}[t!]
	\centering
	
	\subfloat[sea]{\includegraphics[scale=0.15]{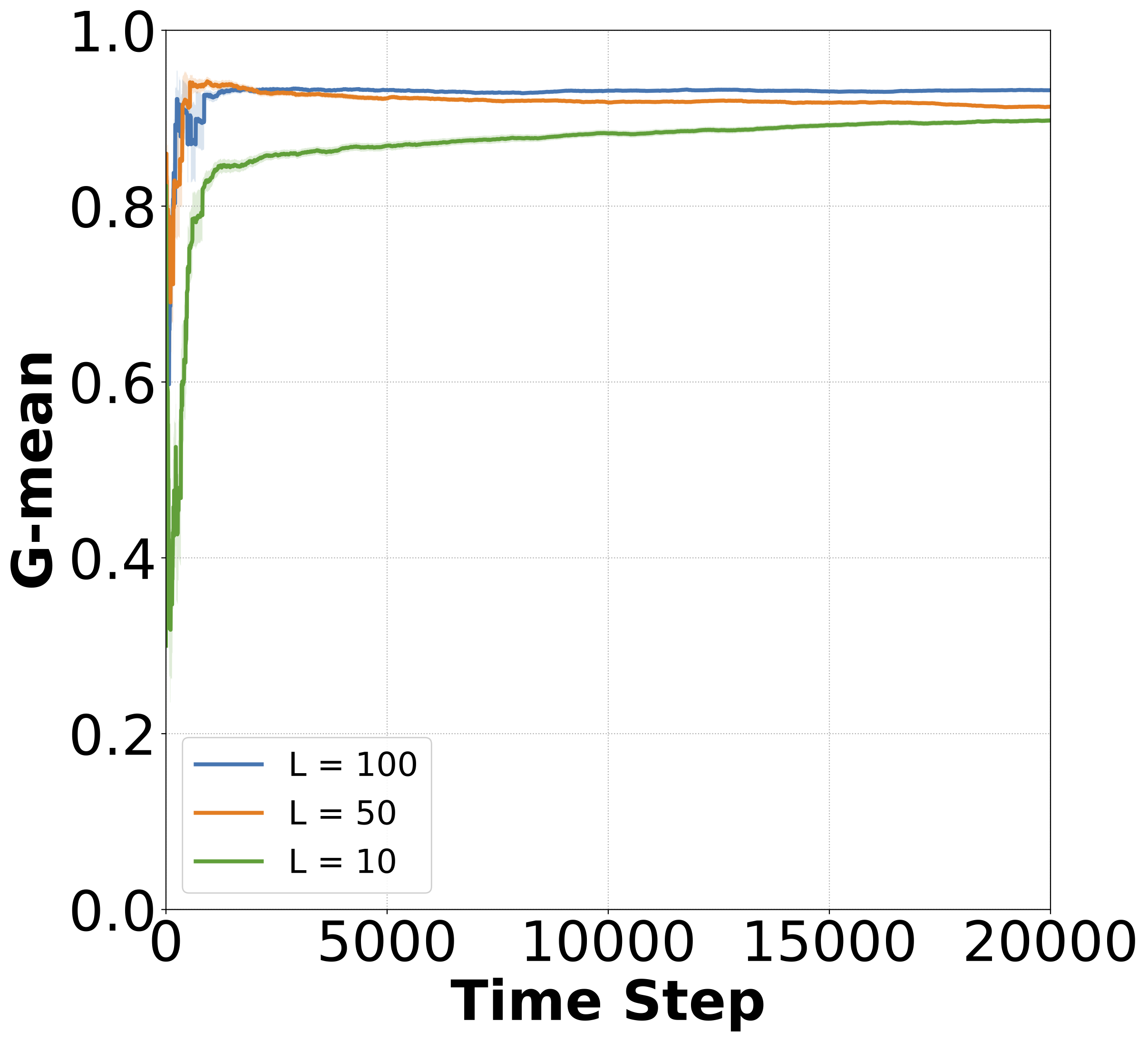}%
		\label{fig:role_sea10_actisiamese_imbalanced}}
	\subfloat[circles]{\includegraphics[scale=0.15]{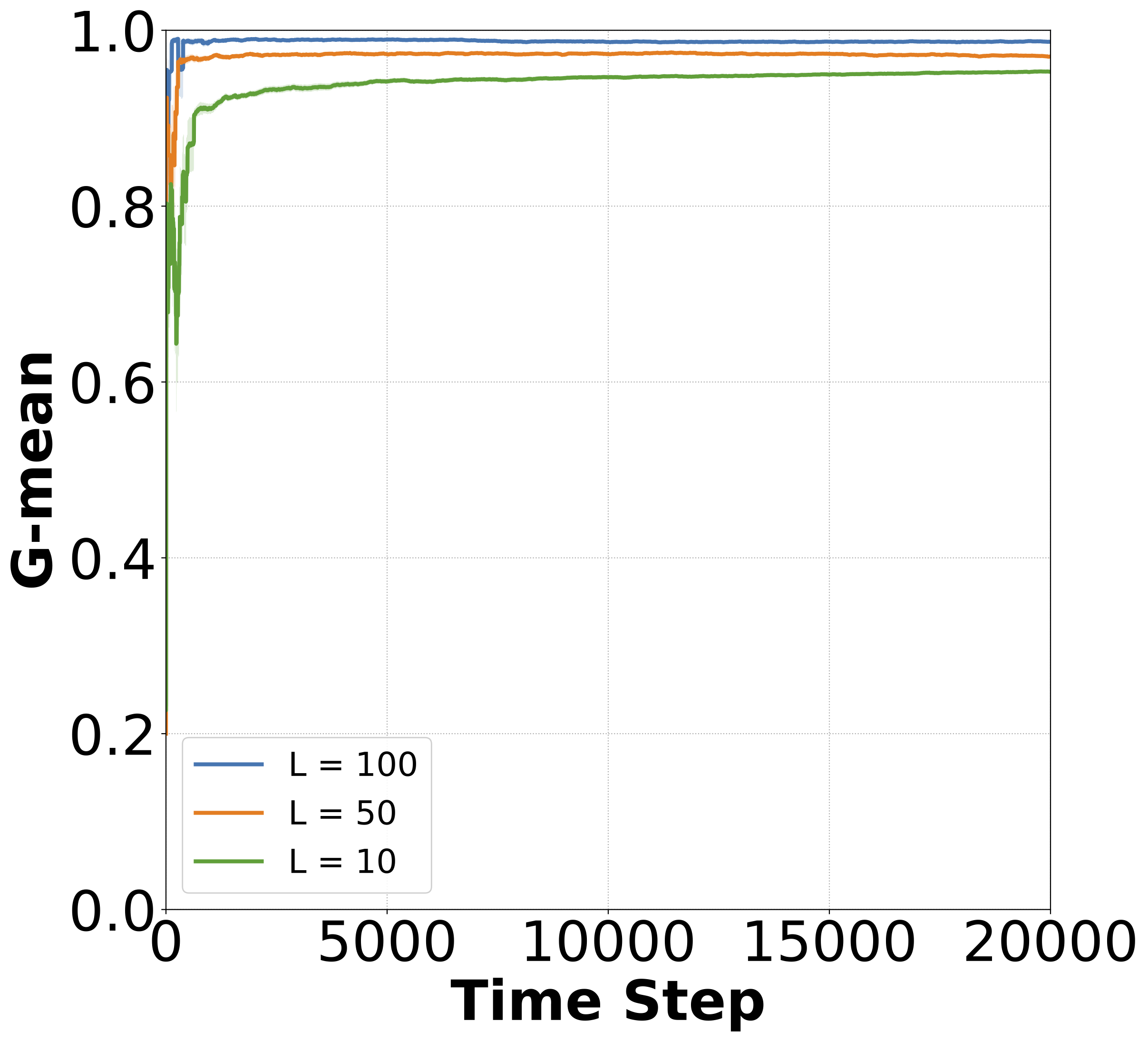}%
		\label{fig:role_circles10_actisiamese_imbalanced}}
	
	\subfloat[sea drifted]{\includegraphics[scale=0.15]{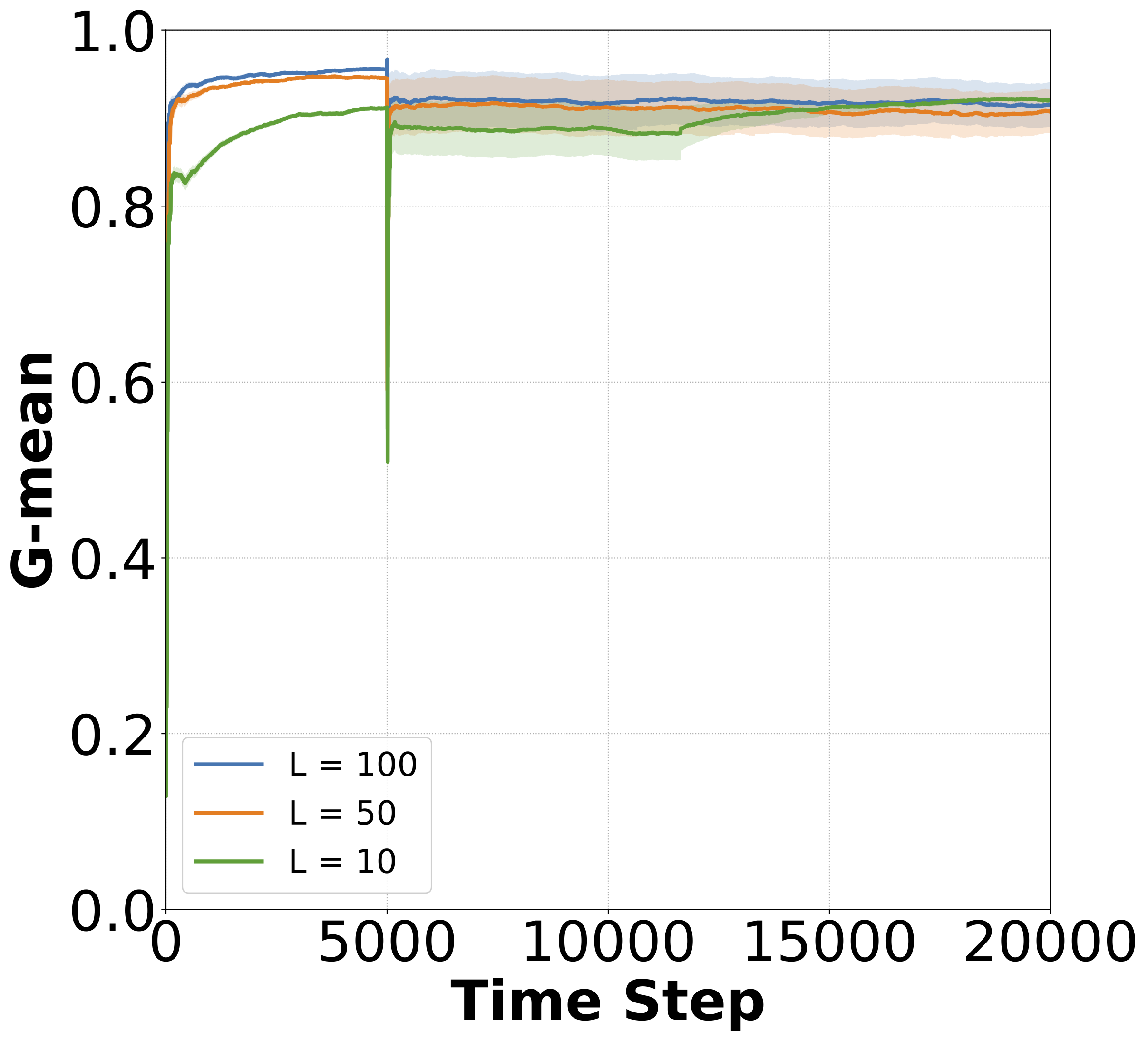}%
		\label{fig:role_sea10_actisiamese_abrupt}}
	\subfloat[circles drifted]{\includegraphics[scale=0.15]{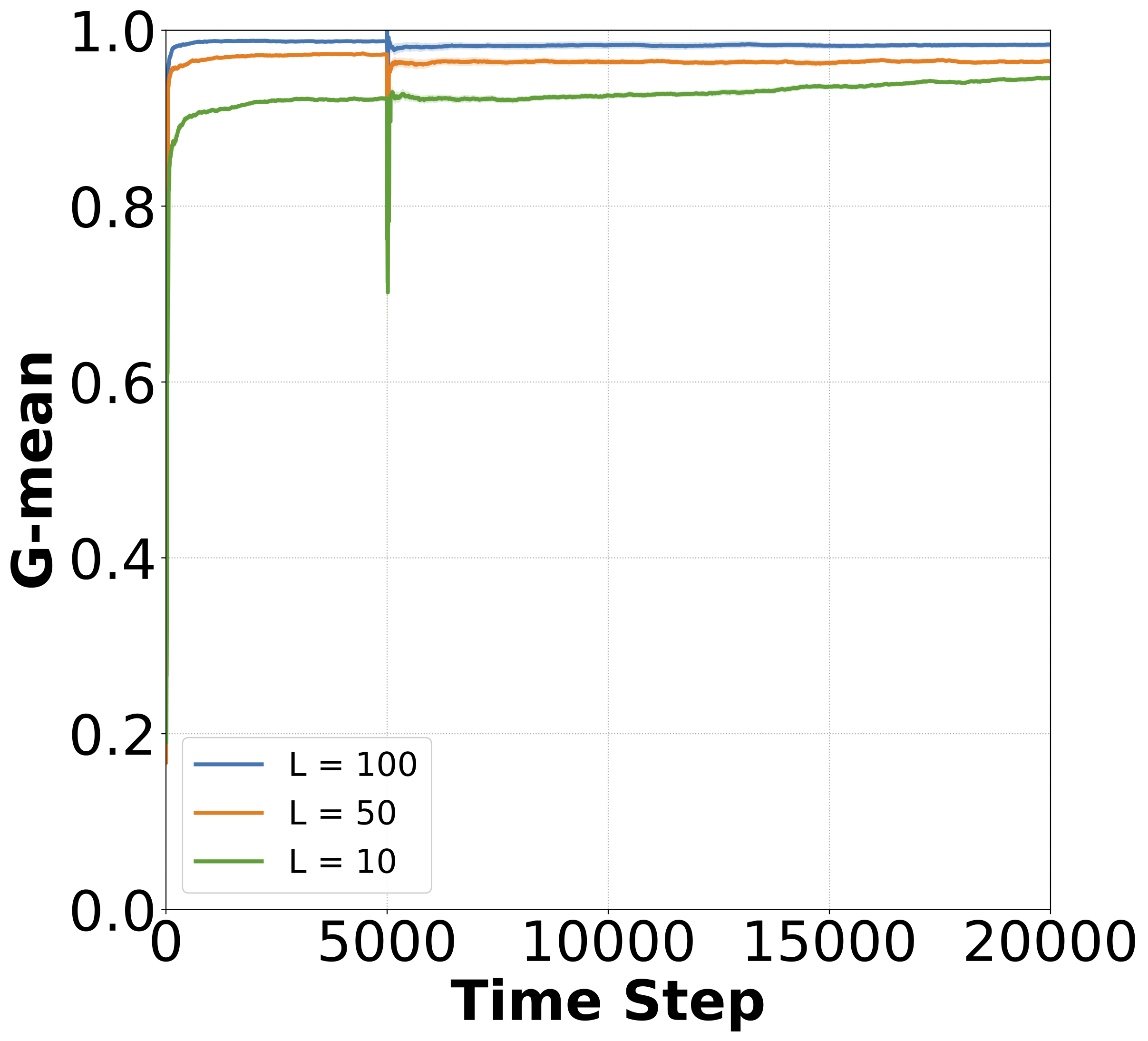}%
		\label{fig:role_circles10_actisiamese_abrupt}}
	
	\caption{The effect of $L$ on ActiSiamese under severe ($1\%$) imbalance.}
\end{figure}

\subsection{Role of the queue capacity \textit{L}}
\begin{figure}[t!]
	\centering
	
	\subfloat[sea]{\includegraphics[scale=0.15]{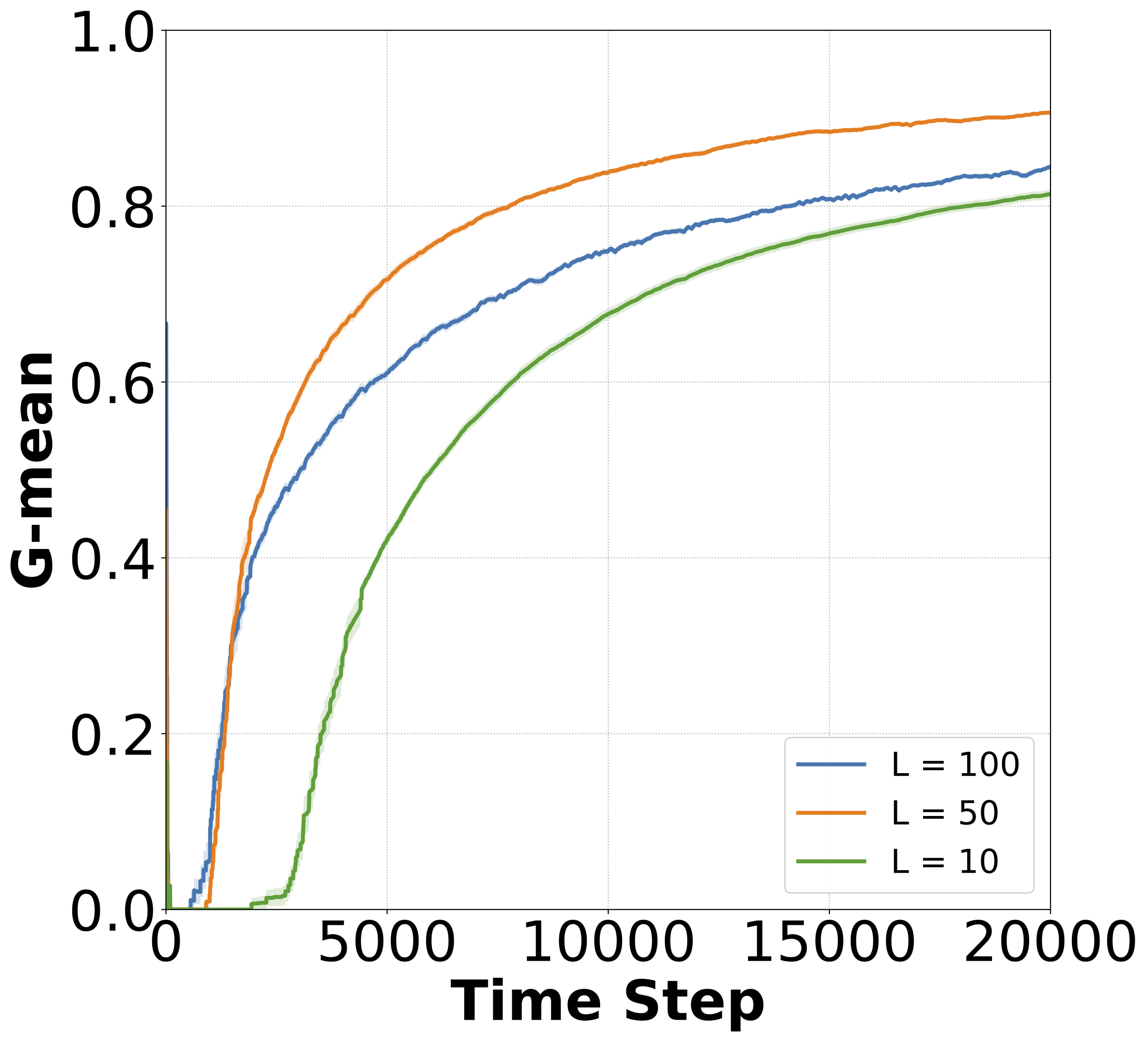}%
		\label{fig:role_sea10_actiq_imbalanced}}
	\subfloat[circles]{\includegraphics[scale=0.15]{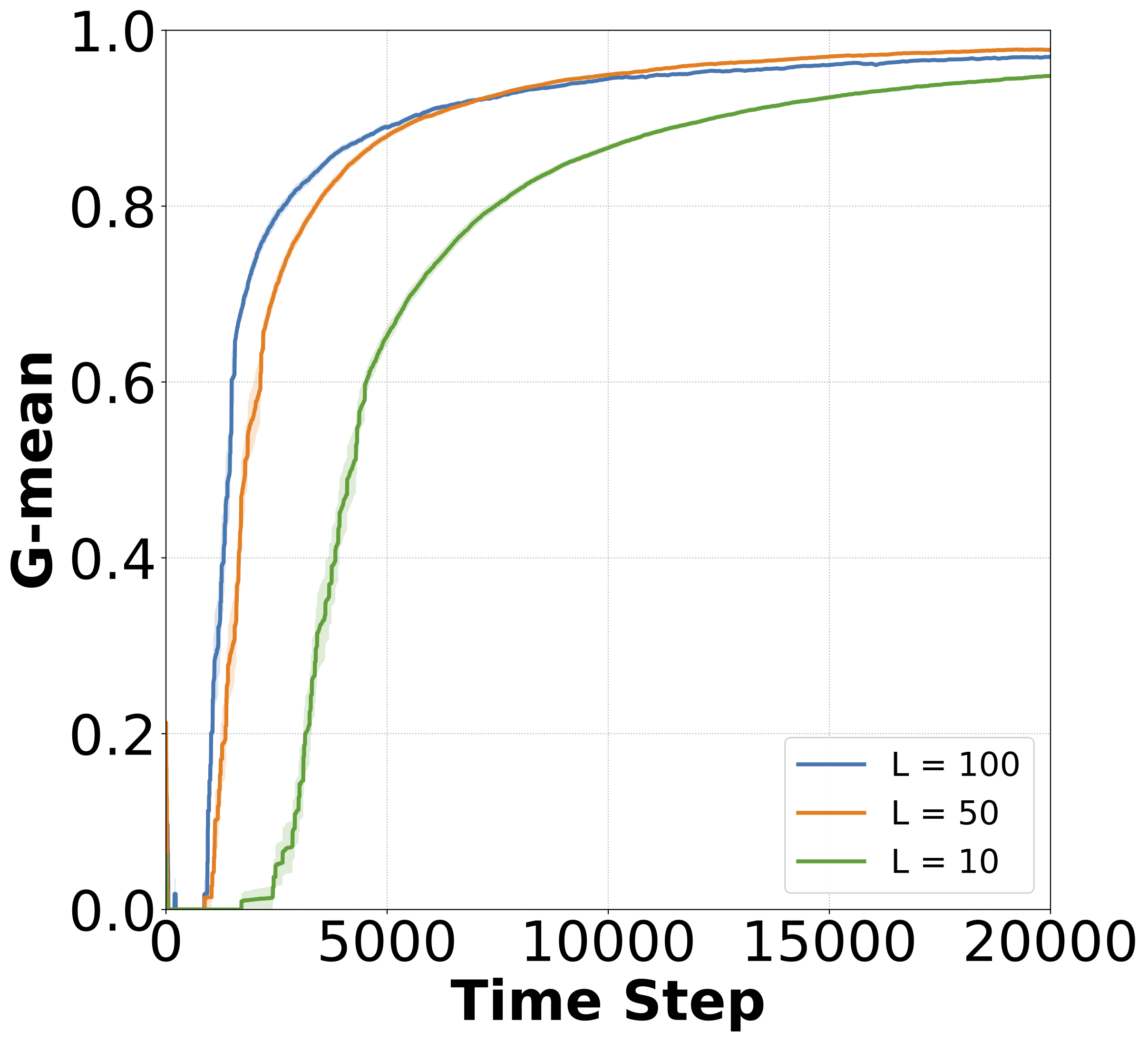}%
		\label{fig:role_circles10_actiq_imbalanced}}
	
	\subfloat[sea drifted]{\includegraphics[scale=0.15]{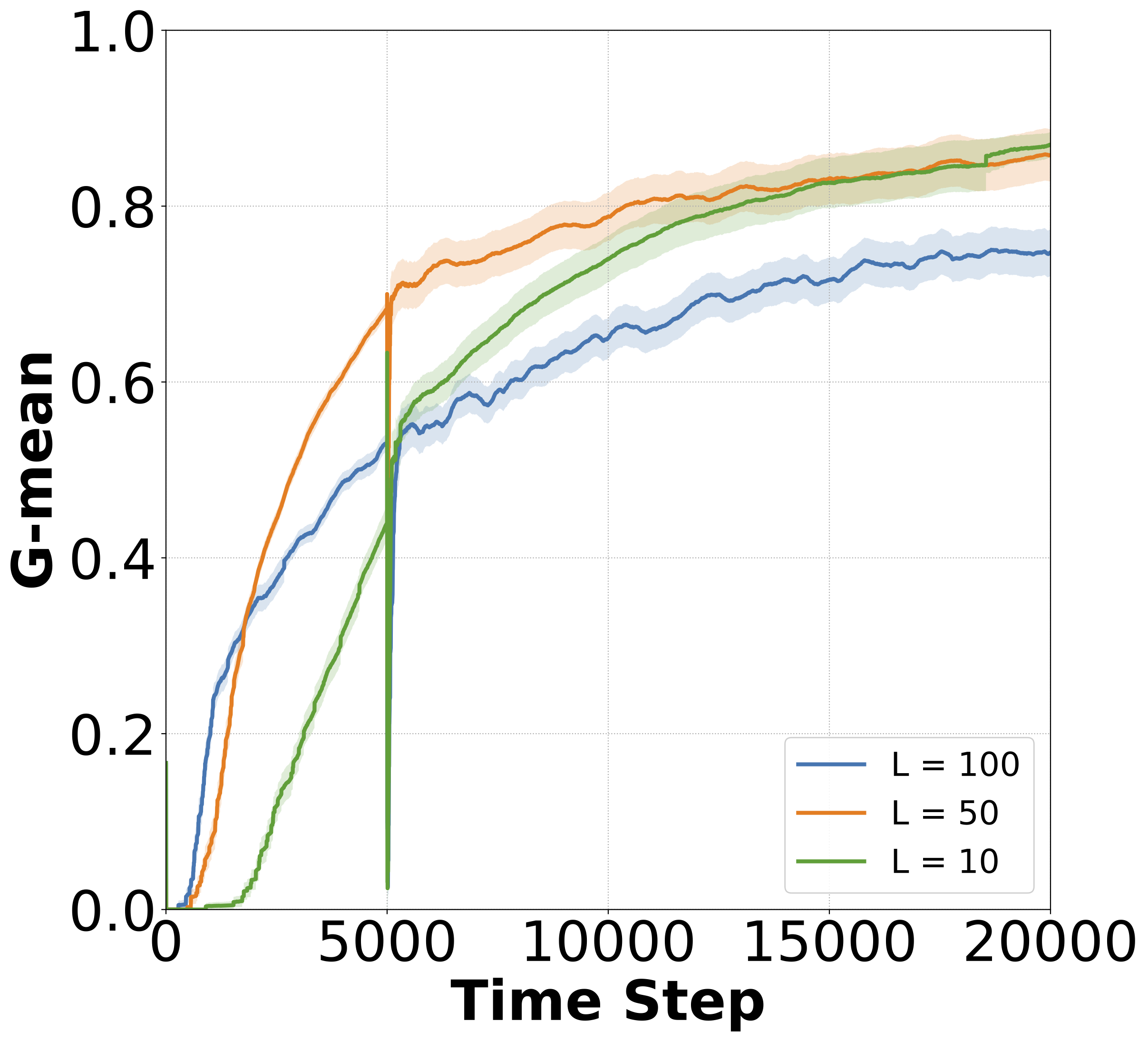}%
		\label{fig:role_sea10_actiq_abrupt}}
	\subfloat[circles drifted]{\includegraphics[scale=0.15]{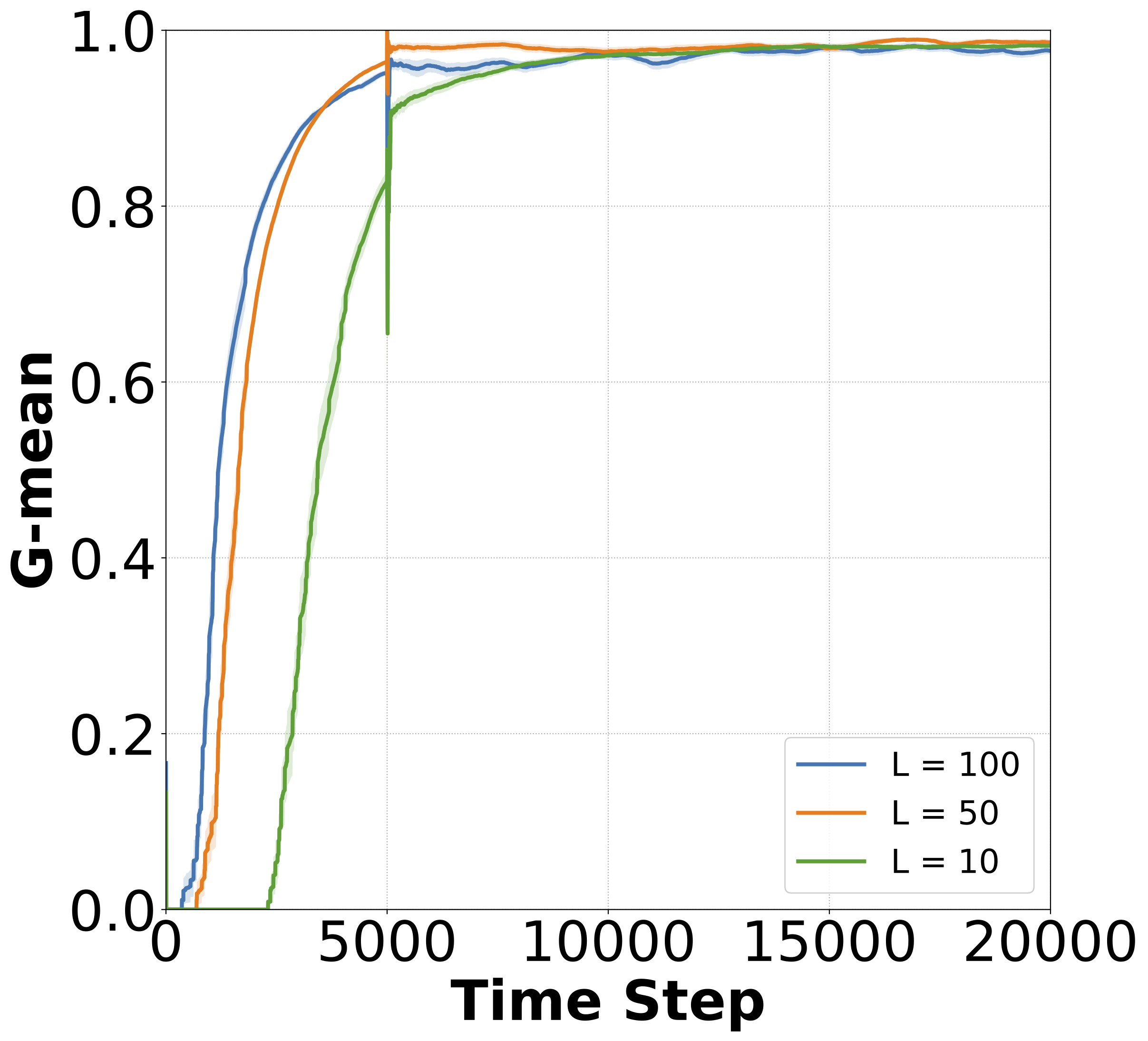}%
		\label{fig:role_circles10_actiq_abrupt}}
	
	\caption{The effect of $L$ on ActiQ under severe ($1\%$) imbalance.}
\end{figure}

\textbf{Stationary data}: Figs~\ref{fig:role_sea10_actisiamese_imbalanced}~-~\ref{fig:role_circles10_actisiamese_imbalanced} show the learning curves for ActiSiamese in \textit{sea} and \textit{circles} respectively. The curves show the prequential G-mean at every step. For all experiments the budget is $B=1\%$ and imbalance is severe ($1\%$). We conclude that the higher the value of $L$ the higher the performance, however, diminishing returns are obtained, e.g., the cases where $L=50$ and $L=100$ do not differ considerably. Importantly, irrespective of the value of $L$, ActiSiamese learns fast (all learning curves are steep). This is in contrast to ActiQ shown in Figs~\ref{fig:role_sea10_actiq_imbalanced}~-~\ref{fig:role_circles10_actiq_imbalanced} where the choice of $L$ significantly affects its learning speed.

\textbf{Nonstationary data}: Figs~\ref{fig:role_sea10_actisiamese_abrupt}~-~\ref{fig:role_circles10_actisiamese_abrupt} show the learning curves for ActiSiamese in \textit{sea} and \textit{circles} respectively where drift occurs abruptly at $t=5000$. Hence, let's focus on the part of the curves after the drift i.e. $t \geq 5000$. In Fig~\ref{fig:role_sea10_actisiamese_abrupt} where the drift is more severe, we observe a variation in performance (indicated by the shaded regions) and a slight decrease in the final performance compared to Fig~\ref{fig:role_sea10_actisiamese_imbalanced}. ActiSiamese is robust to the choice of $L$; for comparison, Figs~\ref{fig:role_sea10_actiq_abrupt}~-~\ref{fig:role_circles10_actiq_abrupt} show the analogous plots for ActiQ. To sum up, important remarks are:
\begin{itemize}
	\item In stationary data, ActiSiamese with larger values of $L$ performs better, however, diminishing returns are obtained.
	
	\item In nonstationary data, ActiSiamese is still robust to the choice of $L$, however, a slight performance variation is observed after the drift.
	
	\item We have the first evidence that the use of a Siamese network is beneficial compared to ActiQ, which uses a standard neural network.
\end{itemize}

\section{Comparative Study}\label{sec:comparative_study}

\subsection{Synthetic datasets}
\begin{figure*}[t!]
	\centering
	
	\subfloat[normal]{\includegraphics[scale=0.13]{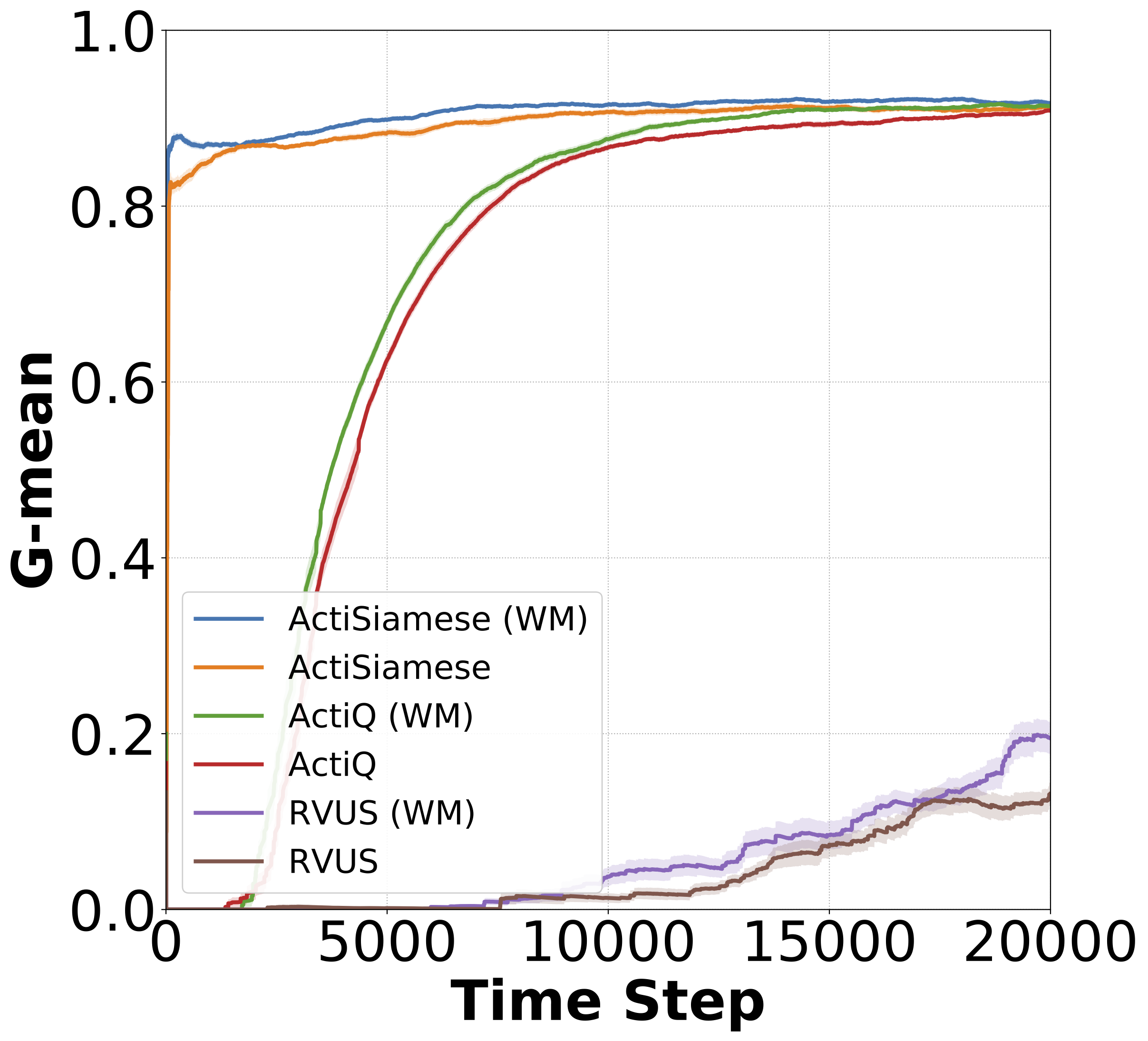}%
		\label{fig:stationary_sea10_budget_001}}
	\subfloat[class imbalance]{\includegraphics[scale=0.13]{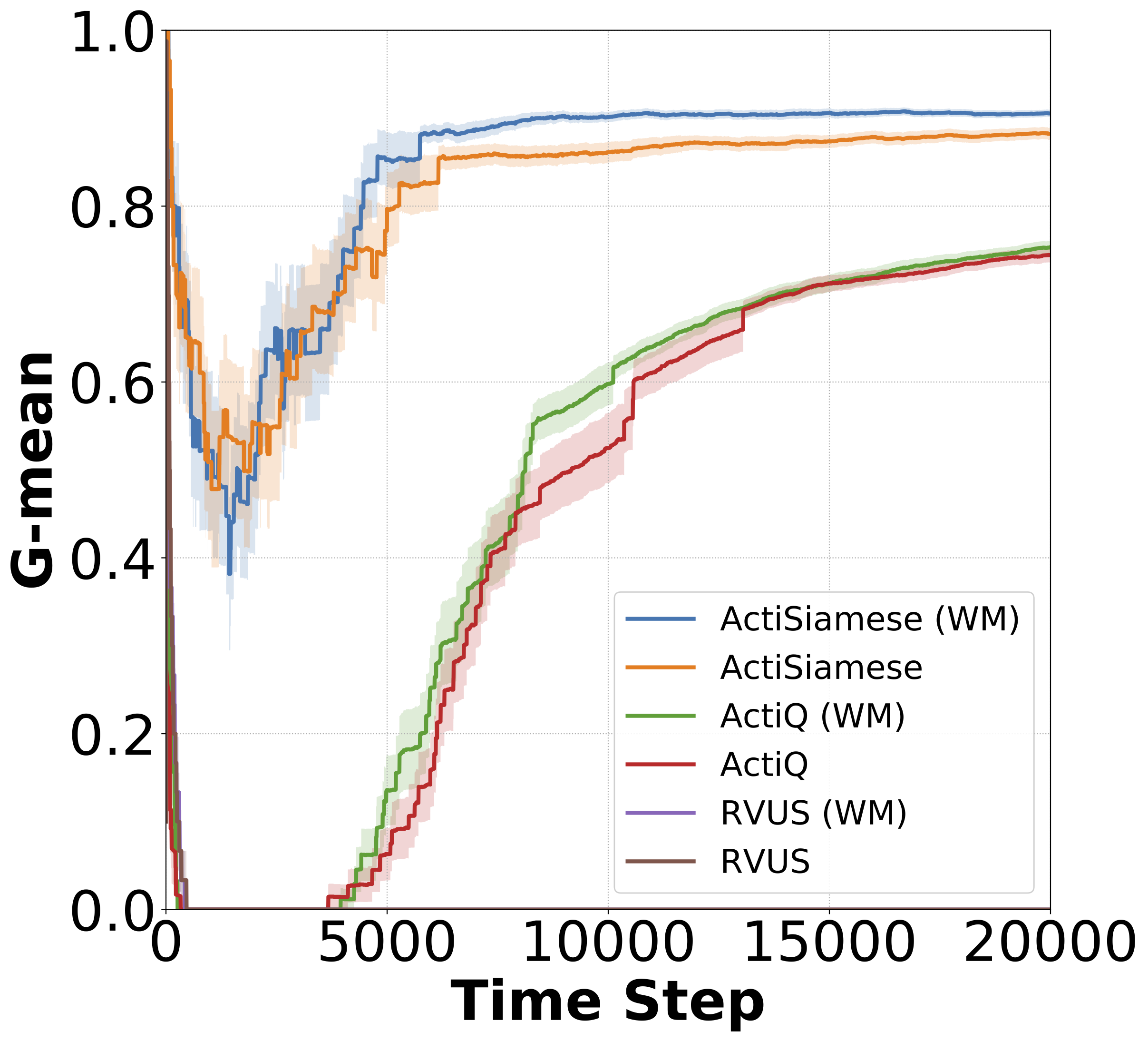}%
		\label{fig:imbalance_sea10_mm_extreme}}
	\subfloat[abrupt drift ]{\includegraphics[scale=0.13]{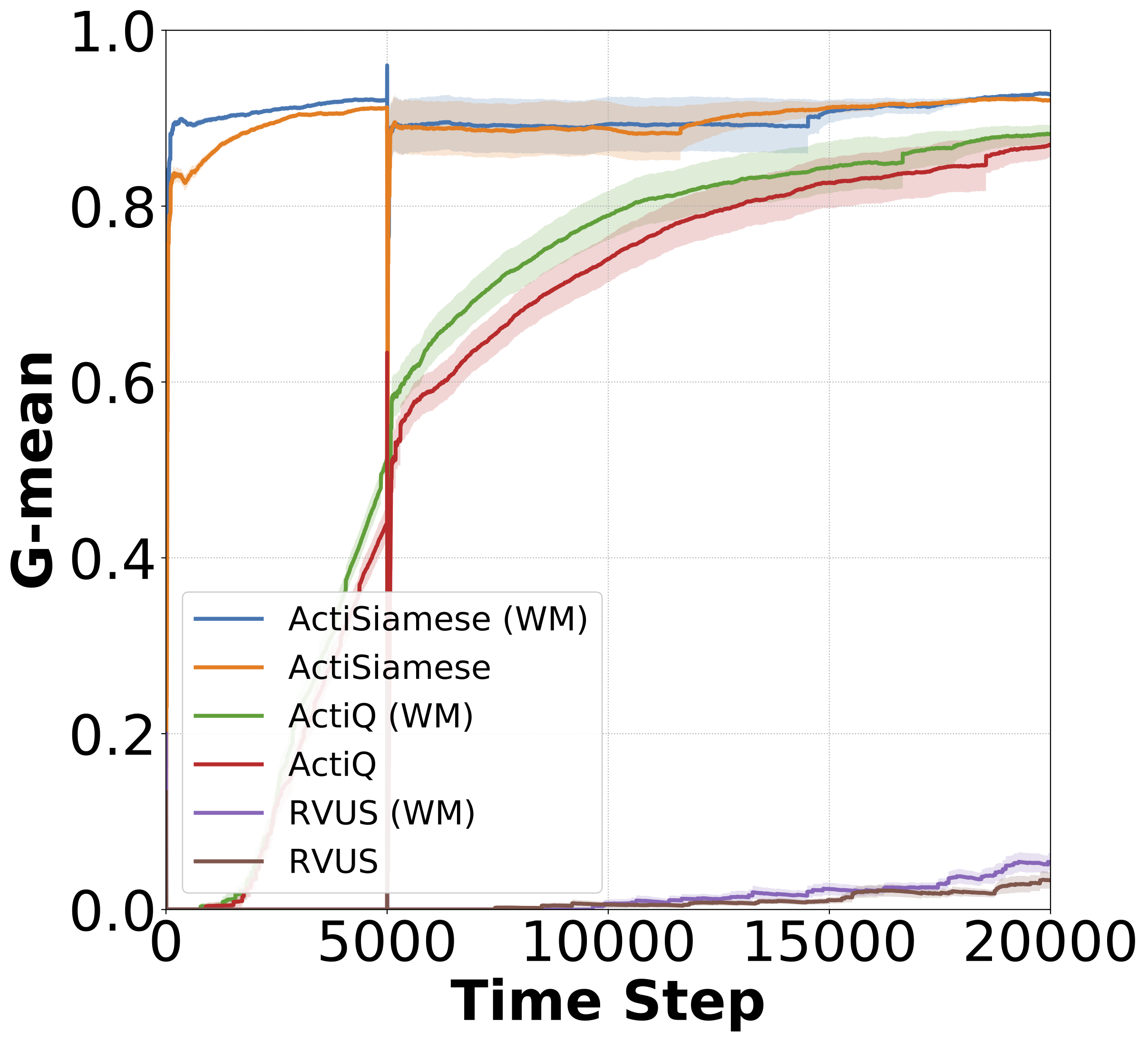}%
		\label{fig:drift_sea10_budget_001}}
	
	\subfloat[imbalance + abrupt ]{\includegraphics[scale=0.13]{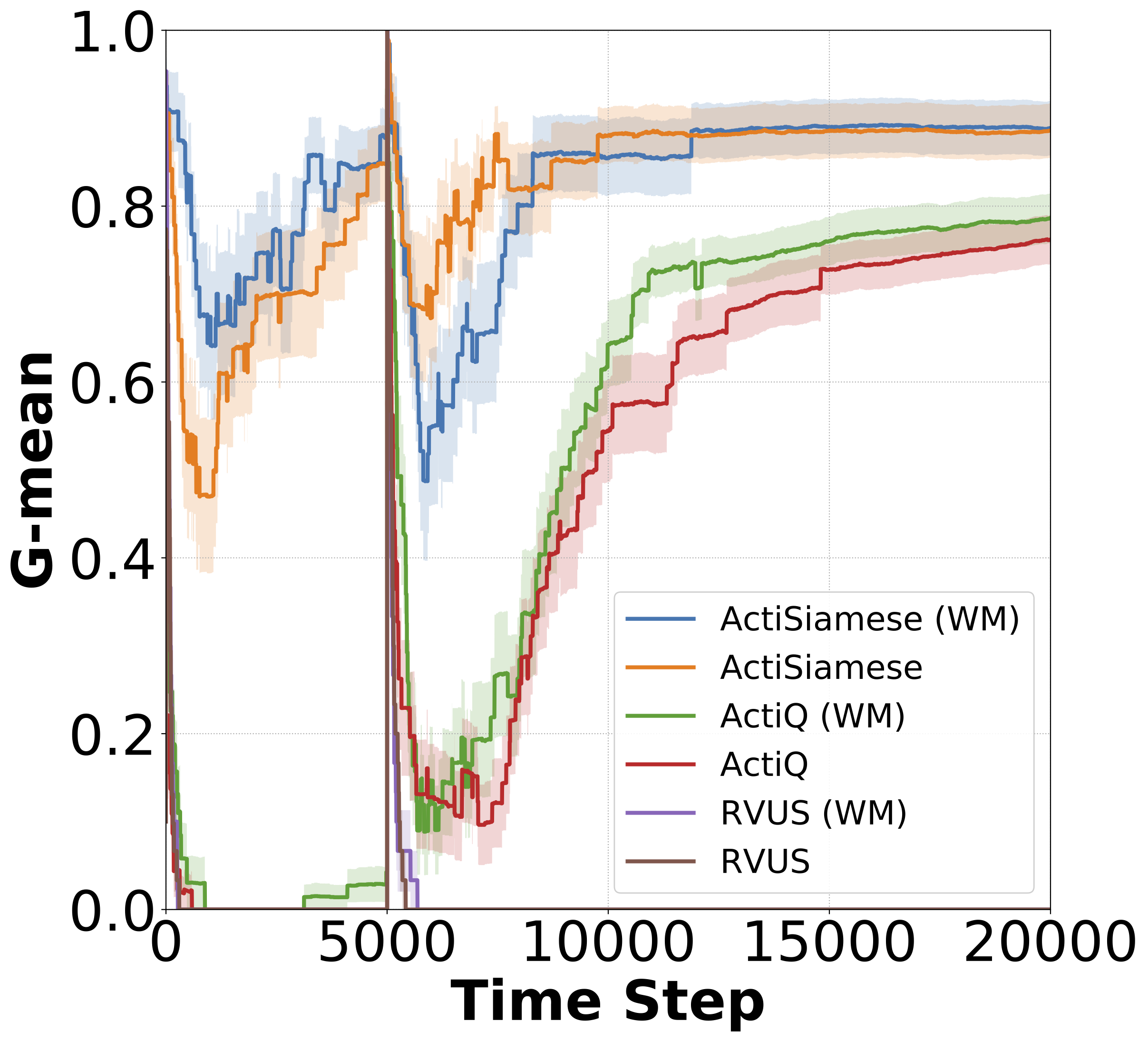}%
		\label{fig:drift_imbalance_sea10_mm_extreme}}
	\subfloat[recurrent drift ]{\includegraphics[scale=0.13]{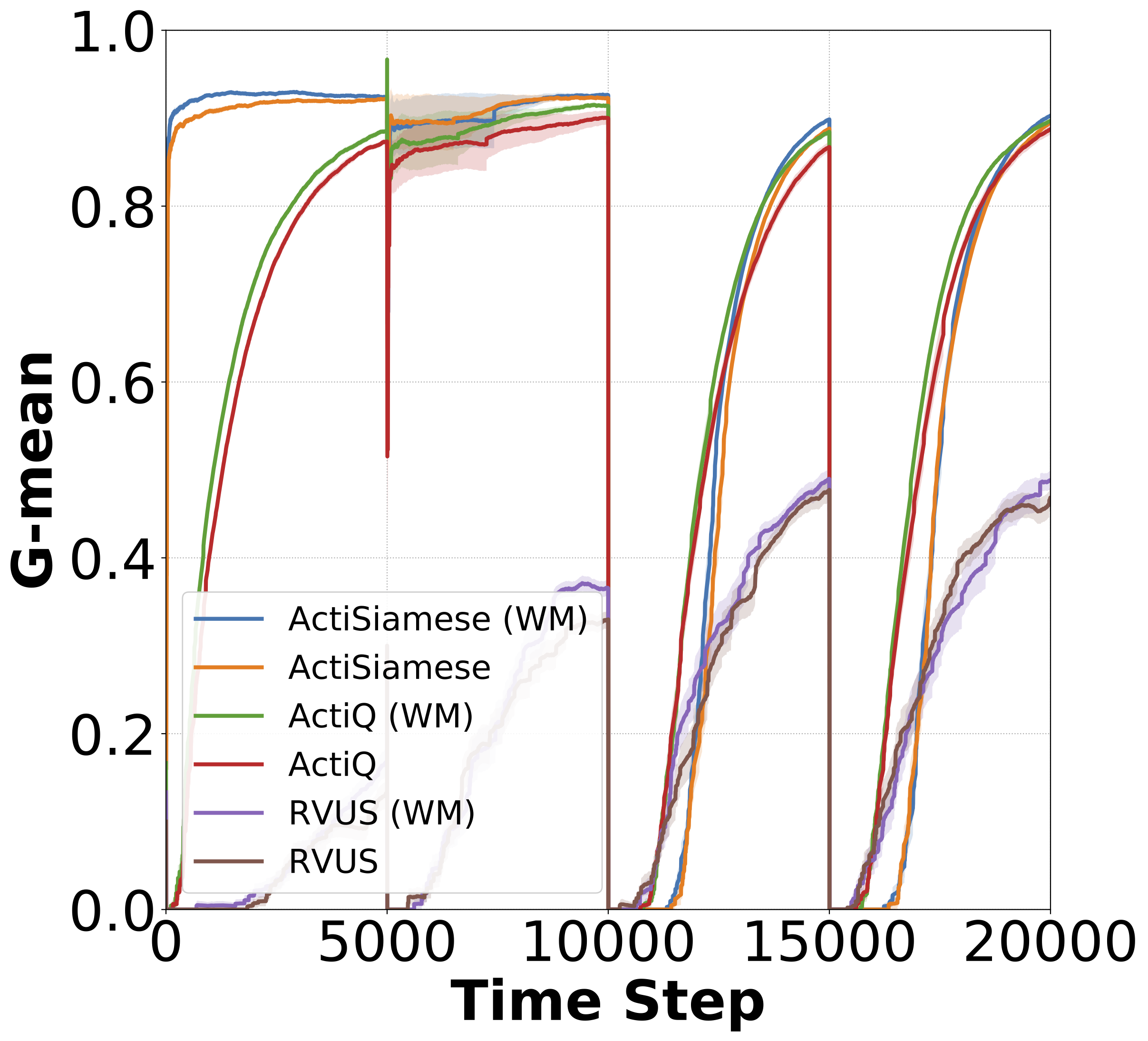}%
		\label{fig:drift_rec_sea10_budget_005}}
	
	\caption{Comparative study in sea: (a) normal (balanced, stationary), (b) extreme ($0.1\%$) imbalance, (c) abrupt drift, (d) extreme imbalance and abrupt drift, and (e) recurrent drift.}
\end{figure*}

Figs.~\ref{fig:stationary_sea10_budget_001}-\ref{fig:drift_rec_sea10_budget_005} show the performance of the methods in the sea dataset under normal conditions (stationary, balanced), extreme ($0.1\%$) imbalance, abrupt drift, both extreme imbalance with abrupt drift, and recurrent drift respectively. In all experiments, the memory size is $L=10$, and the budget is $B=1\%$ unless otherwise stated.

In Fig.~\ref{fig:stationary_sea10_budget_001}, ActiSiamese learns significantly faster. Given more time, ActiQ will catch up. In Fig.~\ref{fig:imbalance_sea10_mm_extreme}, the superiority of ActiSiamese is shown. In Fig.~\ref{fig:drift_sea10_budget_001} both ActiSiamese and ActiQ (to a lesser degree) deal well with abrupt drift; notice how similar this figure with Fig.~\ref{fig:stationary_sea10_budget_001} is. The combination of drift with imbalance in Fig.~\ref{fig:drift_imbalance_sea10_mm_extreme} causes some variation in the performance throughout the experiment's duration. We have run another simulation experiment with recurrent drift shown in Fig.~\ref{fig:drift_rec_sea10_budget_005}. Given the challenging conditions of recurrent drift, the active learning budget has been increased from $B=1\%$ to $B=5\%$.

\begin{figure*}[t]
	\centering
	
	\subfloat[normal]{\includegraphics[scale=0.13]{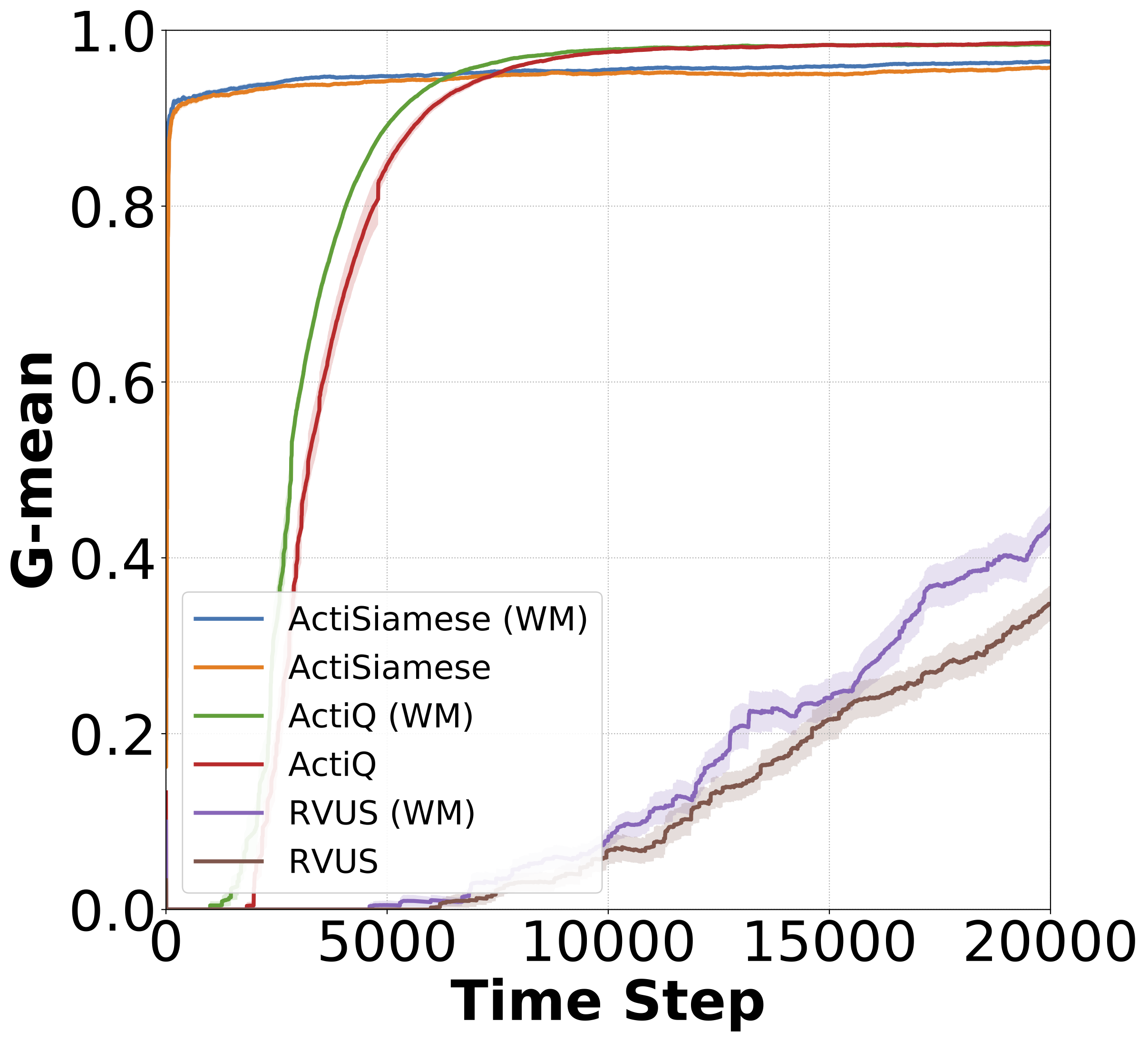}%
		\label{fig:stationary_circles10_budget_001}}
	\subfloat[class imbalance]{\includegraphics[scale=0.13]{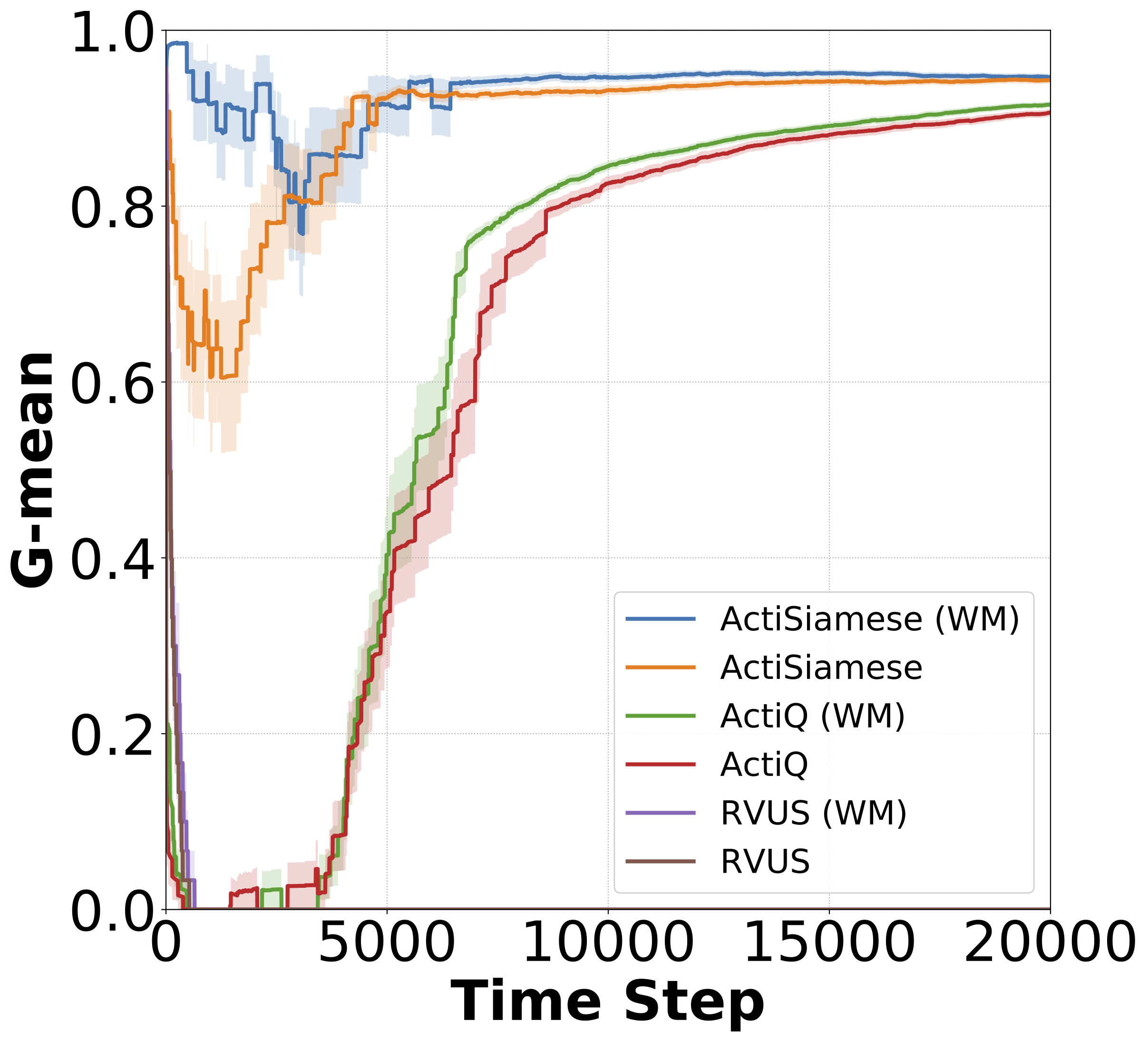}%
		\label{fig:imbalance_circles10_mm_extreme}}
	\subfloat[abrupt drift]{\includegraphics[scale=0.13]{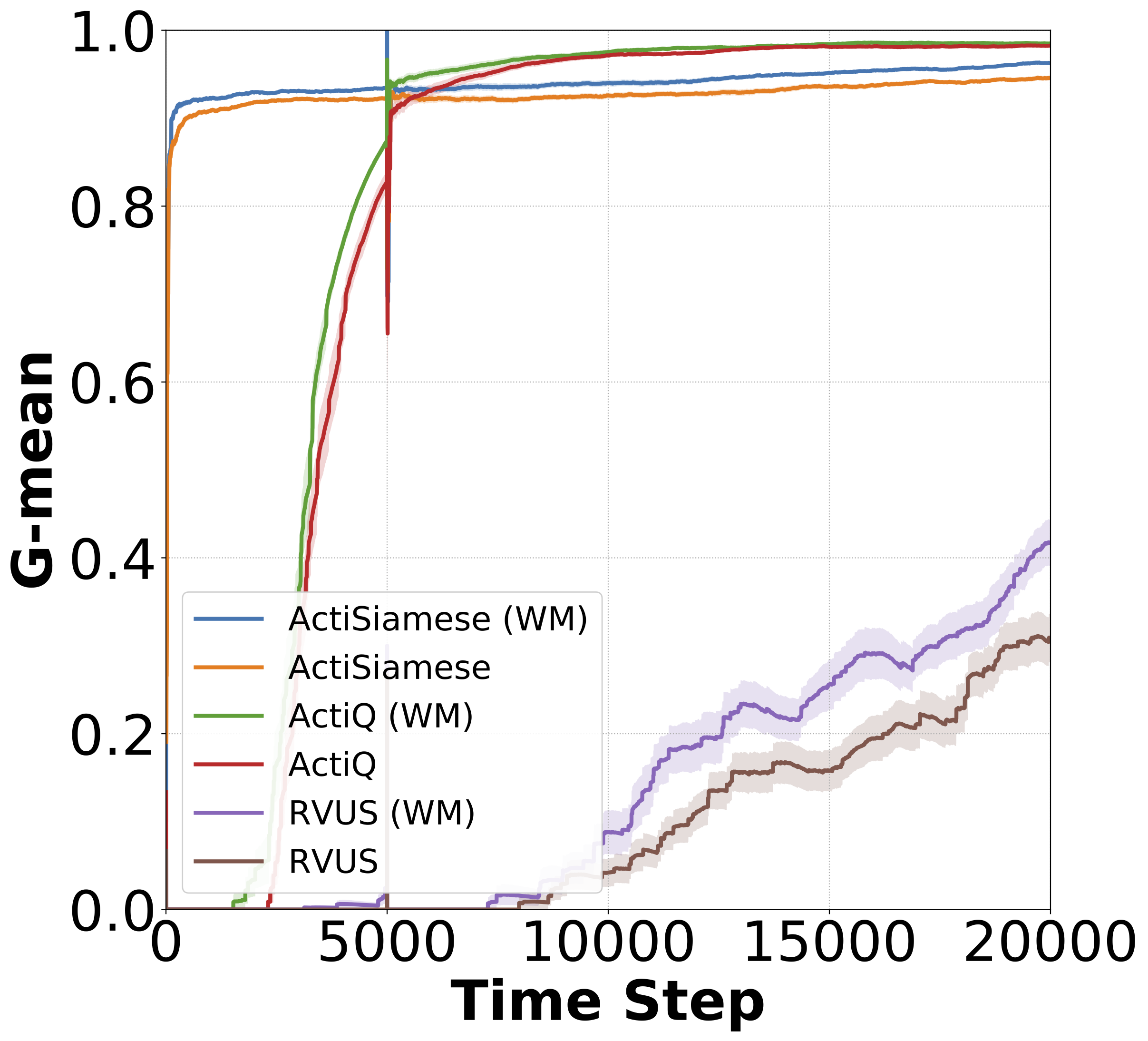}%
		\label{fig:drift_circles10_budget_001}}
	
	\subfloat[imbalance + abrupt]{\includegraphics[scale=0.13]{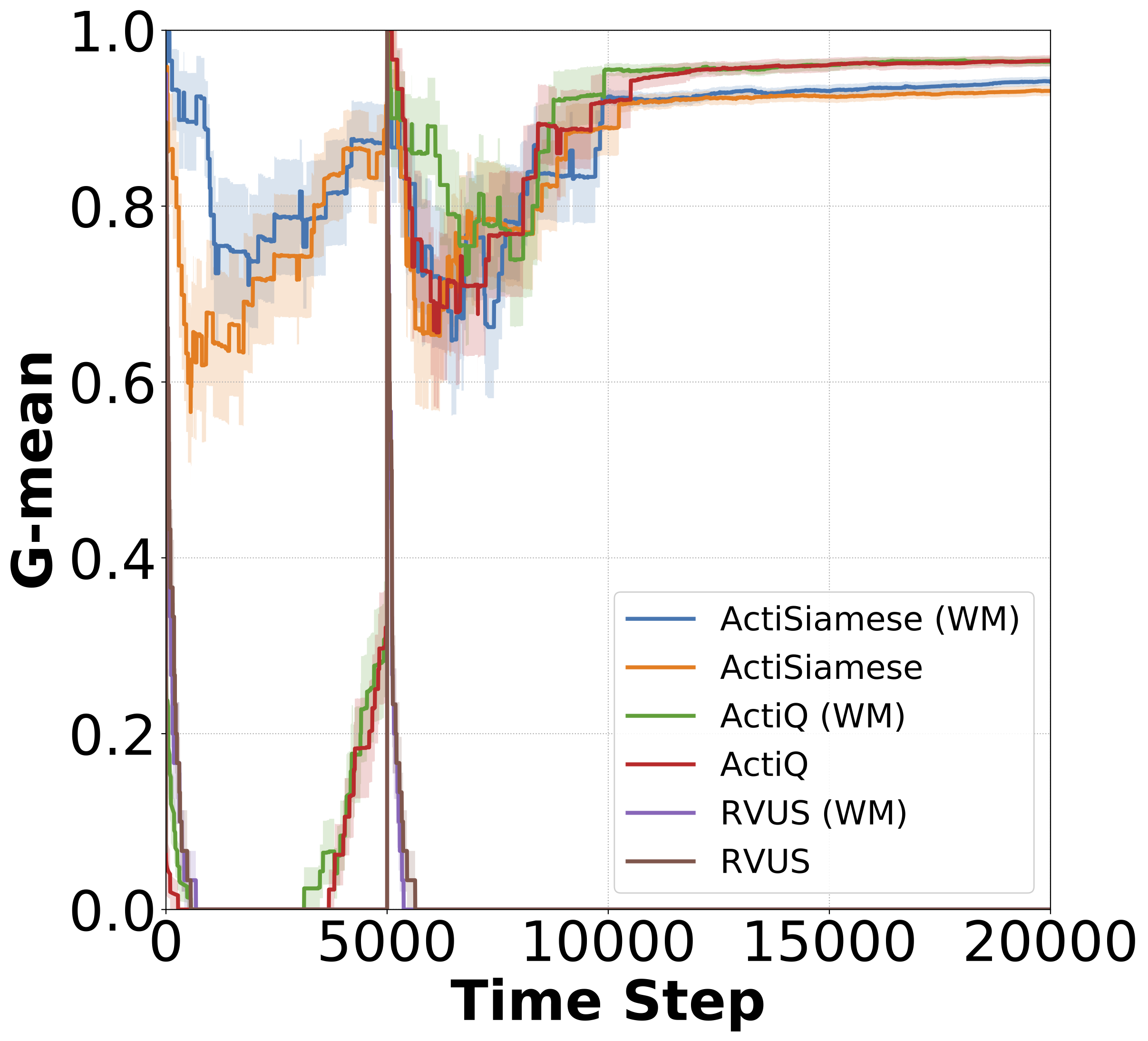}%
		\label{fig:drift_imbalance_circles10_mm_extreme}}
	\subfloat[recurrent drift]{\includegraphics[scale=0.13]{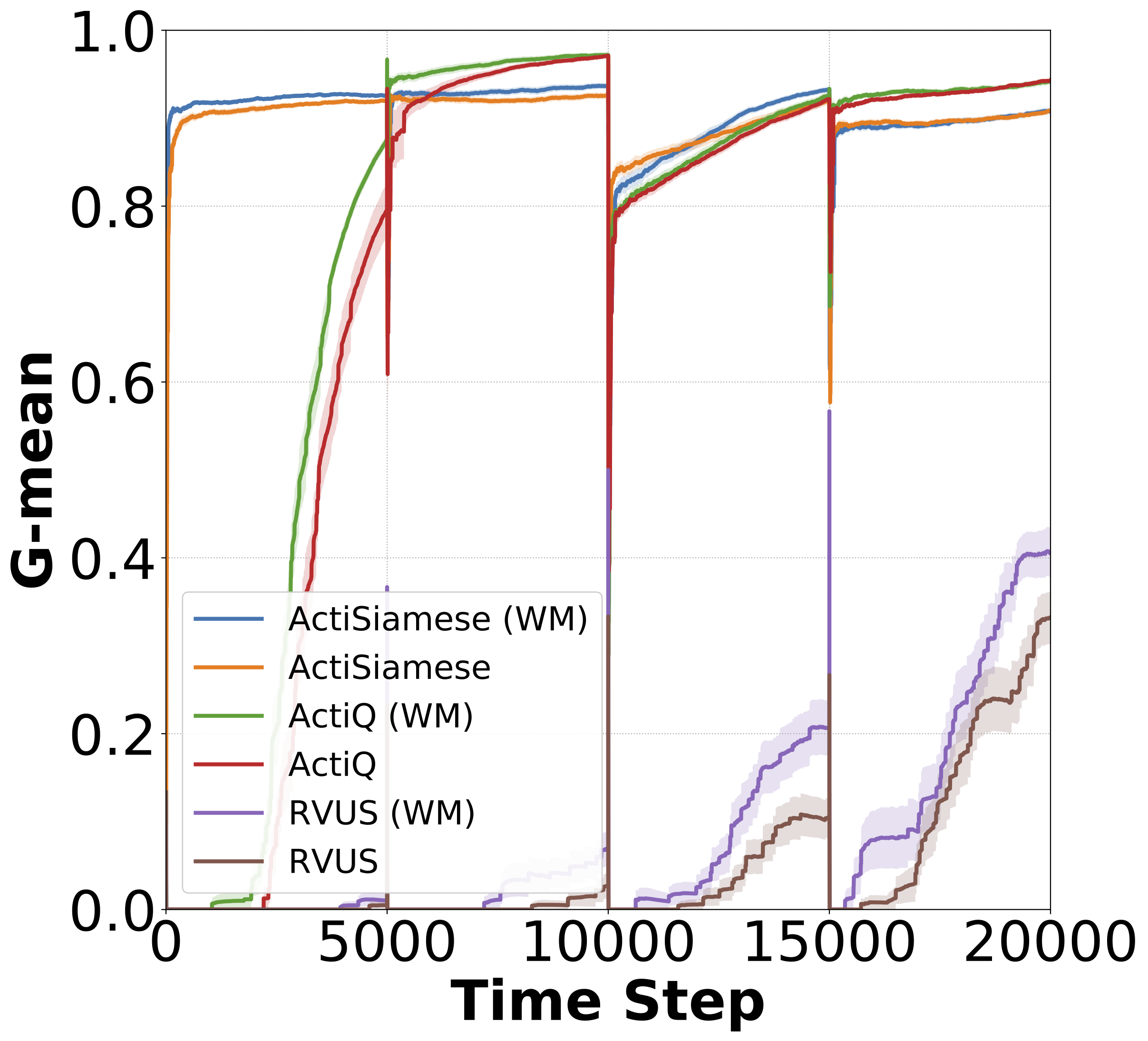}%
		\label{fig:drift_rec_circles10_budget_001}}
	
	\caption{Comparative study in circles: (a) normal (balanced, stationary), (b) extreme ($0.1\%$) imbalance, (c) abrupt drift, (d) extreme imbalance and abrupt drift, and (e) recurrent drift.}
\end{figure*}

\begin{figure*}[t!]
	\centering
	
	\subfloat[normal]{\includegraphics[scale=0.13]{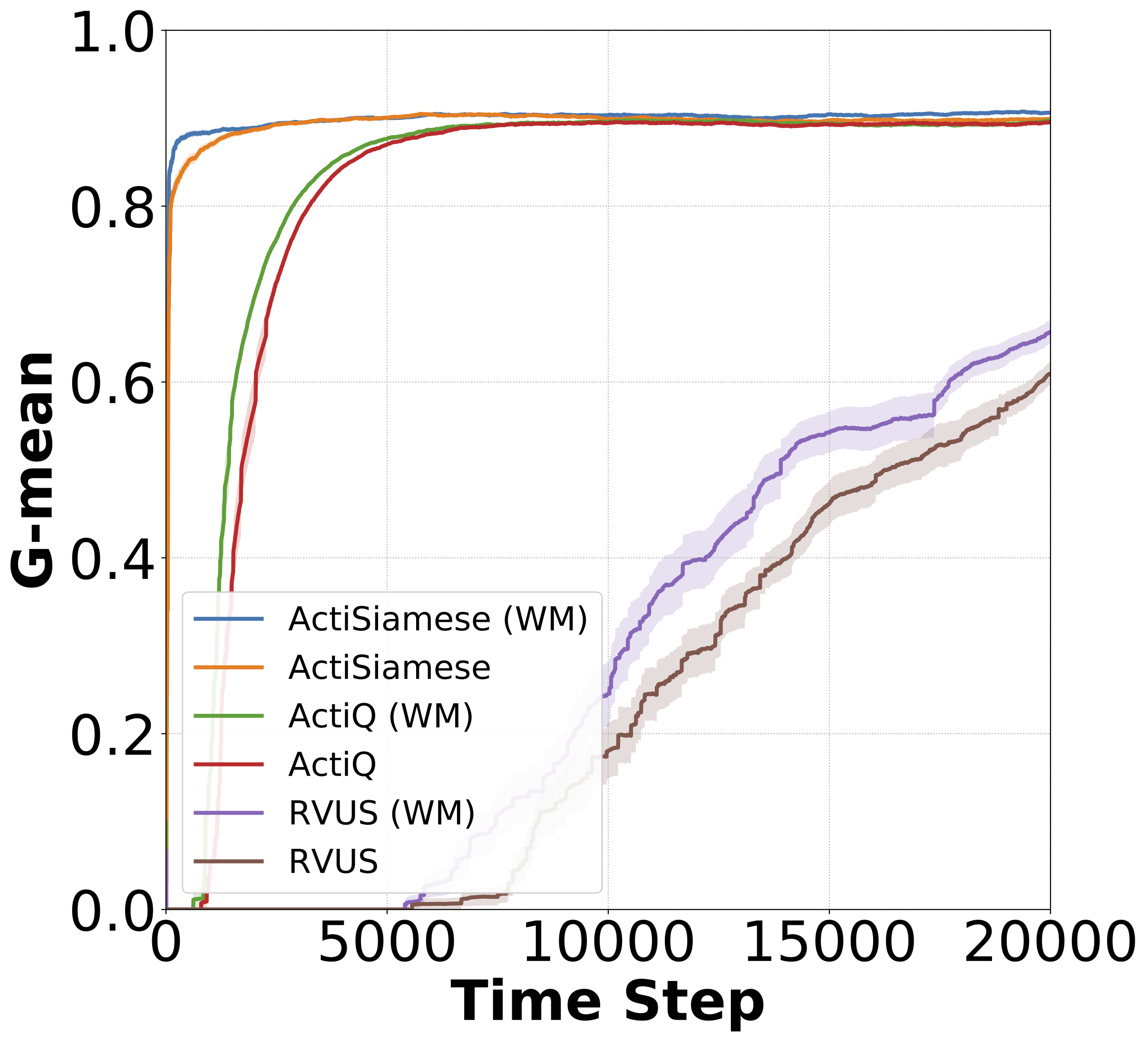}%
		\label{fig:stationary_blobs12_budget_001}}
	\subfloat[class imbalance]{\includegraphics[scale=0.13]{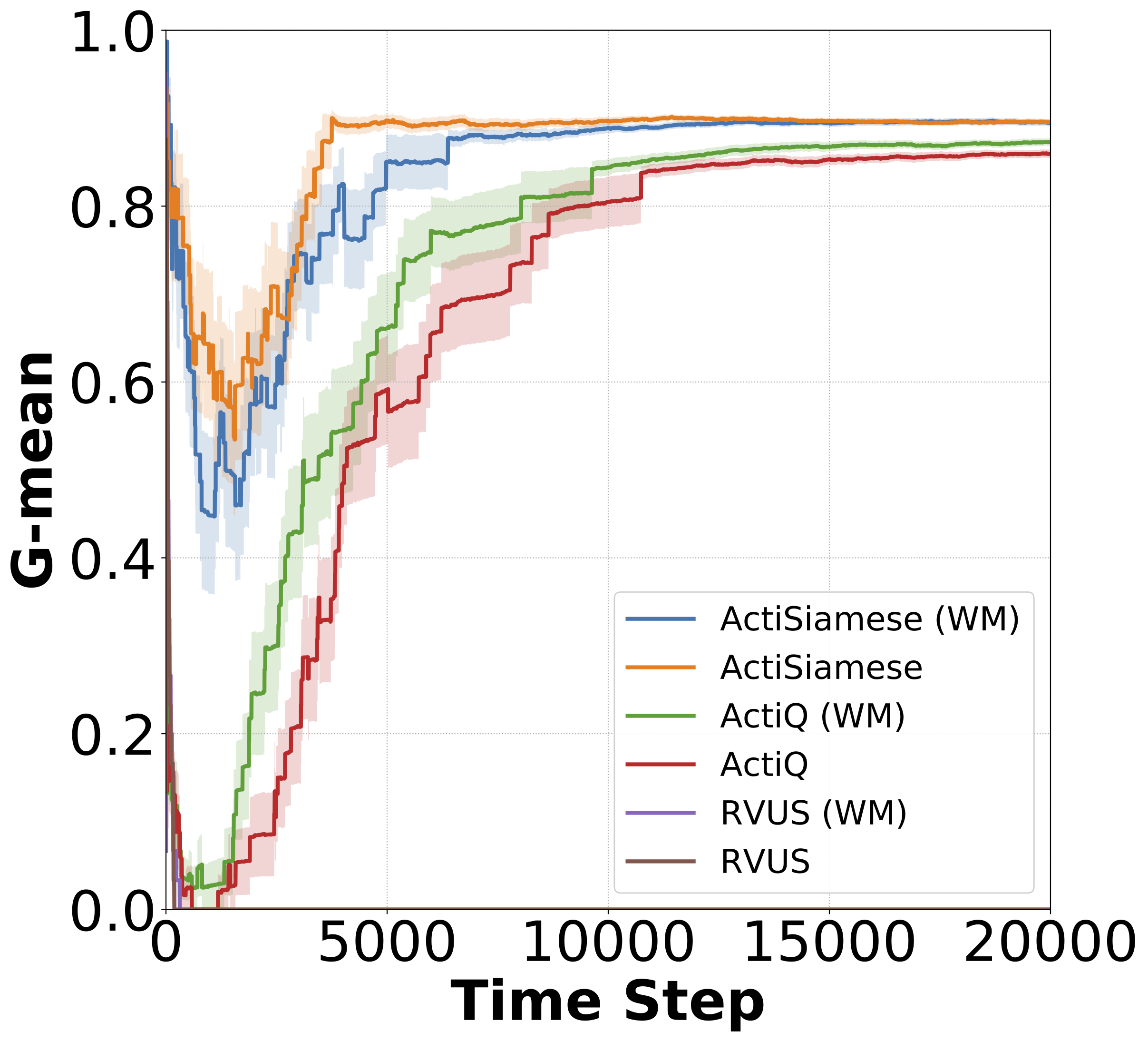}%
		\label{fig:imbalance_blobs12_mm_extreme}}
	\subfloat[abrupt drift]{\includegraphics[scale=0.13]{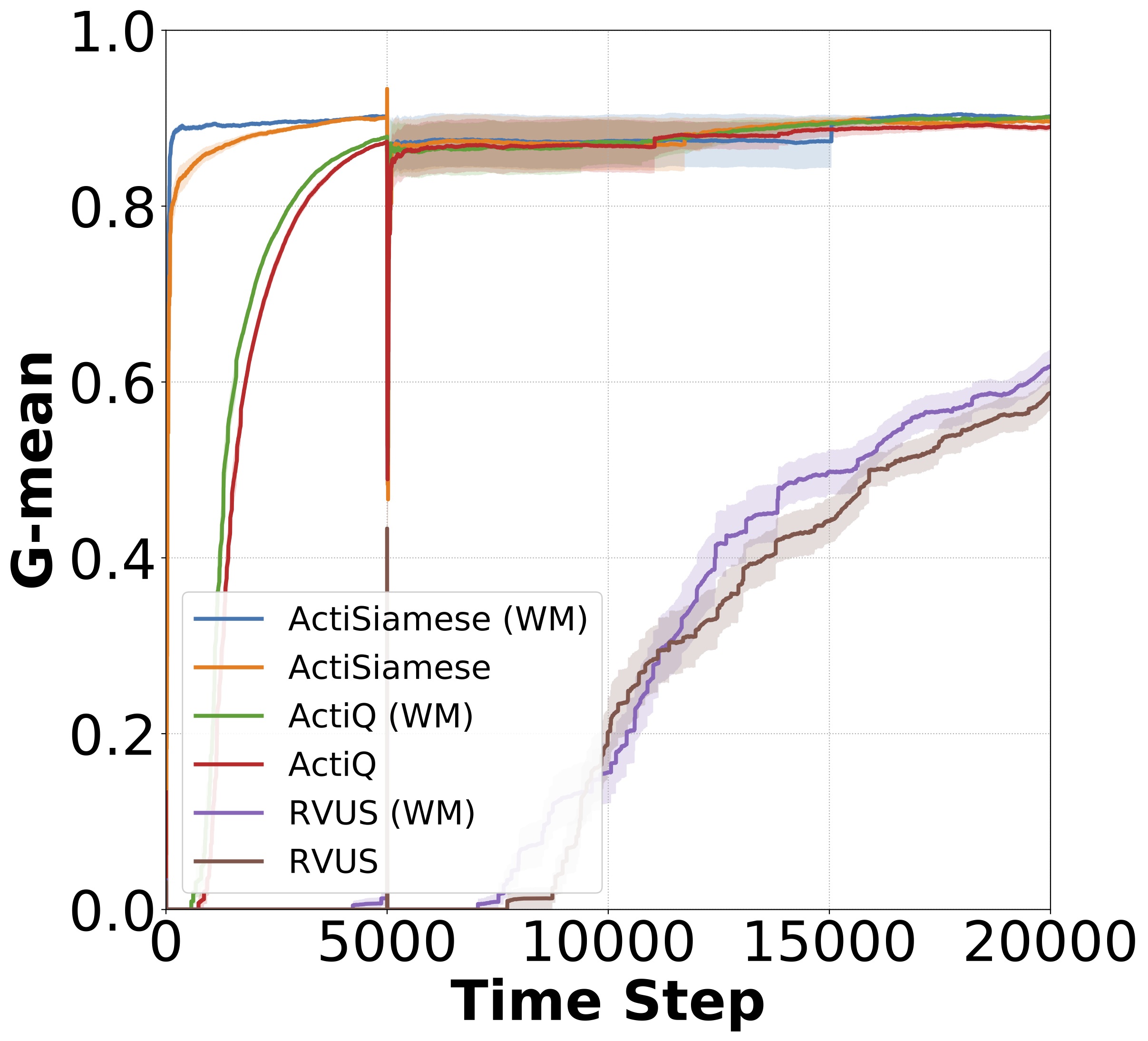}%
		\label{fig:drift_blobs12_budget_001}}
	
	\subfloat[imbalance + abrupt]{\includegraphics[scale=0.13]{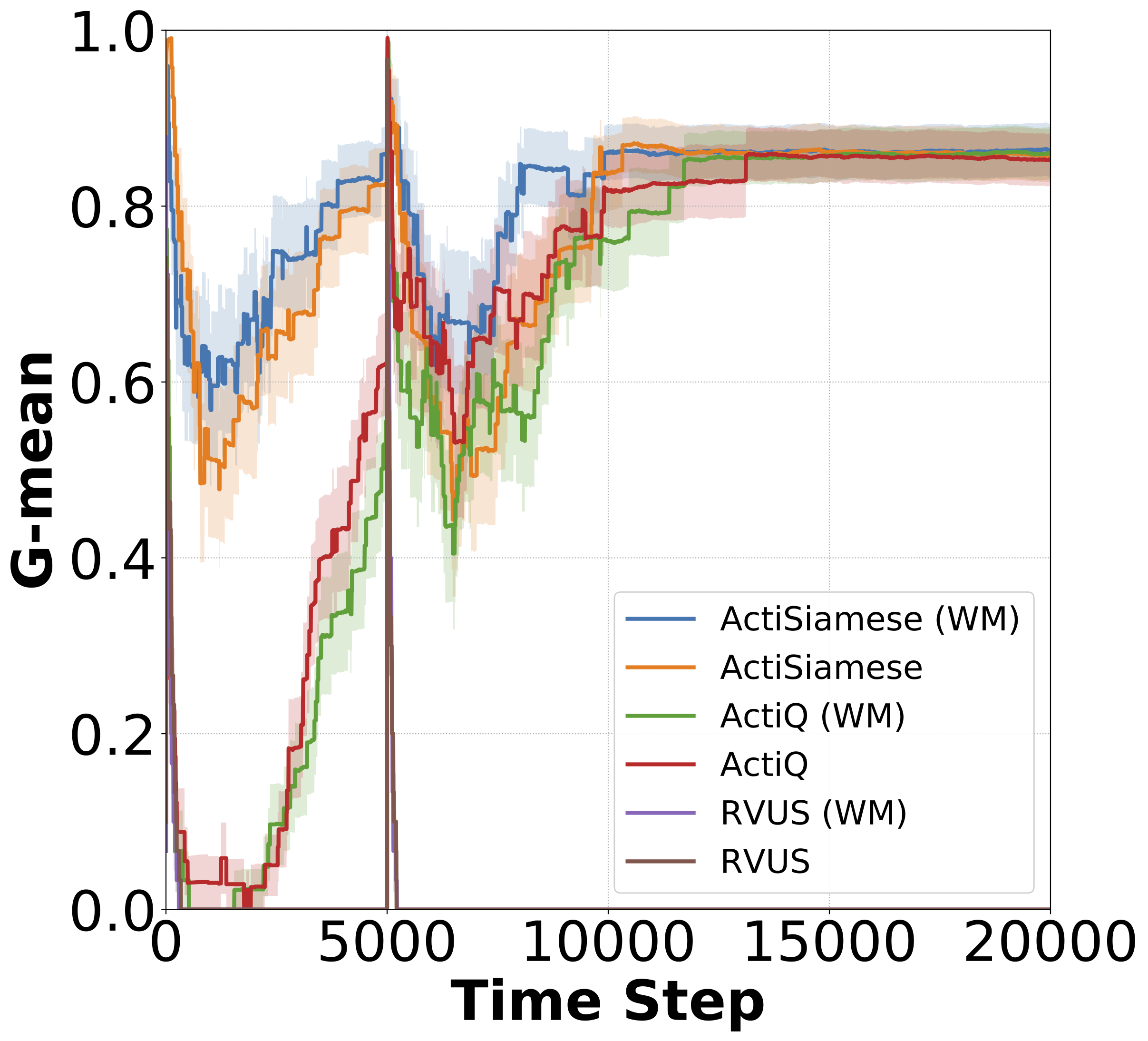}%
		\label{fig:drift_imbalance_blobs12_mm_extreme}}
	\subfloat[recurrent drift]{\includegraphics[scale=0.13]{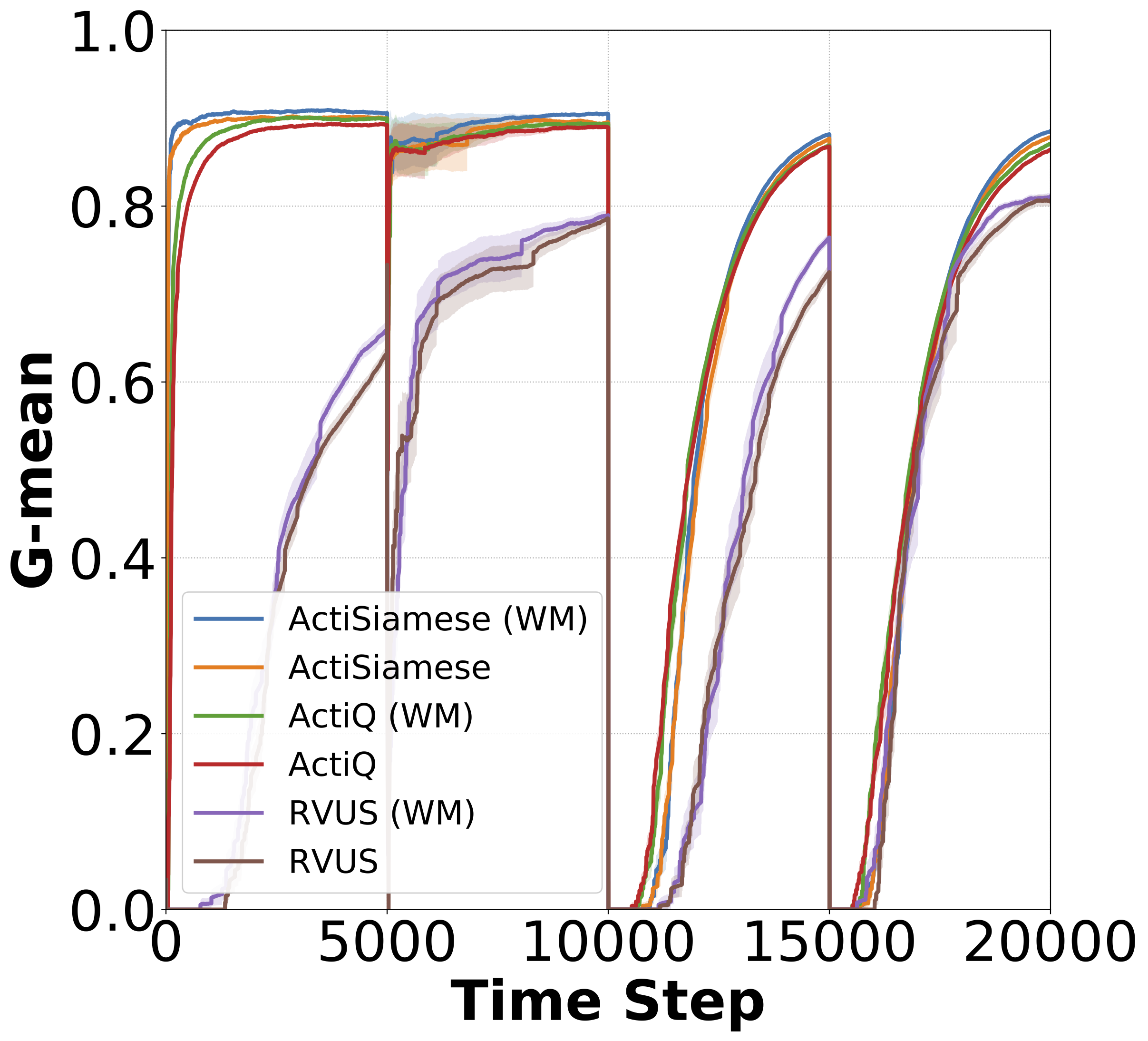}%
		\label{fig:drift_rec_blobs12_budget_005}}
	
	\caption{Comparative study in blobs: (a) normal (balanced, stationary), (b) extreme ($0.1\%$) imbalance, (c) abrupt drift, (d) extreme imbalance and abrupt drift, and (e) recurrent drift.}
\end{figure*}

The analogous plots for circles and blobs are shown in Figs.~\ref{fig:stationary_circles10_budget_001}-\ref{fig:drift_rec_circles10_budget_001} and 
Figs.~\ref{fig:stationary_blobs12_budget_001}-\ref{fig:drift_rec_blobs12_budget_005}. The budget is set to $B=1\%$ in all cases, except for recurrent drift in \textit{blobs} which is $B=5\%$ (Fig.~\ref{fig:drift_rec_blobs12_budget_005}). Overall, similar results are observed. A notable observation is that in Fig.~\ref{fig:stationary_circles10_budget_001}, given additional time, ActiQ slightly outperforms ActiSiamese. 

\begin{figure*}[t!]
	\centering
	
	\subfloat[circles (prior drift)]{\includegraphics[scale=0.13]{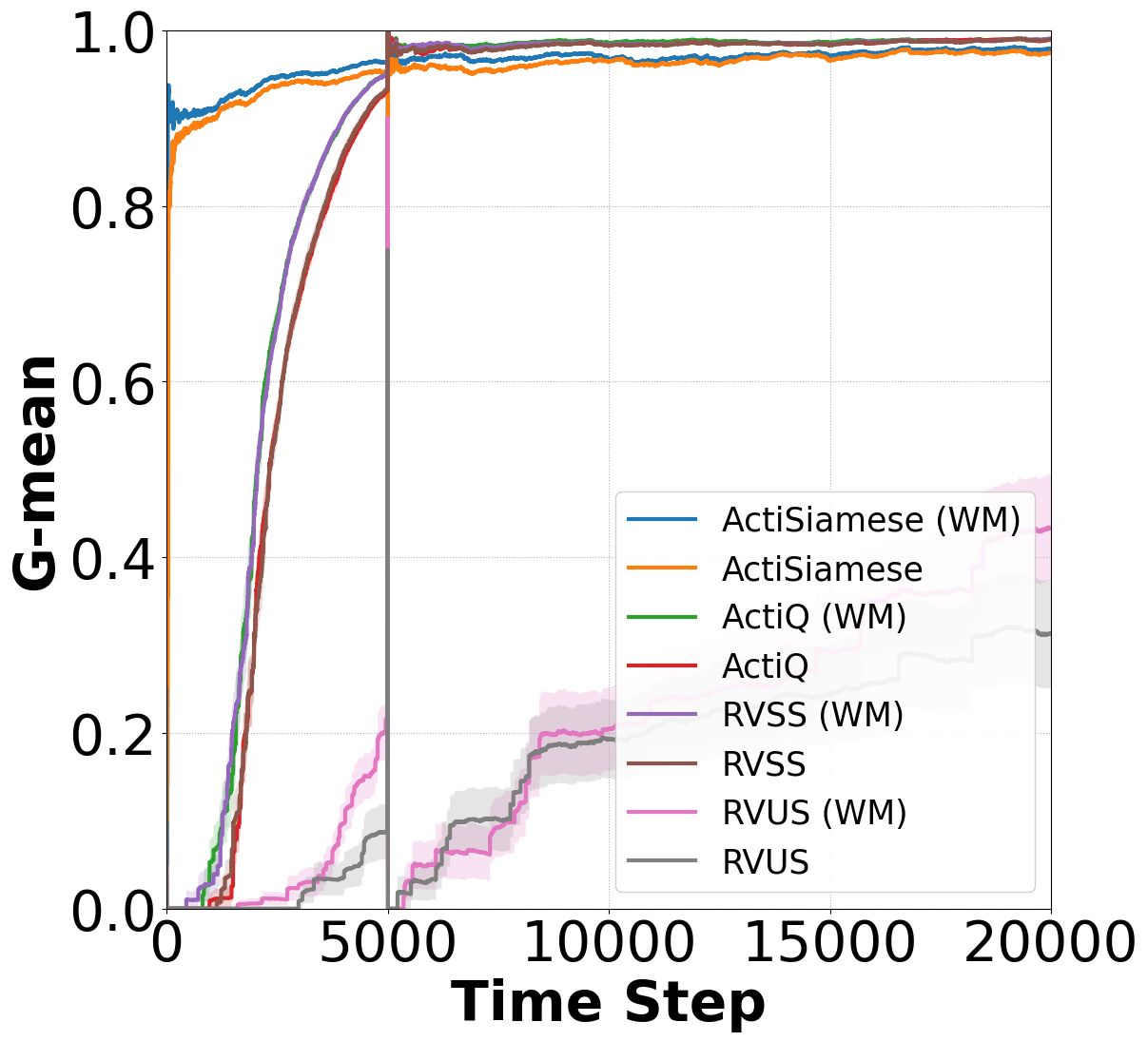}%
		\label{fig:synthetic_circle_10_01}}
	\subfloat[sea (prior drift)]{\includegraphics[scale=0.13]{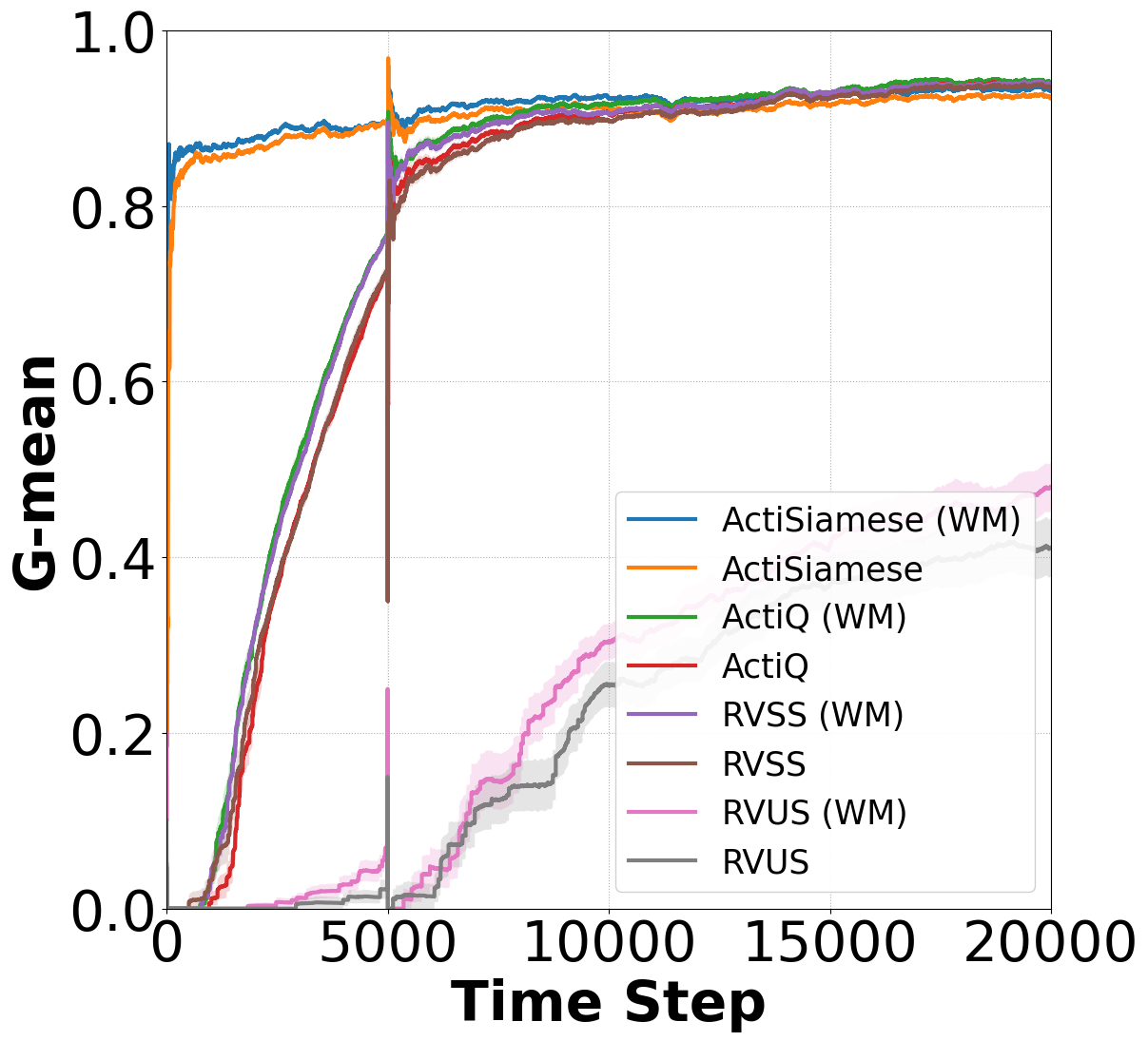}%
		\label{fig:synthetic_sea_10_01}}
	\subfloat[Two Patterns]{\includegraphics[scale=0.13]{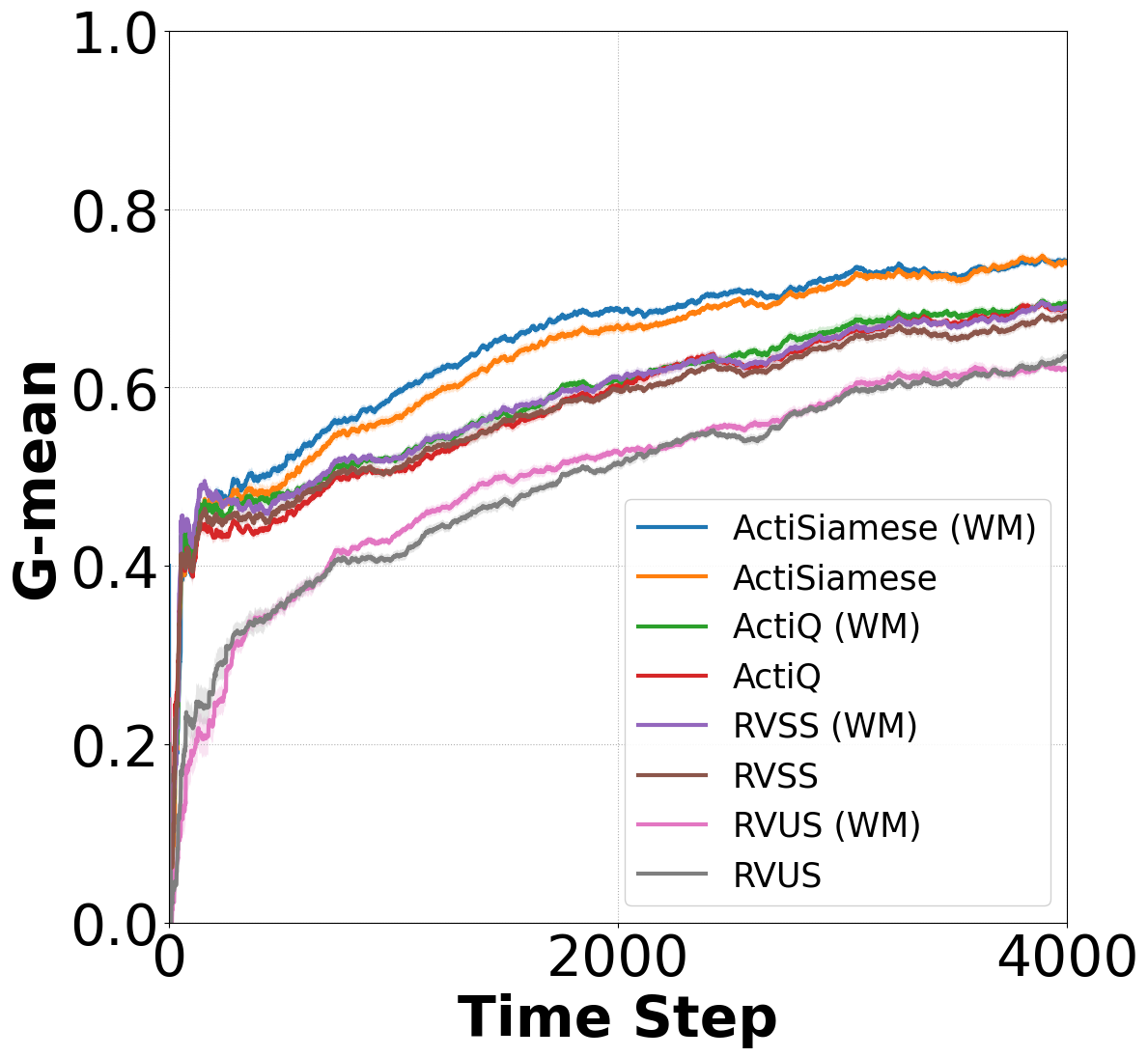}%
		\label{fig:synthetic_TwoPatterns_10_01}}
	
	\subfloat[Interchanging RBF]{\includegraphics[scale=0.13]{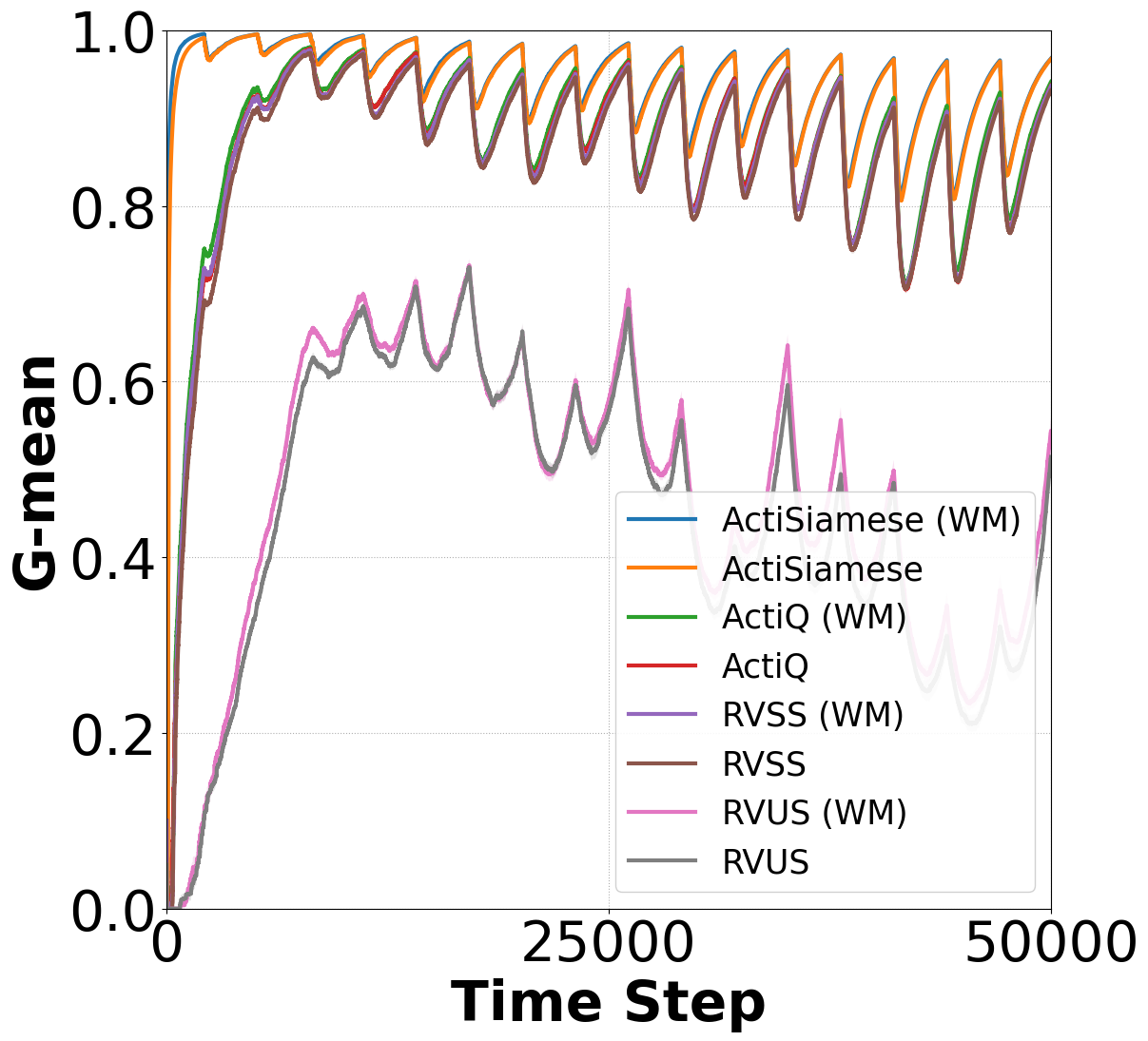}%
		\label{fig:synthetic_InterchangingRBF_10_03}}
	\subfloat[Moving Squares]{\includegraphics[scale=0.13]{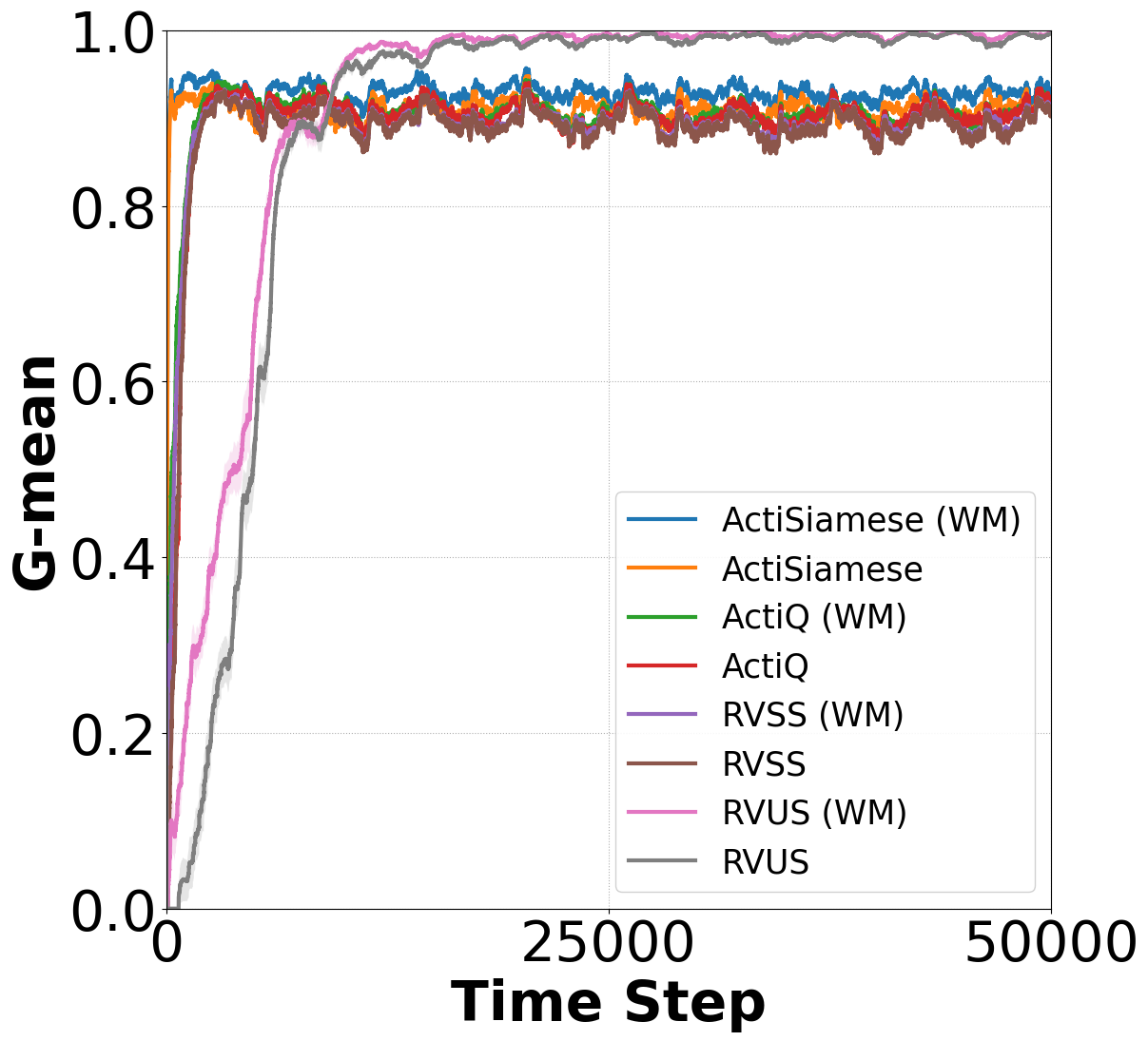}%
		\label{fig:synthetic_MovingSquares_10_03}}
	
	\caption{Comparative study for the rest of the synthetic datasets.}
\end{figure*}

The results for the remaining of the synthetic datasets are depicted in Figs.~\ref{fig:synthetic_circle_10_01}-\ref{fig:synthetic_MovingSquares_10_03}. The active learning is $B=10\%$ for circles, sea and Two Patterns, while for the more challenging Interchanging RBF and Moving Squares datasets it is set to $B=30\%$. In Figs.\ref{fig:synthetic_circle_10_01} and \ref{fig:synthetic_sea_10_01}, ActiSiamese has a superior learning speed; given additional time RVSS and ActiQ equalise its performance in sea and slightly outperform it in circles. In Figs.~\ref{fig:synthetic_TwoPatterns_10_01} and \ref{fig:synthetic_InterchangingRBF_10_03} ActiSiamese outperforms the rest. The ``oscillations'' observed in Fig.~\ref{fig:synthetic_InterchangingRBF_10_03} are due to interchanging nature of drift. Interestingly, in Fig.~\ref{fig:synthetic_MovingSquares_10_03}, RVSS outperforms the rest.

Taking everything into account, important remarks are as follows:
\begin{itemize}
	\item ActiSiamese, ActiQ, and RVSS that use the multi-queue memory significantly outperform the one-pass learner RVUS in all scenarios.
	\item Among the memory-based methods, ActiSiamese learns significantly faster than the rest. Given more time, ActiQ and RVSS may equalise ActiSiamese or may even slightly outperform it.
	\item ActiSiamese is superior under class imbalance.
	\item The methods that use the multi-queue memory deal with drift well, and they are robust to it. However, under challenging drift conditions, we had to increase the memory size and / or active learning budget.
	\item ActiSiamese, which considers similarity in the latent space significantly outperforms RVSS which considers similarity in the input space. An exception was observed in Fig.~\ref{fig:synthetic_MovingSquares_10_03}.
	\item Overall, ensembles gain a few percentages of performance for all methods.
\end{itemize}

\begin{figure}[t!]
	\centering
	
	\subfloat[Gestures (L=50,B=1\%)]{\includegraphics[scale=0.13]{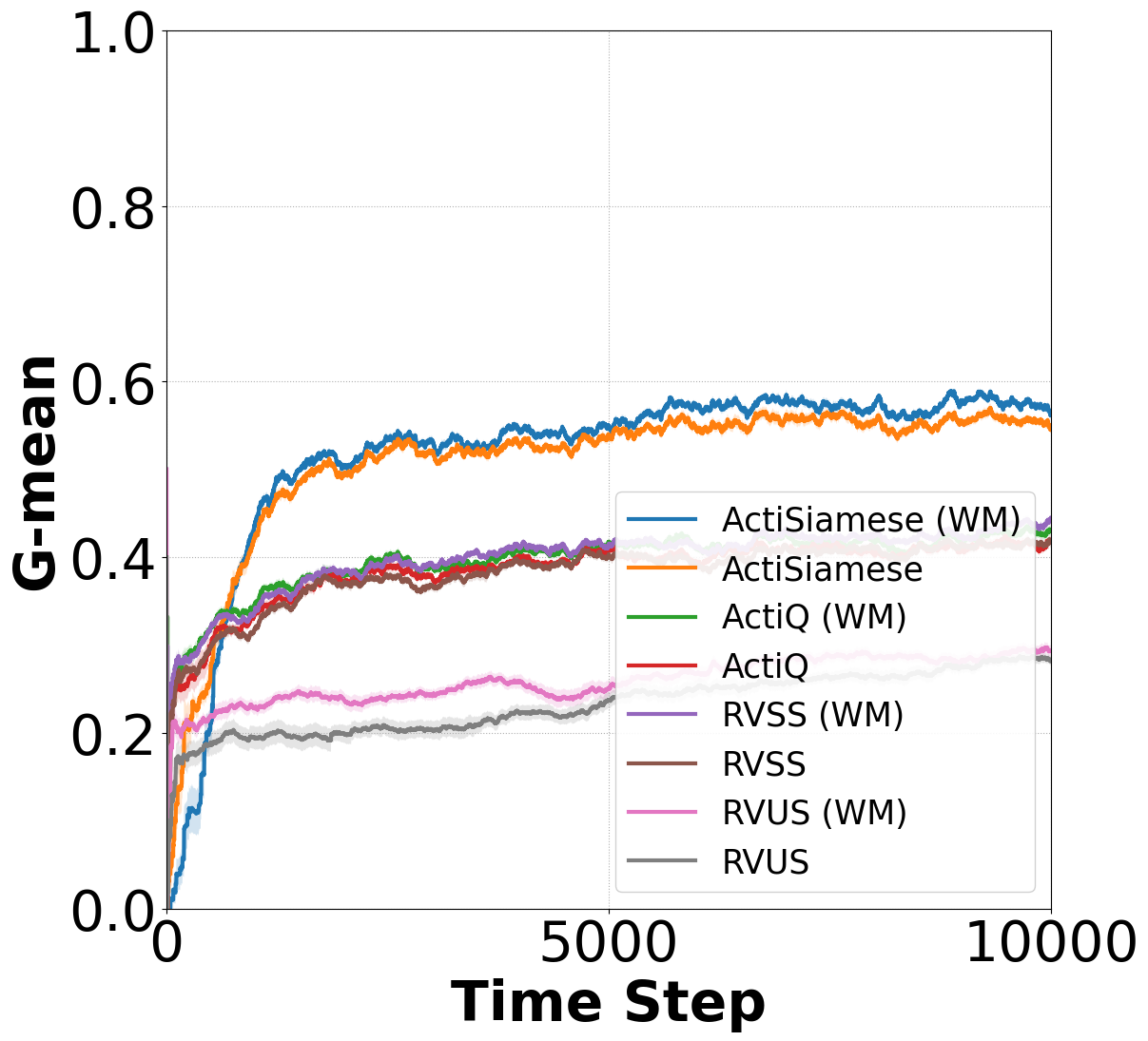}%
		\label{fig:real_gestures50_001}}
	\subfloat[MNIST (L=10,B=1\%)]{\includegraphics[scale=0.13]{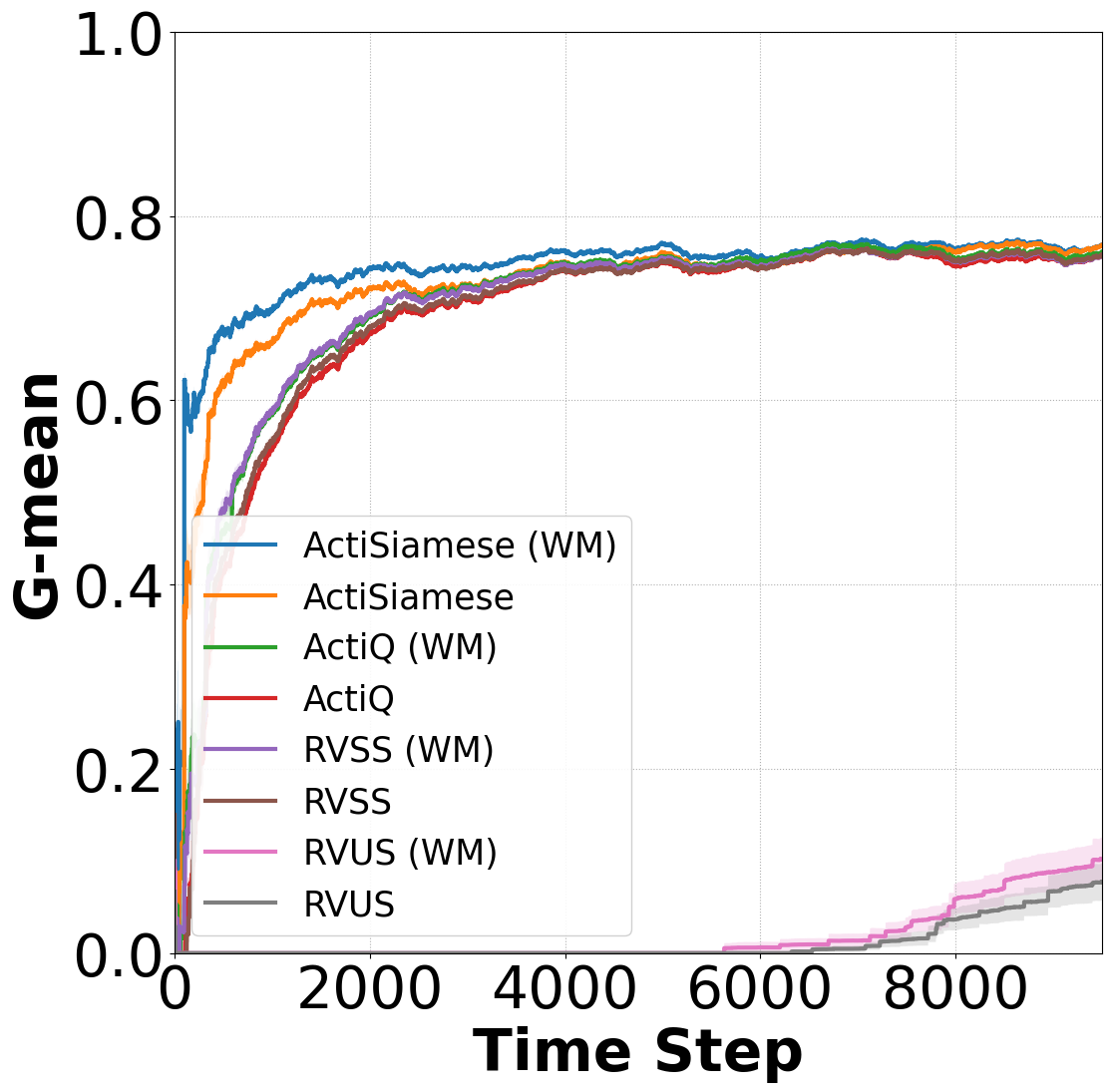}%
		\label{fig:real_mnist10_001}}
	\subfloat[Forest (L=50,B=1\%)]{\includegraphics[scale=0.13]{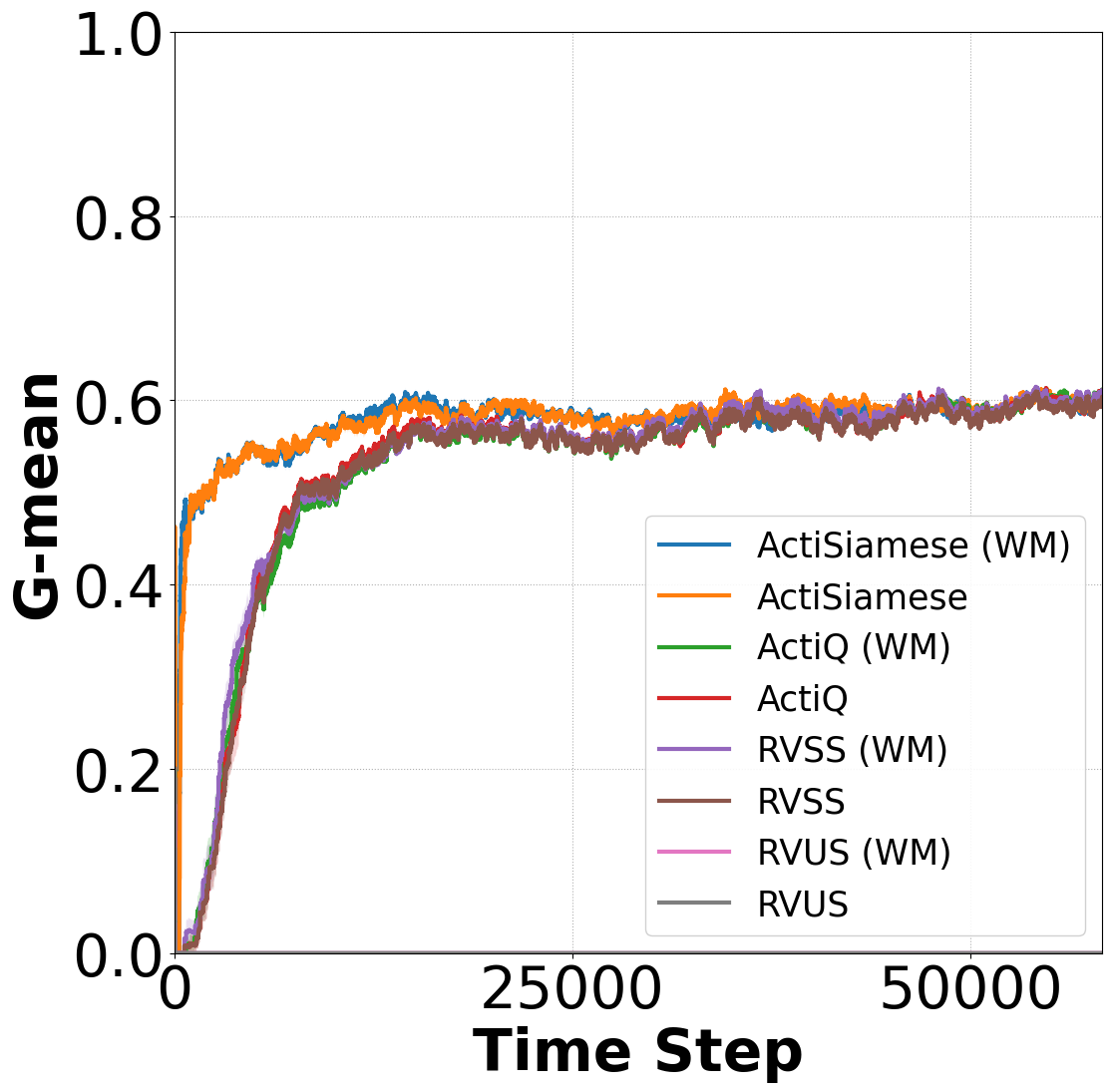}%
		\label{fig:real_forest50_001}}
	
	\subfloat[Keystroke (L=50,B=10\%)]{\includegraphics[scale=0.13]{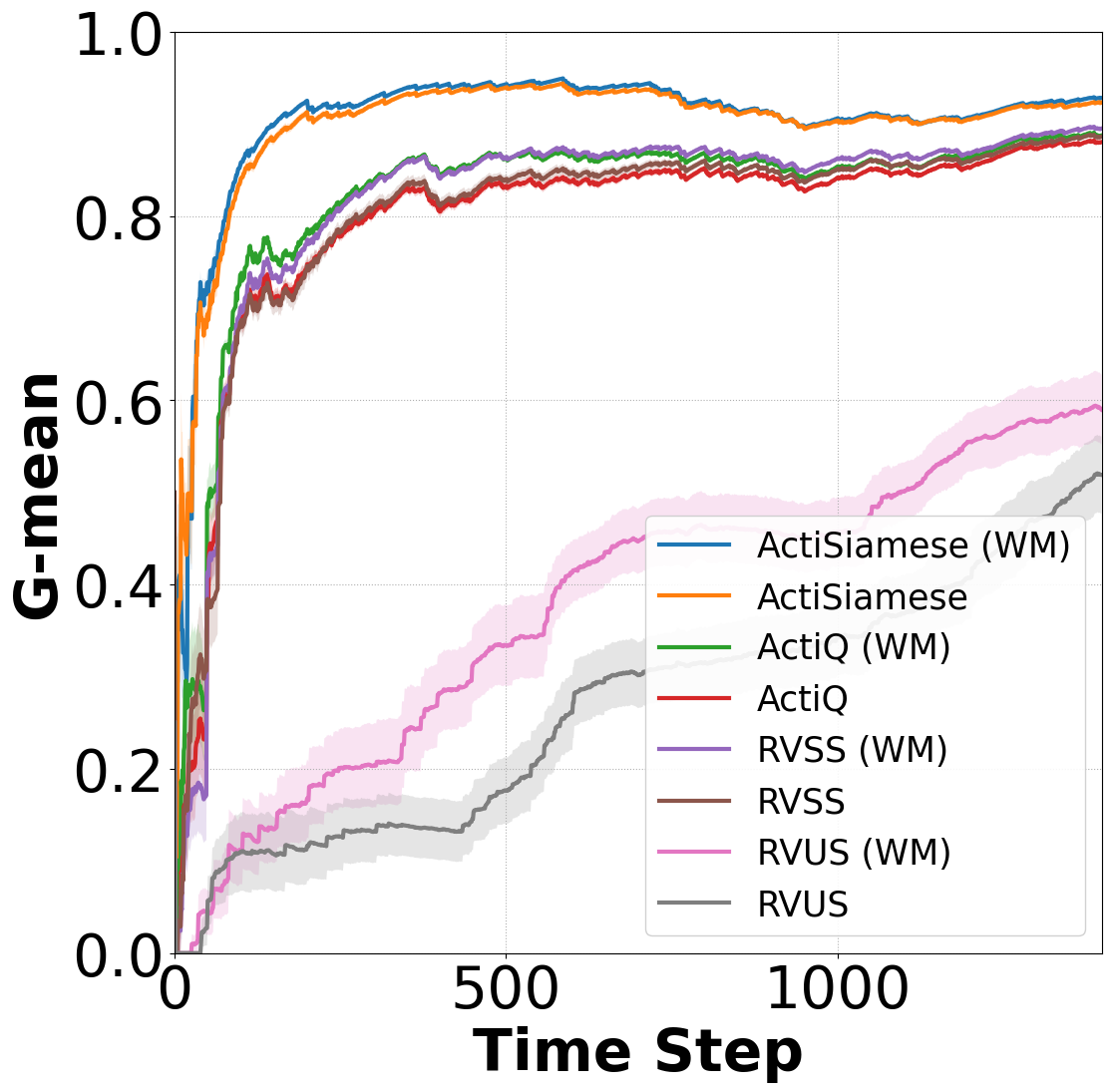}%
		\label{fig:real_keystroke50_01}}
	\subfloat[UWave Gestures (L=50,B=10\%)]{\includegraphics[scale=0.13]{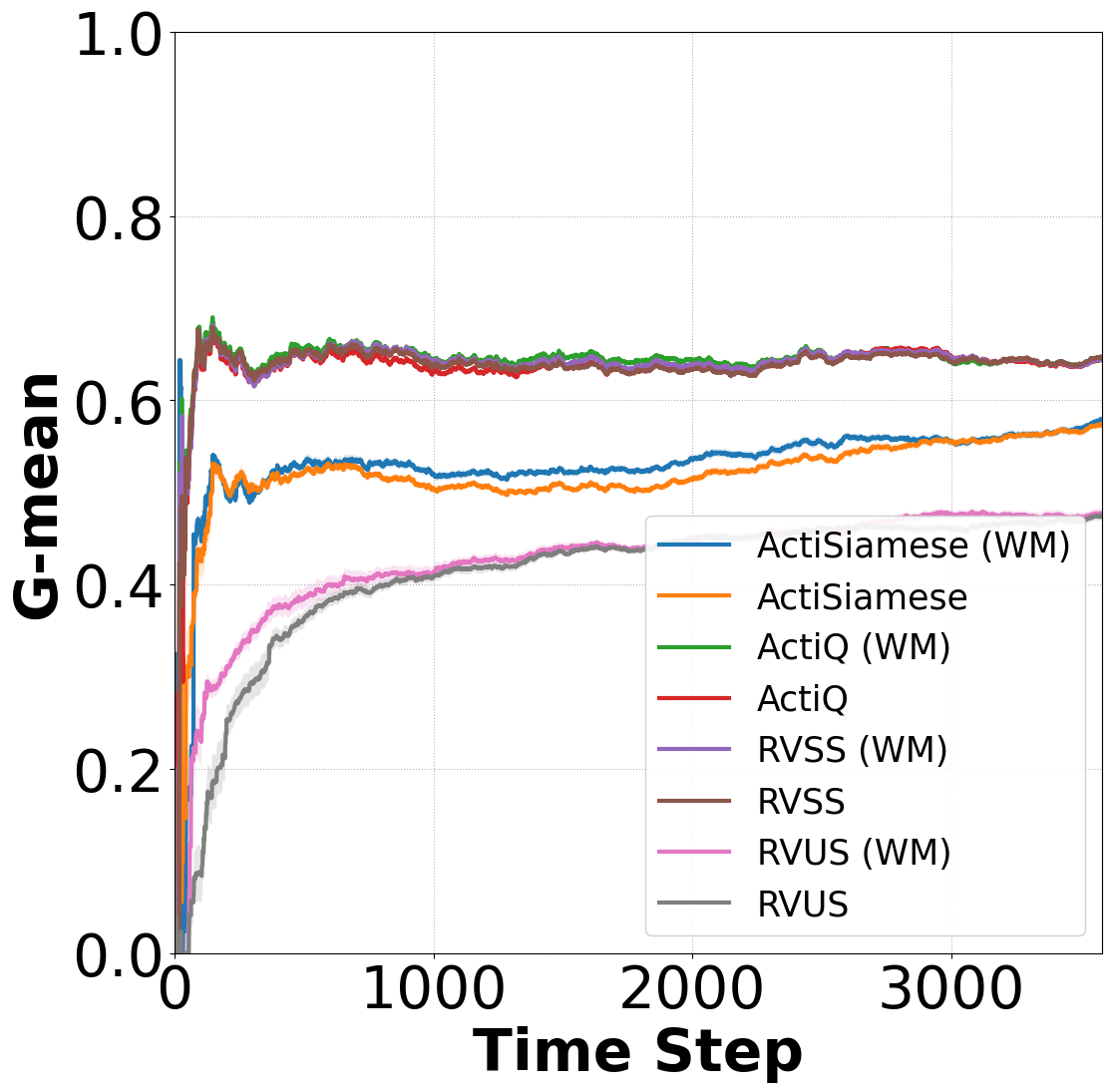}%
		\label{fig:real_uwavegestures50_01}}
	\subfloat[Insects (L=10,B=10\%)]{\includegraphics[scale=0.13]{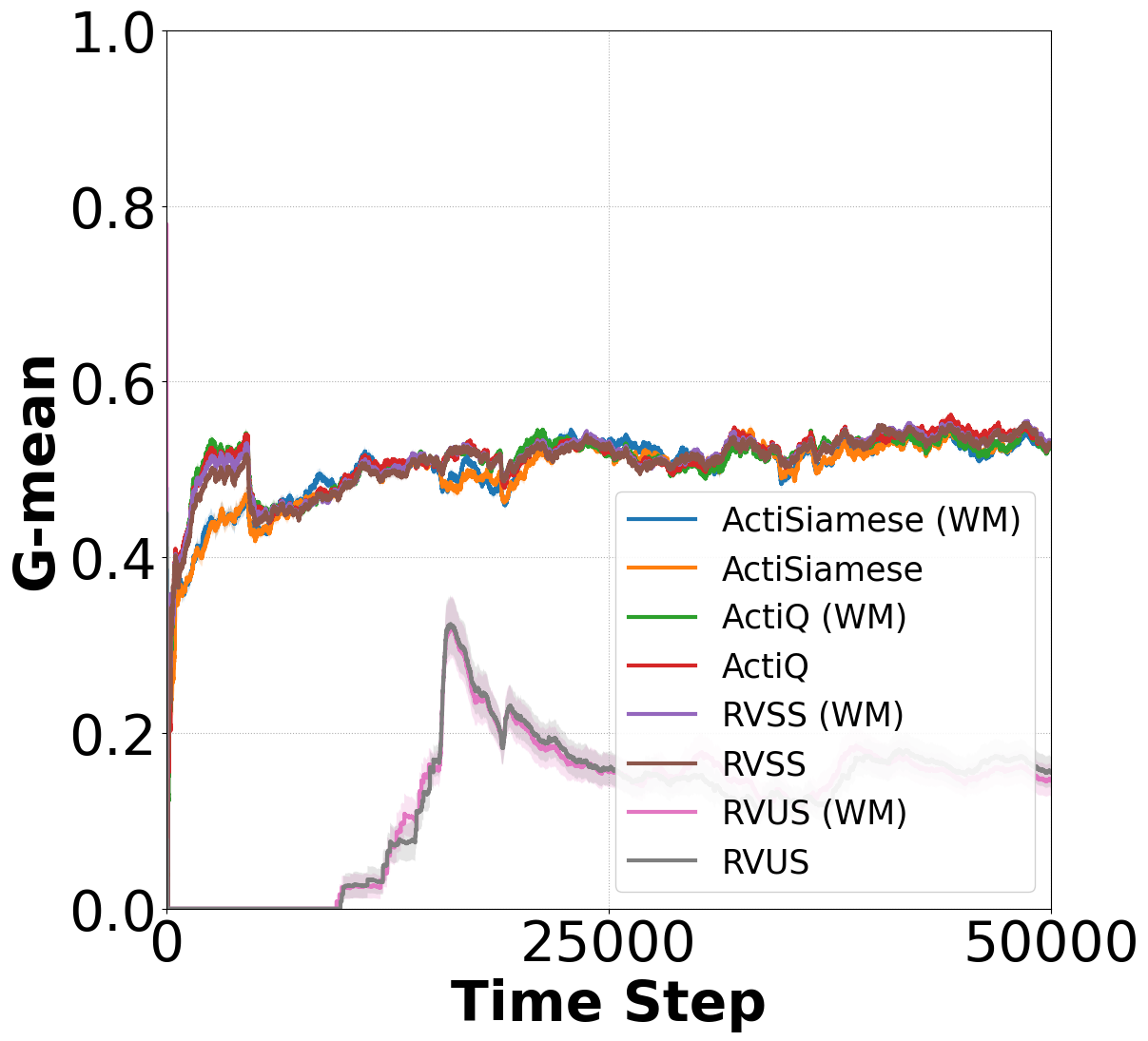}%
		\label{fig:real_insects10_01}}
	
	\caption{Comparative study for the real-world datasets.}
\end{figure}

\subsection{Real datasets}
This section describes our study in real-world datasets. Figs. \ref{fig:real_gestures50_001} - \ref{fig:real_insects10_01} show the performance of each method in the Gestures, MNIST, Forest, Keystroke, UWave Gestures, and Insects dataset respectively.

Recall that the active learning budget is critical to the success of a real-world application. A budget too large would correspond to too frequent interactions with a human expert, i.e., requesting ground truth information too often which could potentially cause a method to be impractical. It is for this reason, we have kept the budget to a low value; in Gestures, MNIST, and Forest (Figs.~\ref{fig:real_gestures50_001} - \ref{fig:real_forest50_001}) the budget is $B=1\%$, while in Keystroke, UWave Gestures, and Insects (Figs.~\ref{fig:real_keystroke50_01} - \ref{fig:real_insects10_01}) it is $B=10\%$. The memory size is $L=10$ for MNIST and Insects, and $L=50$ for the rest which have proven to be more challenging due to the small budget.

In summary, the results are in alignment with those obtained in the synthetic datasets:
\begin{itemize}
	\item The proposed methods (ActiSiamese, ActiSiamese-WM) significantly outperform the rest in two datasets (Gestures, Keystroke).
	
	\item ActiSiamese and ActiSiamese-WM learn significantly faster in two datasets (MNIST, Forest). In these two datasets, the final performance is similar to RVSS(-WM) and ActiQ(-WM) which they make use of the multi-queue memory of the proposed methods.
	
	\item An exception where ActiSiamese performs worse is in the UWave Gestures. Expectedly, the proposed method is not the silver bullet to all problems.
	
	\item The one-pass learner RVUS (no memory) is significantly worse than the rest, except in Forest which obtains a similar performance to them.
	
	\item Ensemble learning gains a few percentages of performance for all methods.
	
	\item In challenging problems, larger values of the memory $L$ and budget $B$, typically, improve the performance of all methods.
\end{itemize}

To verify the above, we further conducted a statistical analysis. The final performance (i.e., at the last time step) is examined for Gestures, UWave Gestures, and Insects, while the learning speed is examined early in the learning process ($t=500$) for MNIST, Forest, and Keystroke. Table~\ref{tab:stats} shows the mean performances and standard deviations. The method which yields the higher performance based on ANOVA and its posthoc tests (Section~\ref{sec:evaluation}) is shown in bold font which denotes statistical significance over the others.

\begin{table}[t!]\label{tab:stats}
	\centering
	\caption{Mean performance and standard deviation at the last time step (Gestures, UWave Gestures, Insects), and early in the learning process at $t=500$ (MNIST, Forest, Keystroke).}
	\label{tab:hyperparameters}
				\resizebox{\columnwidth}{!}{%
	\begin{tabular}{l|c|c|c|c|c|c}
		& \textbf{Gestures}        & \textbf{MNIST}           & \textbf{Forest}          & \textbf{Keystroke}       & \textbf{UWave Gestures}  & \textbf{Insects}          \\ 
		\hline
		\textbf{ActiSiamese (WM)} & \textbf{0.5616 (0.0189)} & \textbf{0.6785 (0.0238)} & \textbf{0.4654 (0.0571)} & \textbf{0.9428 (0.0041)} & 0.5778 (0.0175)          & \textbf{0.5263 (0.0208)}  \\ 
		\hline
		\textbf{ActiSiamese}      & 0.5456 (0.0191)          & 0.6193 (0.0376)          & 0.3653 (0.1857)          & \textbf{0.9393 (0.0061)} & 0.5729 (0.0180)          & \textbf{0.5258 (0.0179)}  \\ 
		\hline
		\textbf{ActiQ (WM)}       & 0.4311 (0.0239)          & 0.4415 (0.1563)          & 0.0000 (0.0000)          & 0.8627 (0.0091)          & \textbf{0.6440 (0.0139)} & \textbf{0.5247 (0.0185)}  \\ 
		\hline
		\textbf{ActiQ}            & 0.4175 (0.0191)          & 0.4172 (0.0518)          & 0.0000 (0.0000)          & 0.8322 (0.0212)          & \textbf{0.6441 (0.0097)} & \textbf{0.5269 (0.0168)}  \\ 
		\hline
		\textbf{RVSS (WM)}        & 0.4446 (0.0172)          & 0.4826 (0.0533)          & 0.0000 (0.0000)          & 0.8634 (0.0115)          & \textbf{0.6442 (0.0095)} & \textbf{0.5293 (0.0167)}  \\ 
		\hline
		\textbf{RVSS}             & 0.4193 (0.0284)          & 0.4289 (0.0414)          & 0.0000 (0.0000)          & 0.8396 (0.0253)          & \textbf{0.6463 (0.0116)} & \textbf{0.5259 (0.0159)}  \\ 
		\hline
		\textbf{RVUS (WM)}        & 0.2937 (0.0296)          & 0.0000 (0.0000)          & 0.0000 (0.0000)          & 0.3340 (0.1957)          & 0.4770 (0.0212)          & 0.1456 (0.0804)           \\ 
		\hline
		\textbf{RVUS}             & 0.2831 (0.0220)          & 0.0000 (0.0000)          & 0.0000 (0.0000)          & 0.1761 (0.1605)          & 0.4739 (0.0209)          & 0.1548 (0.0794)          
	\end{tabular}}
\end{table}

\section{Discussion}\label{sec:discussion}
In this section we discuss important remarks concerning ActiSiamese's computational aspects, advantages, and limitations.

\subsection{Computational aspects}\label{sec:cost}

\textbf{Memory requirements}: The memory size is $|Q^t| = K \times L$, and while the number of classes $K$ depends on the problem,
this work has demonstrated that ActiSiamese is effective when the queue capacity $L$ is, typically, small (e.g., ten). Due to the pair creation process (discussed in the point below), ActiSiamese makes use of $|Q^t_{train}|$ training data which is, typically, larger than $|Q^t|$. Importantly, ActiSiamese satisfies one of the most important and desired properties of learning in nonstationary environments, which is having a fixed amount of memory for any storage \cite{gama2014survey}. Ideally, however, an algorithm should be capable of one-pass learning; we discuss this below.

\textbf{Training stage}: A computational step of ActiSiamese is the creation of pairs $Q^t_{train}$ from the original examples $Q^t$. First, we have showed that ActiSiamese is effective even when the budget is low, hence, the pair creation process, which is initiated only during training, doesn't occur frequently. Second, only a fraction of the training pairs need to be re-calculated at each training step. Third, ActiSiamese is updated once ($num\_epochs = 1$) whenever trained. Not only this helps to reduce the training cost, it can also prevent overfitting.

\textbf{Prediction stage}: For prediction, ActiSiamese computes the average similarity of $x^t$ to its elements as shown in Eq.~(\ref{eq:siamese_predict}). Significant computation speed can be gained by pre-computing the encodings, therefore, avoiding multiple and repeated forward propagation computations. For the ensemble version ActiSiamese-WM, significant gains can be achieved using parallelisation.

\subsection{Advantages}\label{sec:advantages}

\textbf{Class imbalance}: ActiSiamese has three ``embedded'' mechanisms which make it robust to imbalance. First, the use of \textit{separate} and \textit{balanced} queues per class alleviates the problem as propagating old examples in the most recent training set is a form of oversampling. Second, the data preparation step creates $|Q^t_{train}|$ training pairs from the $K \times L$ examples found in the original $Q^t$. The number of generated pairs depends on the values of $K$ and $L$, however, most of the times it is expected that $|Q^t_{train}|$ will be considerably larger than $|Q^t|$, thus constituting another form of oversampling. Third, the number of positive and negative pairs is always balanced. As a result, we will later show that ActiSiamese achieves a superior performance even under extreme imbalance.

\textbf{Concept drift}: ActiSiamese also has three mechanisms which make it robust to drift. First, it uses incremental learning to continually update the model to reflect new changes as environments evolve. We have chosen this for a few reasons. Drift is a multifaceted problem which can be classified according to type, severity, speed, predictability, frequency, and recurrence \cite{minku2010impact}. As a result, it is hard to characterise and explicitly detect it in practise. Furthermore, there is no explicit drift detection mechanism that can perform well under any combination of drift characteristics \cite{barros2018large}. Therefore, we have decided to \textit{learn} the concept drift using incremental training, rather than performing a complete re-training when drift is detected. We discuss this further in our future work section. Second, ActiSiamese uses a multi-queue memory. The fact that examples are carried over a series of steps allows the classifier to ``remember'' old concepts. ``Forgetting'' old concepts is achieved when obsolete examples eventually drop off the queues. ActiSiamese is effective with small queue sizes, therefore, it ensures that obsolete examples will be discarded quicker. Third, ActiSiamese-WM which uses ensembling helps by de-prioritising obsolete classifiers.

\subsection{Limitations}

\textbf{Ome-pass learning}: In our diverse study, ActiSiamese has been shown to significantly outperform strong baseline and state-of-the-art methods under various conditions. We attribute this to the many characteristics of ActiSiamese which have been combined in a seamless and effective manner. These include, few-shot learning for performing similarity learning, incremental learning, and the various mechanisms to handle imbalance and drift as discussed in Section~\ref{sec:advantages}.

This is, however, at the expense of ActiSiamese not being a one-pass learner. That is, if learning occurs using the most recently queried example, and then discarding it (i.e., without storing it in a memory), it is termed one-pass learning. Having said this, ActiSiamese satisfies one of the most important and desired properties of learning in data streams, which is having a fixed amount of memory for any storage. Also, ActiSiamese has been shown to be effective when the queue capacity $L$ is, typically, small (e.g., in the order of ten). These are discussed in Section~\ref{sec:cost}.

\section{Conclusion}\label{sec:conclusion}
We proposed ActiSiamese which synergistically merges online active learning, a multi-queue memory, and siamese networks. It proposes a new density-based active learning strategy which considers similarity in the latent space rather than the input space. We conducted an extensive study where we show that ActiSiamese outperforms baseline and state-of-the-art algorithms, and is effective even under extreme imbalance, and even when only a fraction of the arriving instances' labels is available. Future work will examine the following:

\textbf{Semi-supervised learning}: A limitation of active learning is that it solely attempts to explore the search space by querying selected instances for their class labels. In other words, it ignores all the arriving instances for which their label is not requested. In contrast, semi-supervised learning assumes the initial availability of labelled data which the classifier is trained with, but it later uses its learnt knowledge to automatically classify the arriving instances \cite{chapelle2009semi}. Future work will consider combining ActiSiamese with semi-supervise learning.

\textbf{Explicit concept drift detection}: ActiSiamese has an implicit way of dealing with drift using incremental learning and a multi-queue memory. An explicit drift detection mechanism (e.g., using statistical tests) could allow better reaction to drift. Future work will consider incorporating such a mechanism.

\bibliographystyle{elsarticle-num}
\bibliography{mybibfile}

\end{document}